\documentclass{article}

\usepackage{microtype}
\usepackage{subfigure}
\usepackage{booktabs} 

\usepackage{hyperref}       
\usepackage{url}            
\usepackage{booktabs}       
\usepackage{amsfonts}       
\usepackage{nicefrac}       
\usepackage{microtype}      
\usepackage[linesnumbered,ruled,vlined, algo2e]{algorithm2e}
\usepackage{algorithmic,eqparbox,array}
\renewcommand\algorithmiccomment[1]{%
\hfill\#\ \eqparbox{COMMENT}{#1}%
}

\SetKw{KwBy}{by}
\usepackage{tikzit}

\tikzstyle{new style 0}=[fill=black, draw=black, shape=circle]
\tikzstyle{red}=[fill={rgb,255: red,191; green,0; blue,64}, draw={rgb,255: red,191; green,0; blue,64}, shape=circle]
\tikzstyle{orange}=[fill={rgb,255: red,255; green,128; blue,0}, draw={rgb,255: red,255; green,128; blue,0}, shape=circle]
\tikzstyle{green trans no line}=[opacity=0.4, fill={rgb,255: red,0; green,162; blue,162}]

\tikzstyle{thick line}=[-, line width=0.5mm, draw=black]
\tikzstyle{density 1}=[-, draw={rgb,255: red,84; green,167; blue,222}, fill={rgb,255: red,84; green,167; blue,222}]
\tikzstyle{density 2}=[-, draw={rgb,255: red,192; green,210; blue,222}, fill={rgb,255: red,192; green,210; blue,222}]
\tikzstyle{linear}=[-, line width=0.75mm, draw={rgb,255: red,109; green,0; blue,38}]
\tikzstyle{green trans}=[-, draw={rgb,255: red,0; green,162; blue,162}, opacity=0.4, fill={rgb,255: red,0; green,162; blue,162}, line width=0.5mm]

\usepackage{pgfplots}
\usepgfplotslibrary{fillbetween}
\usepackage{xcolor}
\definecolor{n_red}{RGB}{191, 0, 64}
\definecolor{n_orange}{RGB}{255, 128, 0}
\definecolor{n_purple}{RGB}{109, 0, 38}
\definecolor{n_turq}{RGB}{0, 162, 162}

\usepackage{multirow}

\def\arrvline{\hfil\kern\arraycolsep\vline\kern-\arraycolsep\hfilneg}

\usepackage{nicefrac}
\usepackage{amsmath}
\usepackage{amssymb}
\usepackage{bm}
\usepackage{esvect}
\usepackage{todonotes}
\usepackage{graphicx}

\usepackage{tikz}
\usepackage{bbm}
\usetikzlibrary{arrows}
\usetikzlibrary{bayesnet}
\usepackage{mathrsfs} 
\usepackage{hyperref}
\usepackage{xcolor}
\usepackage{comment}
\usepackage{array}
\usepackage[export]{adjustbox}
\usepackage{balance}
\usepackage[format=plain,labelfont={bf},textfont=it]{caption} 

\makeatletter
\newcommand*{\rsimdots}{%
\mathrel{%
\mathpalette\@rsimdots{}
}%
}
\newcommand*{\@rsimdots}[2]{%
\sbox0{$#1\sim\m@th$}%
\sbox2{$#1\vcenter{}$}
\dimen@=.75\ht2\relax 
\sbox2{$#1\cdot\m@th$}
\sbox2{
\rlap{\raisebox{\dimen@}{\copy2}}%
\raisebox{-\dimen@}{\copy2}%
}%
\sbox2{$#1\rotatebox[origin=c]{-45}{\copy2}$}%
\rlap{\hbox to \wd0{\hss\copy2\hss}}%
\copy0 %
}
\makeatother

\newcolumntype{?}{!{\vrule width 1.2pt}}

\newcommand{\pp}{\Omega}
\newcommand{\rr}{\mathbf{r}}
\newcommand{\RR}{\mathbf{R}}

\usepackage{hyperref}

\usepackage[accepted]{icml2021}

\icmltitlerunning{
Black-box density function estimation using recursive partitioning
}

\begin{document}

\twocolumn[
\icmltitle{Black-box density function estimation using recursive partitioning}



\icmlsetsymbol{equal}{*}

\begin{icmlauthorlist}
\icmlauthor{Erik Bodin}{bris}
\icmlauthor{Zhenwen Dai}{spot}
\icmlauthor{Neill D. F. Campbell}{bath}
\icmlauthor{Carl Henrik Ek}{camb}
\end{icmlauthorlist}

\icmlaffiliation{bris}{University of Bristol, United Kingdom}
\icmlaffiliation{bath}{University of Bath, United Kingdom}
\icmlaffiliation{spot}{Spotify, United Kingdom}
\icmlaffiliation{camb}{University of Cambridge, United Kingdom}

\icmlcorrespondingauthor{Erik Bodin}{mail@erikbodin.com}

\icmlkeywords{
Machine Learning,
ICML, Bayesian inference, Black-box Bayesian inference,
Bayesian Computation, Active sampling, Tree-structured domain partitioning,
}

\vskip 0.3in
]



\printAffiliationsAndNotice{}  

\begin{abstract}
We present a novel approach to Bayesian inference and general Bayesian computation
that is defined through a sequential decision loop.
Our method defines a recursive partitioning of the sample space.
It neither relies on gradients nor requires any problem-specific tuning,
and is asymptotically exact for any density function with a bounded domain.
The output is an approximation to the whole density function including the normalisation
constant, via partitions organised in efficient data structures.
Such approximations may be used for evidence estimation or fast posterior sampling,
but also as building blocks to treat a larger class of estimation problems.
The algorithm shows competitive performance to recent state-of-the-art methods
on synthetic and real-world problems including
parameter inference for gravitational-wave physics.
\end{abstract}

\section{Introduction}
\label{sec:introduction}

Bayesian methods require the computation of posterior densities, expectations, and model evidence.
In all but the simplest conjugate models, these calculations are intractable and require numerical approximations to be made.
In many application areas, multiple computational tasks are carried out on the same distribution or model of interest.
Using traditional methods, this typically involves specialised algorithms and expertise for different tasks,
reducing ease-of-use of the overall methodology,
and there is often a need for many expensive computations.
The individual computations can have low utilisation of past results,
increasing the total computational burden
and slowing down experimentation.

The shared object on which the methods and computations operate is a density function,
typically being an unnormalised joint distribution of parameters and fixed data.
What is preventing tractable computations is that the density function does not have an amenable functional form.
An ideal functional form would allow us properties that are typically
only associated with specific families of parametric distributions,
such as fast proportional sampling,
tractable expectations of functions,
deriving conditional and marginal densities,
and compute quantities like divergences.
There are families of methods dedicated to approximating the density function with a function of such form.
However,
these methods either assume specific original functional forms,
require known gradients,
or can be prohibitively expensive.
This is today a challenge often even for problems which are moderate in the number of random variables
\footnote{Such as constituting up to around ten random variables.},
as many applications areas entail joint distributions with difficult properties.
Such areas include the physical sciences,
hyperparameter inference,
and active learning, such as approximating distributions of inputs to functions.

In this work, we construct approximations which are amenable for tractable computation.
The approximation is a tree structure which can be computed ahead of time, and from which we can produce piecewise constant approximations amenable for respective task.
We present an efficient algorithm to produce such trees, driving a recursive partitioning of the domain.
The algorithm is designed for black-box settings,
does not rely on gradients nor a known form,
and addresses challenges such as multi-modality, discontinuous structures, and zero density regions.
Our approach is asymptotically exact and has no sensitive free parameters;
this leads to an algorithm that accommodates general density functions of moderate dimension, and that is easy to use.
The method defines a sequential decision loop sampling the integrand,
prioritising regions of high probability mass within a recursive refinement of the approximation.
The decision loop has similar algorithmic efficiency as MCMC and nested sampling,
with a time complexity allowing for sub-millisecond decision times and scalability to a large number of observations to obtain sufficient precision.
It is competitive to recent state-of-the-art implementations of these methods for their individually supported tasks,
whilst in contrast to these methods produce a \emph{density function approximation} defined over the whole domain,
useful for many down-stream tasks beyond efficient sampling and evidence estimation.
This algorithm we refer to as DEnsity Function Estimation using Recursive partitioning or DEFER.

\begin{figure}[h]

  \centering
  \includegraphics[trim=10 10 10 10, clip, width=0.38\columnwidth, frame]{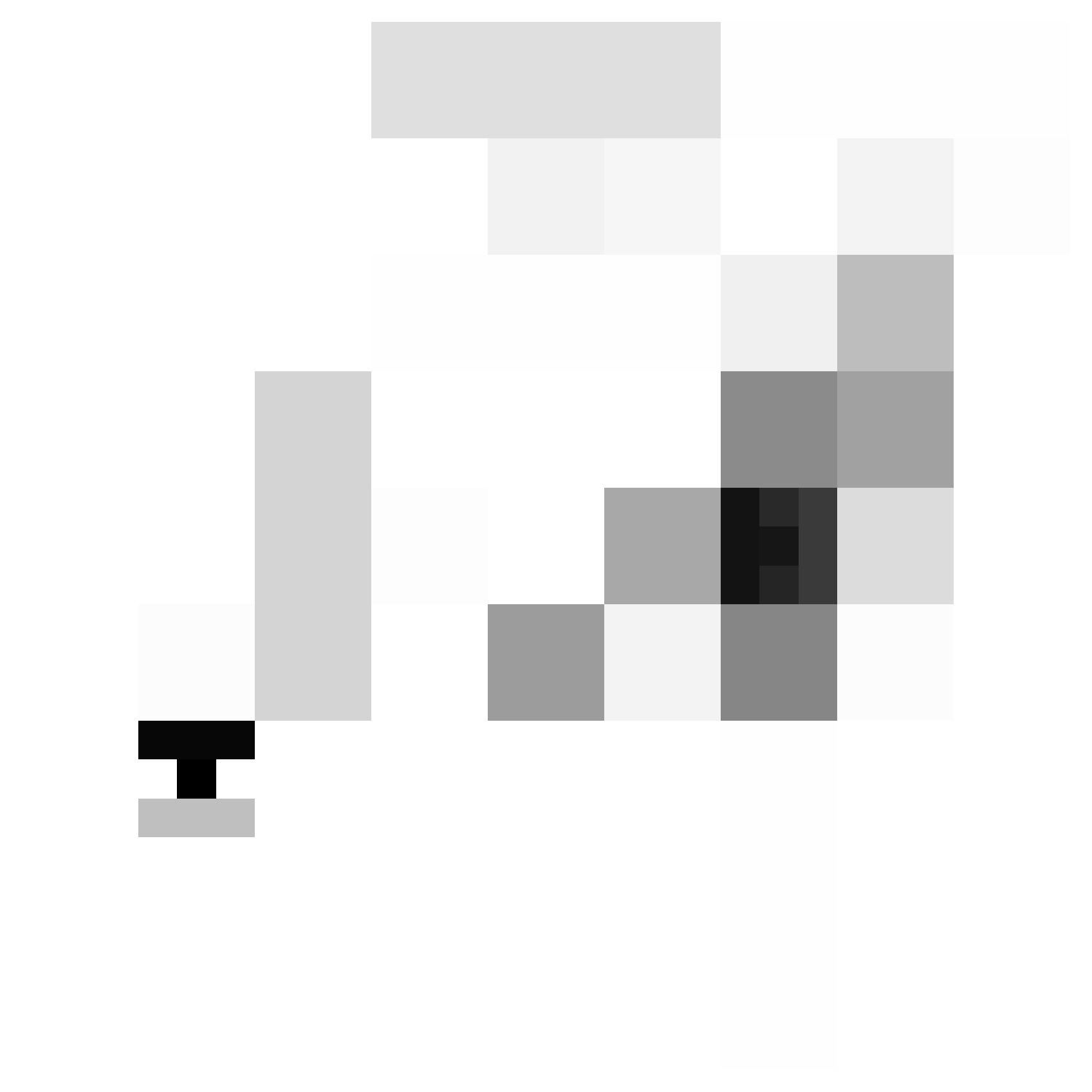}
  \hspace{-0.05cm}
  \includegraphics[trim=10 10 10 10, clip, width=0.38\columnwidth, frame]{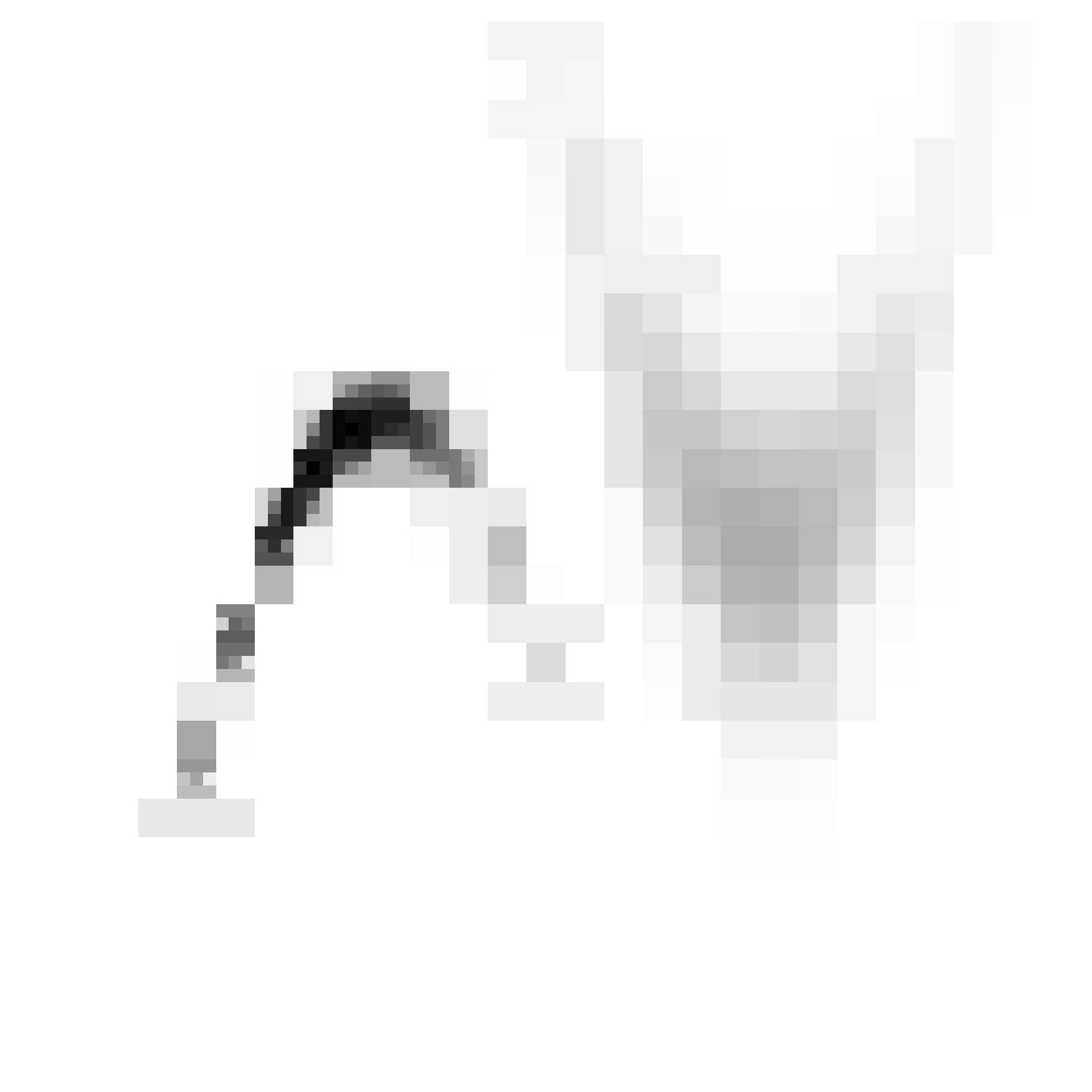} \\
  \vspace{0.08cm}
  \includegraphics[trim=10 10 10 10, clip, width=0.38\columnwidth, frame]{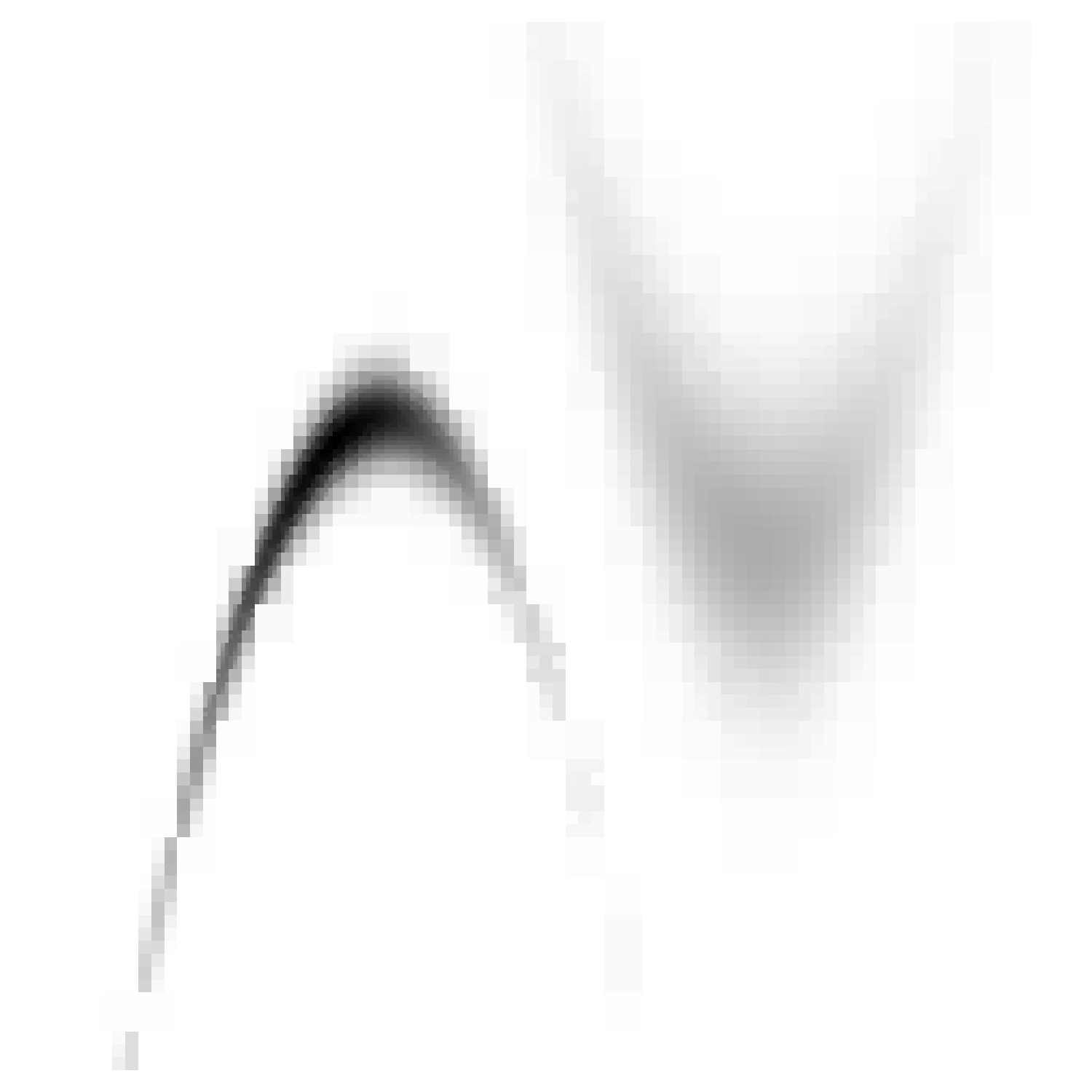}
  \hspace{-0.05cm}
  \includegraphics[trim=10 10 10 10, clip, width=0.38\columnwidth, frame]{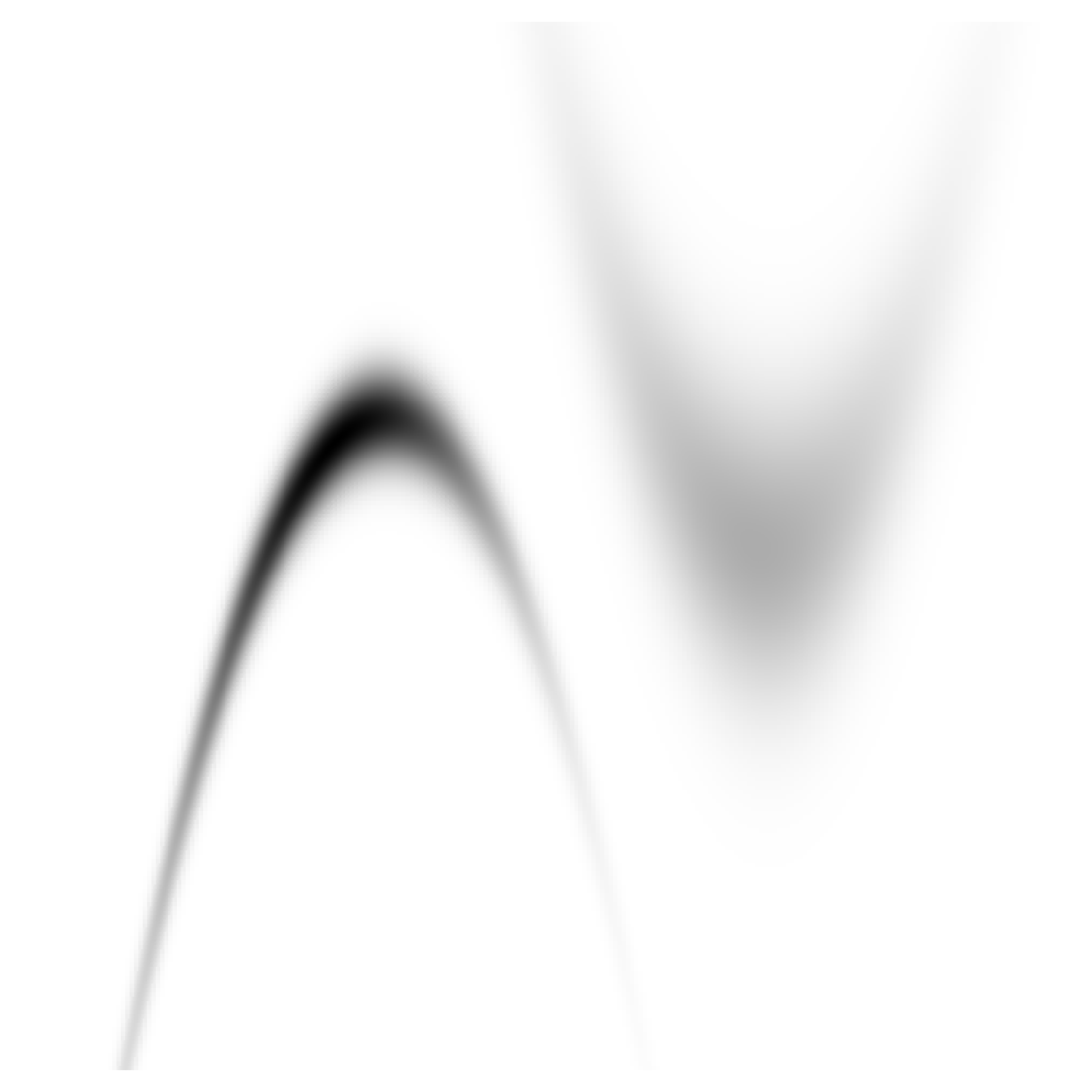}

  \caption{
  Partitions produced using DEFER, after 53, 303, 2\,101, and 100\,003 density function evaluations, respectively.
  }
  \label{fig:partitions}
\end{figure}

\section{Background}
\label{sec:background}

\begin{figure}[t]
  \includegraphics[width=0.485\linewidth]{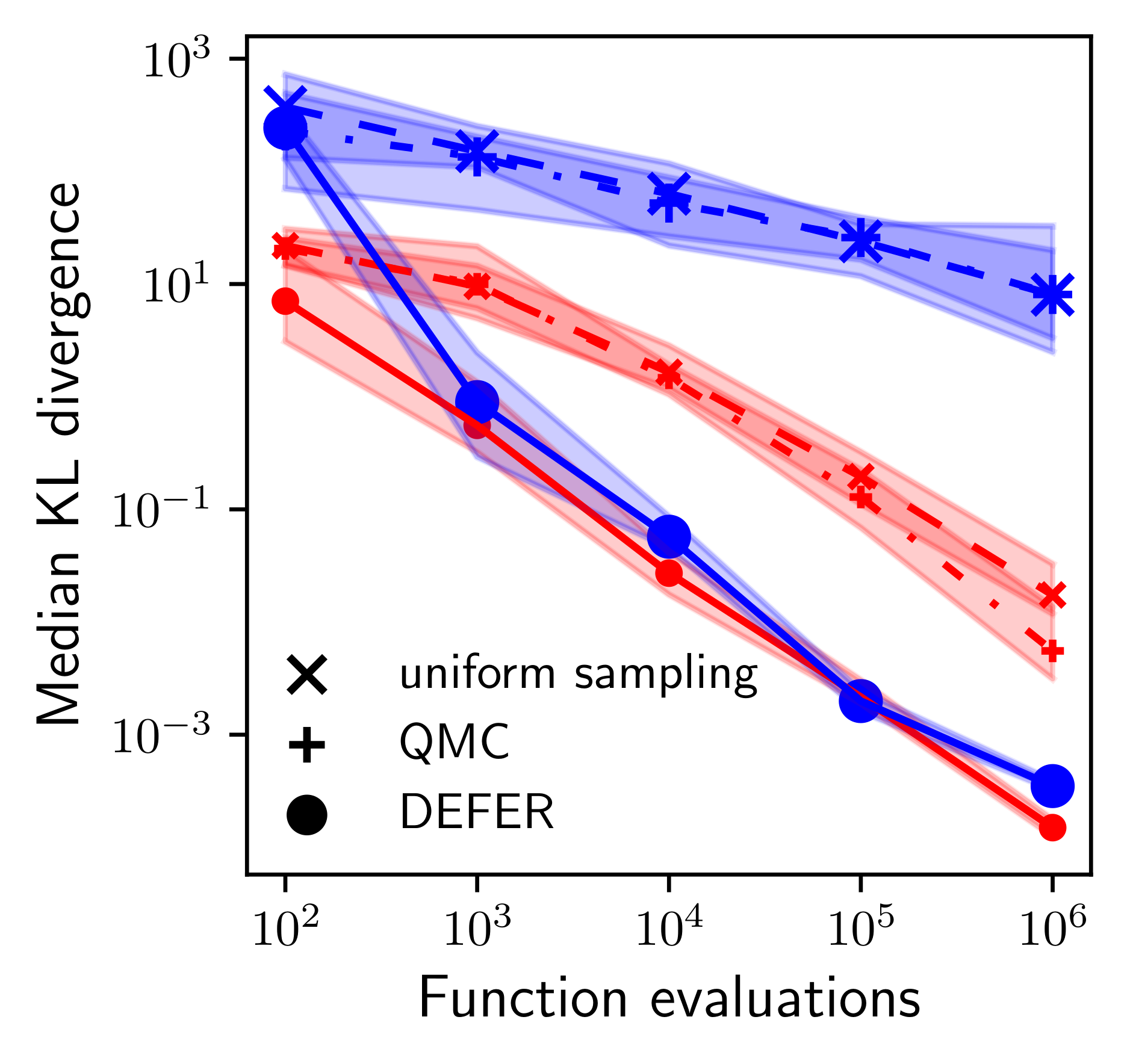}
  \includegraphics[width=0.495\linewidth]{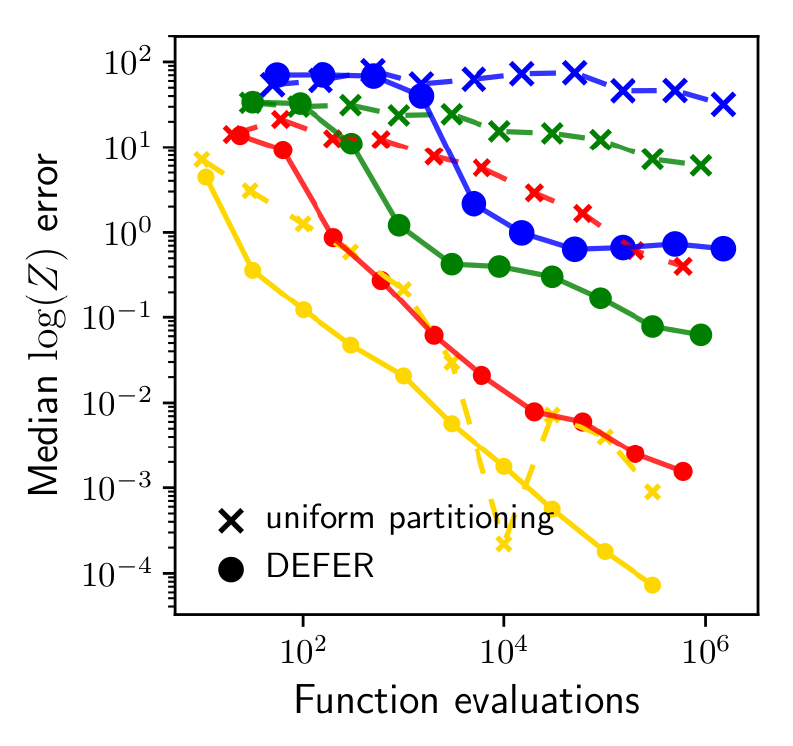}
  \caption{
  Shown on the left is an illustrative comparison to using standard rejection sampling for parameter inference
  on a Gaussian distribution in 5D, with a true scale parameter of $\nicefrac{1}{20}$ and $\nicefrac{1}{100}$ of the domain sides,
  corresponding to small and large marker sizes, respectively.
  The rejection methods use proposals drawn uniformly or in a Sobol sequence (Quasi-Monte Carlo), respectively.
  Shown on the right is a comparison to using a uniform grid for evidence estimation on the Student’s t-distribution
  in 2D, 4D, 6D, and 10D corresponding to the markers of increasing size.
  The true scale parameter corresponds to $\nicefrac{1}{100}$ of the domain sides.
  Both experiments were run 20 times with uniformly sampled true means,
  and the Student's t-distribution has $2.5 + (D / 2)$ degrees of freedom.
  }
  \label{fig:naive_methods}
\end{figure}

The need for substituting continuous distributions or processes with approximations exists in various forms in literature.
For optimisation problems in economics~\cite{tanaka2013discrete,tanaka2015discretizing,farmer2017discretizing},
and engineering~\cite{nguyen2018optimal,ai2013discrete} discrete approximations are often used,
but these methods generally rely of knowing the parametric form~\cite{miller1983discrete, devuyst2007gaussian}.
Variational Inference (VI)~\cite{blei16_variat_infer,Hoffman:2013tz} seeks a distribution approximation from a choice of variational family.
However, VI typically requires differentiability and known gradients, which is not met in our setting.
~\citet{acerbi2018variational} combined VI with Bayesian Quadrature~\cite{o1991bayes} and a Gaussian Process (GP) surrogate to
fit the variational distribution of an explicitly unknown distribution,
and~\citet{jarvenpaa2020batch} uses a GP surrogate for Approximate Bayesian Computation (ABC).
These methods are useful for scenarios where the unknown density function is significantly more expensive
to evaluate than the predictive distribution of the GP surrogate,
but this is typically not the case in the settings we address.
To put these costs in perspective,
evaluating the density function in the typical inference scenarios we address comes at a cost in the range of \emph{milliseconds},
whilst the method in~\cite{acerbi2018variational} takes around a \emph{second} to decide where to evaluate already after a few evaluations.
Furthermore, the time complexity of that method is $O(N^4)$,
whilst we need a near-linear algorithm scaling to many thousands or millions of evaluations.
Our method is, e.g., suited for the inner-loop of such methods, such as inference of the hyperparameters of the Gaussian Process
model these methods use.

\begin{figure*}
  \centering

  \raisebox{-0.56\height}{
  \resizebox{0.31\linewidth}{!}{
  \begin{tabular}{@{}c@{}}
    \resizebox{1.05\textwidth}{!}{%
    \centering\begin{tikzpicture}[scale=0.9]
	\begin{pgfonlayer}{nodelayer}
		\node [style=none] (0) at (-8, 6) {};
		\node [style=none] (1) at (12, 6) {};
		\node [style=none] (2) at (12, -6) {};
		\node [style=none] (3) at (-8, -6) {};
		\node [style=none] (4) at (-4, 6) {};
		\node [style=none] (5) at (-4, -6) {};
		\node [style=none] (6) at (-1.33, -6) {};
		\node [style=none] (7) at (-1.33, 6) {};
		\node [style=none] (8) at (5.33, 6) {};
		\node [style=none] (9) at (5.33, -6) {};
		\node [style=none] (10) at (-1.33, -2) {};
		\node [style=none] (11) at (-1.33, 2) {};
		\node [style=none] (12) at (5.33, 2) {};
		\node [style=none] (13) at (5.33, -2) {};
		\node [style=none] (14) at (0.89, 2) {};
		\node [style=none] (15) at (3.11, 2) {};
		\node [style=none] (16) at (0.89, -2) {};
		\node [style=none] (17) at (3.11, -2) {};
		\node [style=new style 0] (18) at (-4.665, 0) {};
		\node [style=new style 0] (19) at (2, 4) {};
		\node [style=new style 0] (20) at (2, -4) {};
		\node [style=new style 0] (21) at (-0.222, 0) {};
		\node [style=new style 0] (22) at (2, 0) {};
		\node [style=new style 0] (23) at (4.22, 0) {};
		\node [style=new style 0] (24) at (8.665, 0) {};
		\node [style=none] (26) at (-7.25, -5.5) {$\pp_{1}$};
		\node [style=none] (27) at (1, 3) {};
		\node [style=none] (28) at (1, 3) {};
		\node [style=none] (29) at (6.08, -5.5) {$\pp_{2}$};
		\node [style=none] (31) at (-0.58, 2.5) {$\pp_{3}$};
		\node [style=none] (33) at (-0.58, -1.5) {$\pp_{4}$};
		\node [style=none] (34) at (-0.58, -5.5) {$\pp_{5}$};
		\node [style=none] (35) at (1.64, -1.5) {$\pp_{6}$};
		\node [style=none] (36) at (3.81, -1.5) {$\pp_{7}$};
		\node [style=none] (37) at (-4.415, 0.75) {$\bm{\theta}_{1}$};
		\node [style=none] (38) at (2.25, 4.75) {$\bm{\theta}_{3}$};
		\node [style=none] (39) at (2.25, -3.25) {$\bm{\theta}_{5}$};
		\node [style=none] (40) at (8.915, 0.75) {$\bm{\theta}_{2}$};
		\node [style=none] (41) at (0.028, 0.75) {$\bm{\theta}_{4}$};
		\node [style=none] (42) at (2.25, 0.75) {$\bm{\theta}_{6}$};
		\node [style=none] (43) at (4.47, 0.75) {$\bm{\theta}_{7}$};
		\node [style=none] (44) at (11.5, -5.5) {$\color{black}{\Omega}$};
		\node [style=none] (45) at (-4.75, 5) {$\Pi := \{ \pp_{1}, \dots, \pp_{7} \}$};
	\end{pgfonlayer}
	\begin{pgfonlayer}{edgelayer}
		\draw [style=thick line] (0.center) to (1.center);
		\draw [style=thick line] (1.center) to (2.center);
		\draw [style=thick line] (2.center) to (3.center);
		\draw [style=thick line] (3.center) to (0.center);
		\draw (7.center) to (6.center);
		\draw (9.center) to (8.center);
		\draw (11.center) to (12.center);
		\draw (10.center) to (13.center);
		\draw (14.center) to (16.center);
		\draw (15.center) to (17.center);
	\end{pgfonlayer}
\end{tikzpicture}
    }
    \vspace{1cm}
    \\[\abovecaptionskip]
    \Huge (a) Partitioning definitions
  \end{tabular}
  }
  }
  \raisebox{-0.5\height}{
  \resizebox{0.277\linewidth}{!}{
  \begin{tabular}{@{}c@{}}
    \resizebox{1.05\textwidth}{!}{%
    \centering\begin{tikzpicture}[scale=0.9]
	\begin{pgfonlayer}{nodelayer}
		\node [style=none] (0) at (-8, 6) {};
		\node [style=none] (1) at (-8, -5.5) {};
		\node [style=none] (2) at (8, -5.5) {};
		\node [style=none] (3) at (8, 6) {};
		\node [style=none] (4) at (-9, 0) {};
		\node [style=none] (5) at (8, 0) {};
		\node [style=none] (6) at (0, 6) {};
		\node [style=none] (7) at (0, -5.5) {};
		\node [style=none] (8) at (-4, 0) {};
		\node [style=none] (9) at (-4, -5.5) {};
		\node [style=none] (10) at (4, 6) {};
		\node [style=none] (11) at (4, 0) {};
		\node [style=red] (12) at (2, 3) {};
		\node [style=new style 0] (13) at (6, 3) {};
		\node [style=new style 0] (14) at (4, -3) {};
		\node [style=new style 0] (15) at (-2, -3) {};
		\node [style=red] (16) at (-6, -3) {};
		\node [style=orange] (17) at (-4, 3) {};
		\node [style=none] (18) at (3, 4.25) {};
		\node [style=none] (19) at (-3.25, 2.5) {};
		\node [style=none] (20) at (-7.5, -2) {};
		\node [style=none] (21) at (-0.75, 0.5) {};
		\node [style=none] (22) at (5.25, 5.25) {};
		\node [style=none] (23) at (-4.75, 3.25) {};
		\node [style=none] (24) at (-8.5, -3) {};
		\node [style=none] (25) at (-0.5, -1) {};
		\node [style=none] (26) at (-9, -5.25) {};
		\node [style=none] (27) at (6, 6) {};
		\node [style=none] (28) at (-3.75, 4) {$\color{n_orange}{\bm{\theta}_{1}}$};
		\node [style=none] (29) at (2.25, 4) {$\color{n_red}{\bm{\theta}_{2}}$};
		\node [style=none] (30) at (6.25, 4) {$\bm{\theta}_{3}$};
		\node [style=none] (31) at (4.25, -2) {$\bm{\theta}_{6}$};
		\node [style=none] (32) at (-1.75, -2) {$\bm{\theta}_{5}$};
		\node [style=none] (33) at (-5.75, -2) {$\color{n_red}{\bm{\theta}_{4}}$};
		\node [style=none] (35) at (-7, -5) {$\color{n_purple}{\bm{\Phi}_{\mathrm{linear}}}$};
		\node [style=none] (36) at (-6, 5.4) {$\color{n_red}{H = \{ \bm{\theta}_{2}, \bm{\theta}_{4} \}}$};
	\end{pgfonlayer}
	\begin{pgfonlayer}{edgelayer}
		\draw [style=density 2, in=30, out=135, looseness=0.50] (22.center) to (23.center);
		\draw [style=density 2, in=120, out=-150, looseness=0.50] (23.center) to (24.center);
		\draw [style=density 2, in=-150, out=-60, looseness=0.75] (24.center) to (25.center);
		\draw [style=density 2, in=-45, out=30, looseness=0.50] (25.center) to (22.center);
		\path [draw, style=density 2] (22.center) -- (23.center) -- (24.center) -- (25.center) -- (22.center);
		\draw [style=density 1, in=30, out=135, looseness=0.50] (18.center) to (19.center);
		\draw [style=density 1, in=135, out=-150, looseness=0.50] (19.center) to (20.center);
		\draw [style=density 1, in=-135, out=-45, looseness=0.75] (20.center) to (21.center);
		\draw [style=density 1, in=-45, out=45, looseness=0.50] (21.center) to (18.center);
		\path [draw, style=density 1] (18.center) -- (19.center) -- (20.center) -- (21.center) -- (18.center);
		\draw (4.center) to (5.center);
		\draw (8.center) to (9.center);
		\draw (7.center) to (6.center);
		\draw (10.center) to (11.center);
		\draw [style=linear] (27.center) to (26.center);
	\end{pgfonlayer}
\end{tikzpicture}
    }
    \vspace{1cm}
    \\[\abovecaptionskip]
    \Huge (b) Locally linear structure
  \end{tabular}
  }
  }
  \raisebox{-0.5\height}{
  \resizebox{0.277\linewidth}{!}{
  \begin{tabular}{@{}c@{}}
    \resizebox{1.05\textwidth}{!}{%
    \centering\begin{tikzpicture}[scale=0.9]
	\begin{pgfonlayer}{nodelayer}
		\node [style=none] (0) at (-9, 6) {};
		\node [style=none] (1) at (-9, -6) {};
		\node [style=none] (2) at (8, -6) {};
		\node [style=none] (3) at (8, 6) {};
		\node [style=none] (4) at (-9, 2) {};
		\node [style=none] (5) at (0, 2) {};
		\node [style=none] (6) at (0, 6) {};
		\node [style=none] (7) at (0, -6) {};
		\node [style=none] (8) at (0, -2) {};
		\node [style=none] (9) at (-9, -2) {};
		\node [style=red] (12) at (4, -4) {};
		\node [style=new style 0] (15) at (-4.5, 4) {};
		\node [style=red] (16) at (-4.5, -4) {};
		\node [style=orange] (17) at (4, 0) {};
		\node [style=none] (18) at (7, 2.5) {};
		\node [style=none] (19) at (4.5, -1.25) {};
		\node [style=none] (20) at (-0.25, -4.75) {};
		\node [style=none] (21) at (4.5, -3.5) {};
		\node [style=none] (22) at (7.25, 3.25) {};
		\node [style=none] (23) at (3.75, -1) {};
		\node [style=none] (24) at (-1.5, -4.75) {};
		\node [style=none] (25) at (4.5, -4.75) {};
		\node [style=none] (28) at (4.25, 1) {$\color{n_orange}{\bm{\theta}_{9}}$};
		\node [style=none] (29) at (4.25, -3) {$\color{n_red}{\bm{\theta}_{10}}$};
		\node [style=none] (32) at (-3.25, 4.25) {$\bm{\theta}_{1}$};
		\node [style=none] (33) at (-4.25, -3) {$\color{n_red}{\bm{\theta}_{5}}$};
		\node [style=none] (37) at (0, 2) {};
		\node [style=none] (38) at (0, -2) {};
		\node [style=none] (39) at (8, -2) {};
		\node [style=none] (40) at (8, 2) {};
		\node [style=orange] (41) at (4, 4) {};
		\node [style=none] (42) at (4.25, 5) {$\color{n_orange}{\bm{\theta}_{2}}$};
		\node [style=none] (43) at (-3, 2) {};
		\node [style=none] (44) at (-6, 2) {};
		\node [style=none] (45) at (-6, 6) {};
		\node [style=none] (46) at (-3, 6) {};
		\node [style=none] (47) at (0, -4) {};
		\node [style=none] (48) at (-9, -4) {};
		\node [style=none] (49) at (-6, 2) {};
		\node [style=none] (50) at (-6, -6) {};
		\node [style=none] (51) at (-3, -6) {};
		\node [style=none] (52) at (-3, 2) {};
		\node [style=orange] (53) at (-4.5, 0) {};
		\node [style=none] (54) at (-4.25, 1) {$\color{n_orange}{\bm{\theta}_{7}}$};
		\node [style=orange] (55) at (-1.5, 0) {};
		\node [style=none] (56) at (-1.25, 1) {$\color{n_orange}{\bm{\theta}_{8}}$};
		\node [style=orange] (57) at (-1.5, -4) {};
		\node [style=none] (58) at (-1.25, -3) {$\color{n_orange}{\bm{\theta}_{6}}$};
		\node [style=orange] (59) at (-7.5, 0) {};
		\node [style=none] (60) at (-7.25, 1) {$\color{n_orange}{\bm{\theta}_{3}}$};
		\node [style=orange] (61) at (-7.5, -4) {};
		\node [style=none] (62) at (-7.25, -3) {$\color{n_orange}{\bm{\theta}_{4}}$};
		\node [style=none] (63) at (-4, -0.5) {};
		\node [style=none] (64) at (-5.5, -3.25) {};
		\node [style=none] (65) at (-7.25, -6) {};
		\node [style=none] (66) at (-2.75, -6) {};
		\node [style=none] (67) at (-3.75, 0.5) {};
		\node [style=none] (68) at (-7.25, -2.5) {};
		\node [style=none] (69) at (-8, -6) {};
		\node [style=none] (70) at (-2, -6) {};
		\node [style=none] (71) at (4, 3) {};
		\node [style=none] (72) at (-3, -4) {};

		\node [style=none] (75) at (8, 1.75) {};
		
		\node [style=none] (77) at (-4.5, -0.5) {};
		\node [style=none] (78) at (-8, -4) {};
		\node [style=none] (79) at (-1, -4) {};
		
		\node [style=none] (81) at (-7.5, -6) {};
		\node [style=none] (82) at (-1.5, -6) {};
		\node [style=none] (84) at (-5, 5.25) {$\color{n_red}{H = \{ \bm{\theta}_{5}, \bm{\theta}_{10} \}}$};
		\node [style=none] (85) at (2, 3.5) {$\color{n_turq}{\bm{\Phi}_{\mathrm{balls}}}$};
	\end{pgfonlayer}
	\begin{pgfonlayer}{edgelayer}
		\draw [style=density 2, in=30, out=135, looseness=0.50] (22.center) to (23.center);
		\draw [style=density 2, in=120, out=-150, looseness=0.50] (23.center) to (24.center);
		\draw [style=density 2, in=-150, out=-60, looseness=0.75] (24.center) to (25.center);
		\draw [style=density 2, in=-45, out=30, looseness=0.50] (25.center) to (22.center);
		\path [draw, style=density 2] (22.center) -- (23.center) -- (24.center) -- (25.center) -- (22.center);

		\draw [style=density 2, in=30, out=135, looseness=0.50] (67.center) to (68.center);
		\draw [style=density 2, in=120, out=-150, looseness=0.50] (68.center) to (69.center);
		\draw [style=density 2, in=-180, out=0, looseness=0.50] (69.center) to (70.center);
		\draw [style=density 2, in=-45, out=30, looseness=0.50] (70.center) to (67.center);
		\path [draw, style=density 2] (67.center) -- (68.center) -- (69.center) -- (70.center) -- (67.center);

		\draw [style=density 1, in=30, out=135, looseness=0.50] (18.center) to (19.center);
		\draw [style=density 1, in=135, out=-150, looseness=0.50] (19.center) to (20.center);
		\draw [style=density 1, in=-135, out=-45, looseness=0.75] (20.center) to (21.center);
		\draw [style=density 1, in=-45, out=45, looseness=0.50] (21.center) to (18.center);
		\path [draw, style=density 1] (18.center) -- (19.center) -- (20.center) -- (21.center) -- (18.center);
		
		\draw [style=density 1, in=30, out=135, looseness=0.50] (63.center) to (64.center);
		\draw [style=density 1, in=135, out=-150, looseness=0.50] (64.center) to (65.center);
		\draw [style=density 1, in=-180, out=0, looseness=0.75] (65.center) to (66.center);
		\draw [style=density 1, in=-45, out=45, looseness=0.50] (66.center) to (63.center);
		\path [draw, style=density 1] (63.center) -- (64.center) -- (65.center) -- (66.center) -- (63.center);

		\draw (4.center) to (5.center);
		\draw (8.center) to (9.center);
		\draw (7.center) to (6.center);
		\draw (38.center) to (39.center);
		\draw (37.center) to (40.center);
		\draw (49.center) to (50.center);
		\draw (52.center) to (51.center);

		\draw [style=green trans, bend right=45] (71.center) to (72.center);
		\draw [style=green trans, in=0, out=135, looseness=0.75] (75.center) to (71.center);
		\draw [style=green trans, in=270, out=105] (66.center) to (72.center);
		\path [style=green trans no line] (75.center) -- (71.center) -- (72.center) -- (66.center) -- (2.center) -- (75.center);

		\draw [style=green trans, bend left=45] (78.center) to (77.center);
		\draw [style=green trans, bend left=45] (77.center) to (79.center);
		\draw [style=green trans, in=60, out=-90] (79.center) to (82.center);
		\draw [style=green trans, in=-90, out=120] (81.center) to (78.center);
		\path [style=green trans no line] (78.center) -- (77.center) -- (79.center) -- (82.center) -- (81.center) -- (78.center);

	\end{pgfonlayer}
\end{tikzpicture}
    }
    \vspace{1cm}
    \\[\abovecaptionskip]
    \Huge (c) Local neighbourhoods
  \end{tabular}
  }
  }
  \vspace{-0.05cm}
  \caption{Illustration of partition division.
  (a)~The domain $\Omega$ is divided according to a \emph{partitioning} $\Pi$ composed of partitions $\{\pp_i\}$ with corresponding centroids $\{\bm{\theta}_i\}$.
  (b)~Locally linear structure: High mass partitions $\color{n_red}{H = \{ \bm{\theta}_{2}, \bm{\theta}_{4} \}}$ define a linear subspace $\color{n_purple}{\bm{\Phi}_{\mathrm{linear}}}$ used to find candidates for division (e.g.~$\color{n_orange}{\bm{\theta}_{1}}$).
  (c)~Local neighbourhoods: High mass partitions $\color{n_red}{H = \{ \bm{\theta}_{5}, \bm{\theta}_{10} \}}$ define neighbourhood D-balls $\color{n_turq}{\bm{\Phi}_{\mathrm{balls}}}$ used to find candidates for division (e.g.~$\color{n_orange}{\bm{\theta}_{6}, \bm{\theta}_{9}, \dots}$).
  The density function is illustrated in \color{blue}{blue}.
  }
  \label{fig:partitioning}
  \vspace{-0.2cm}
\end{figure*}
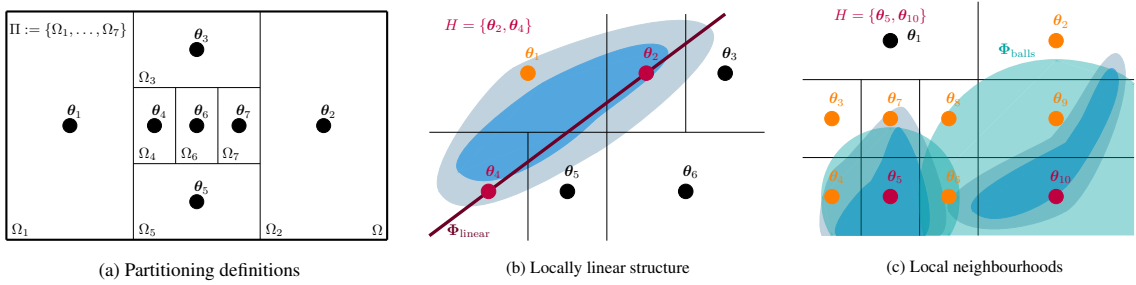

Monte Carlo (MC) methods approximate expectations of functions using a finite collection of samples drawn in proportion to a distribution.
But the samples collected,
for example via Markov Chain Monte Carlo~\cite{geyer1992practical},
only constitute a density function approximation
\footnote{If the associated density values are saved.}
\emph{at the collected samples},
and thus neither provide (probability) mass integral estimates,
nor allow for density queries over the whole sample space $\Omega$ as we require.

One of our method's main requirements is flexibility
and the ability to handle unknown distributions with little to no specification.
This requirement we can fulfil through having the property of asymptotic exactness.
In other words,
that the approximation tends towards the true distribution
and that given enough computational budget, sufficient precision may be achieved.
This is to circumvent the need to specify distribution-specific details upon the usage of the method; which would be difficult,
as the (true) distribution is assumed explicitly unknown and complicated,
in that it may have strong correlations, be discontinuous, multi-modal, or have zero density regions of complex shapes.

To achieve asymptotic exactness, we take inspiration from quadrature,
which is a classic approach to estimating an integral numerically over a bounded domain.
These methods partition the domain and convert the integral into a weighted summation of integrands at each partition.
By creating a partitioning of the full domain, quadrature methods provide asymptotic guarantees by design.
Furthermore, quadrature methods can be robust to the characteristics of the function surface,
such as discontinuities and zero density regions,
as it is the domain partitioning that drives the asymptotic behaviour.
Just as quadrature is asymptotically exact in the limit of decreasing size of the largest partition,
we can similarly achieve this guarantee by ensuring the iterative partitioning algorithm will eventually divide all current partitions.
The downside of the characteristic that provides the asymptotic guarantee is the curse of dimensionality
\cite{gander2000adaptive,Smo63,Gerstner1998NumericalIU,hewitt19_approx_bayes_infer_via_spars},
where the number of partitions in a grid of fixed resolution grows exponentially with the number of dimensions.
We will design the algorithm with this in mind, prioritising regions to refine into finer partitions iteratively.
For a visual depiction of a partitioning, see Figure~\ref{fig:partitions}.
This is especially important as the distribution's typical set may be concentrated to tiny regions,
which would make an equidistant grid computationally infeasible.
We illustrate this in Figure~\ref{fig:naive_methods}.
In the figure, we also demonstrate the common lack of efficiency of using standard rejection sampling for posterior sampling,
a method that otherwise provides much flexibility.
The efficiency problem arises in typical realistic scenarios,
where the likelihood function causes the typical set of the posterior to be concentrated to small regions.
As illustrated, uninformed proposal distributions become infeasibly inefficient already in low dimension (5D in the example) when
the posterior's typical set is even just moderately small.
This is simply a consequence of the vast number of proposals needed on average per proposal landing in the typical set ($\approx10^7$ in the example).

We are not the first to recognise that a tree-based partitioning of the whole domain has several desired properties.
\cite{mccool1997probability} proposed using kd-trees to represent distributions, and
\cite{lu2013multivariate} proposed using partitionings in Bayesian modelling and developed a prior distribution over partitions
and a corresponding inference method.
\cite{li2016density} developed an algorithm to construct piecewise constant density function approximations from iid samples based on discrepancy.
All these methods assume that we have samples drawn from the true distribution and wish to reconstruct the distribution.
As such they address a different problem domain, one where samples are known a-priori.
We do not have access to samples drawn from the distribution,
but are instead provided the unnormalised density function.
In our approach, we evaluate the density function as a part of an active sampling loop,
where at each iteration we pick locations in the sample space for which to evaluate the density function.
We will now proceed with our methodology.

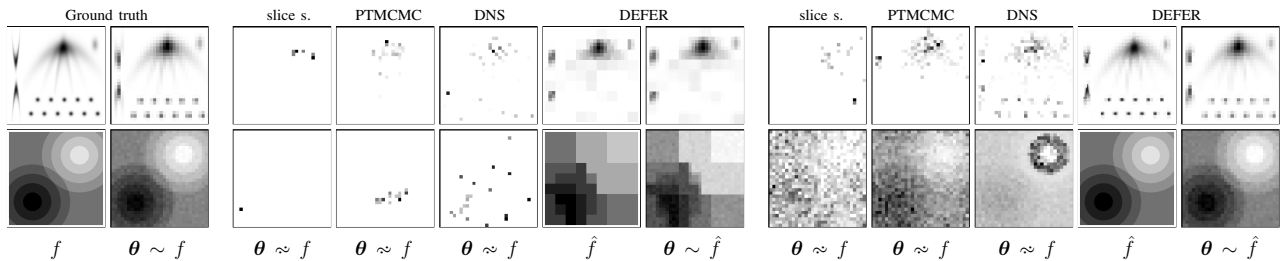
\begin{figure*}[t]
  \resizebox{\textwidth}{!}{
  \begin{tikzpicture}[node distance=0cm]

    \node[draw=black, very thick, inner sep=0pt, minimum size=12pt] (r1c1) {\includegraphics[width=2cm]{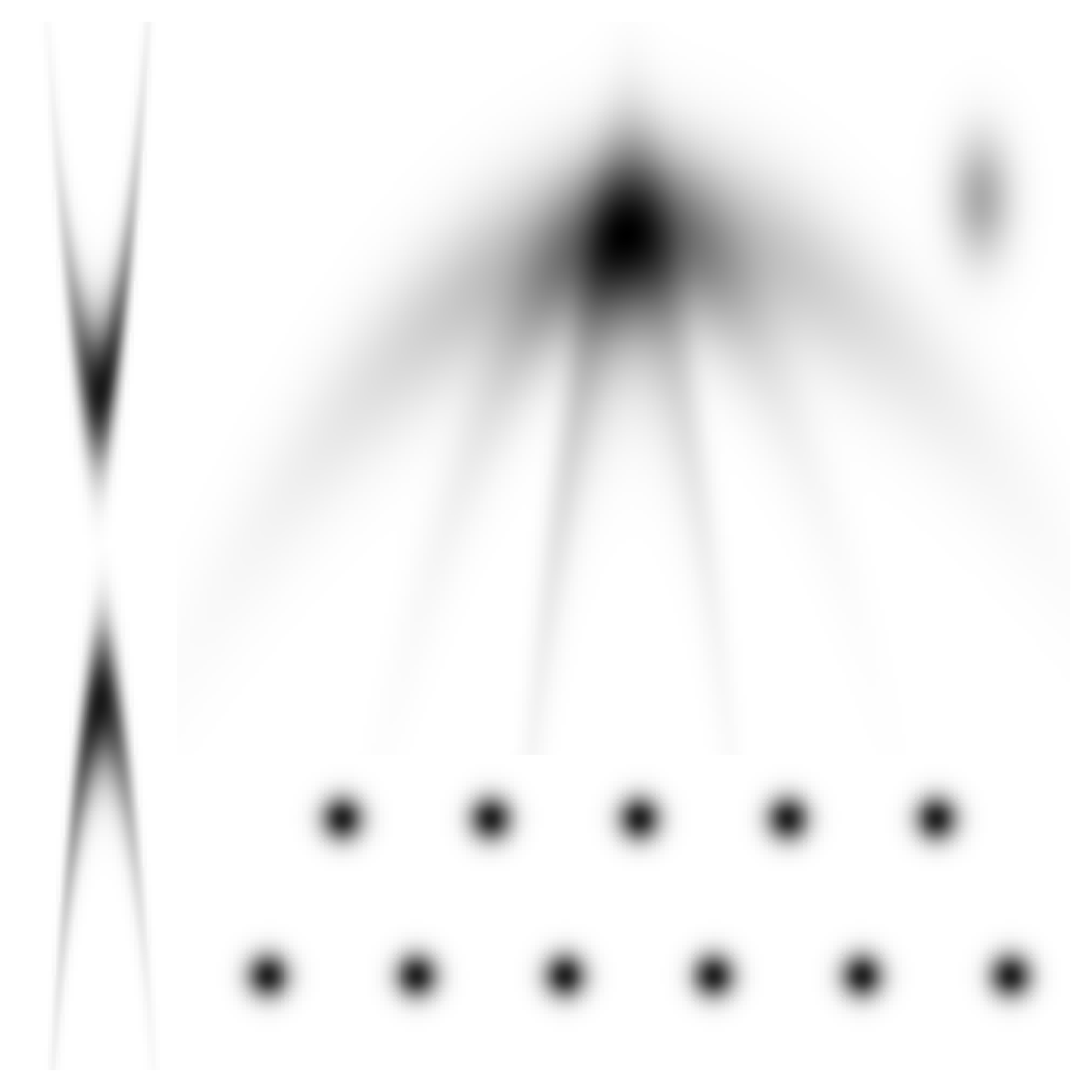}};
    \node[right=of r1c1, xshift=0.1cm, draw=black, very thick, inner sep=0pt, minimum size=12pt] (r1c2) {\includegraphics[width=2cm]{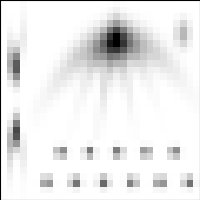}};

    \node[right=of r1c2, xshift=0.5cm, draw=black, very thick, inner sep=0pt, minimum size=12pt] (r1c3) {\includegraphics[width=2cm]{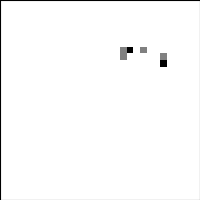}};
    \node[right=of r1c3, xshift=0.1cm, draw=black, very thick, inner sep=0pt, minimum size=12pt] (r1c4) {\includegraphics[width=2cm]{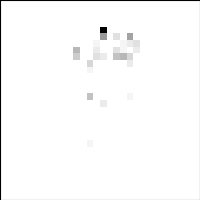}};
    \node[right=of r1c4, xshift=0.1cm, draw=black, very thick, inner sep=0pt, minimum size=12pt] (r1c5) {\includegraphics[width=2cm]{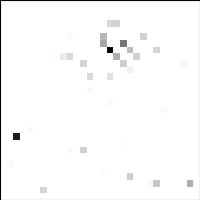}};
    \node[right=of r1c5, xshift=0.1cm, draw=black, very thick, inner sep=0pt, minimum size=12pt] (r1c6) {\includegraphics[width=2cm]{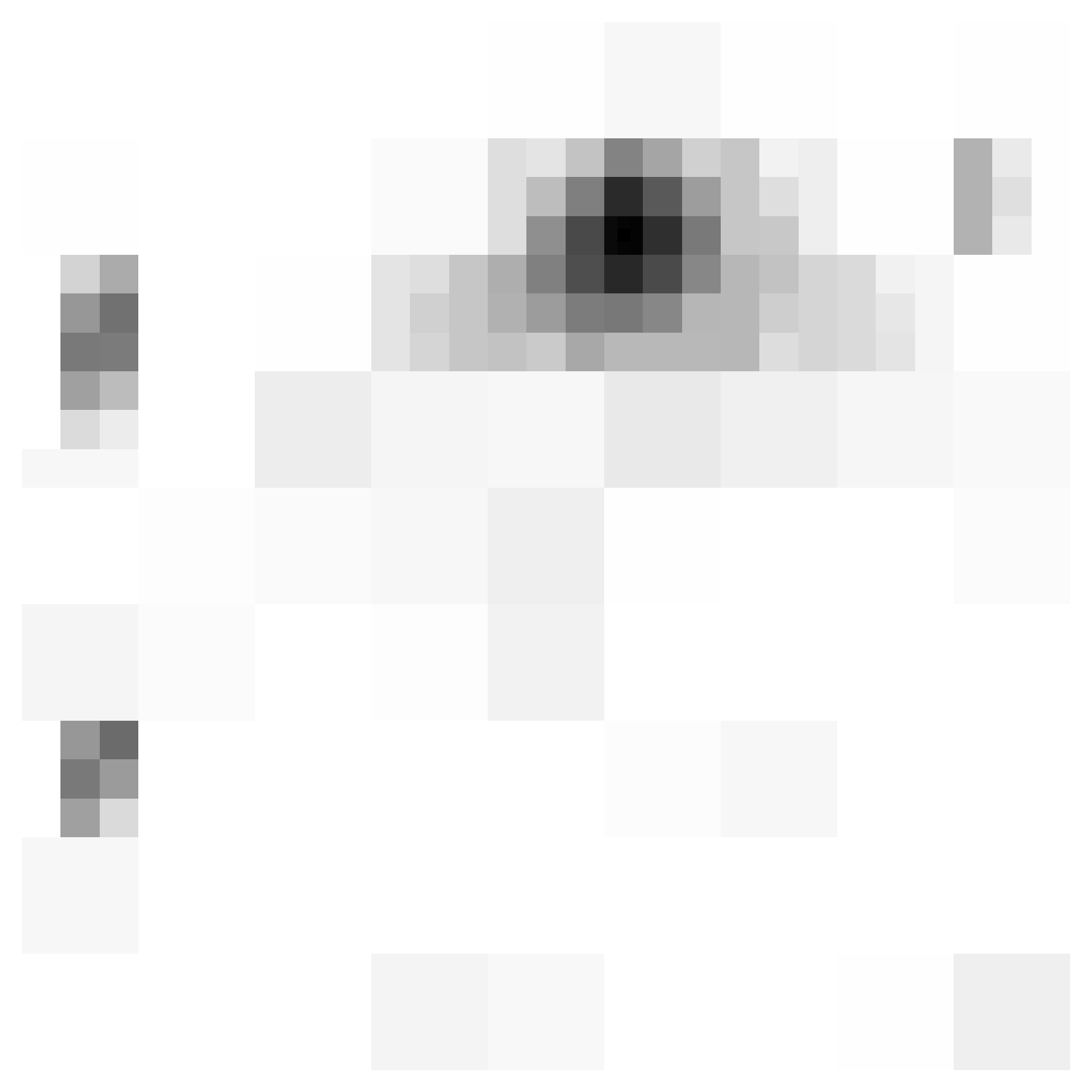}};
    \node[right=of r1c6, xshift=0.1cm, draw=black, very thick, inner sep=0pt, minimum size=12pt] (r1c7) {\includegraphics[width=2cm]{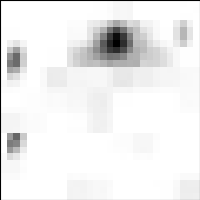}};

    \node[right=of r1c7, xshift=0.5cm, draw=black, very thick, inner sep=0pt, minimum size=12pt] (r1c8) {\includegraphics[width=2cm]{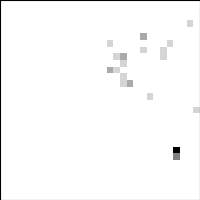}};
    \node[right=of r1c8, xshift=0.1cm, draw=black, very thick, inner sep=0pt, minimum size=12pt] (r1c9) {\includegraphics[width=2cm]{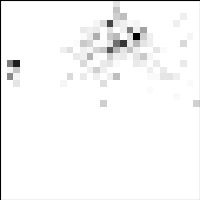}};
    \node[right=of r1c9, xshift=0.1cm, draw=black, very thick, inner sep=0pt, minimum size=12pt] (r1c10) {\includegraphics[width=2cm]{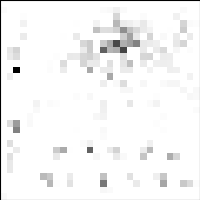}};
    \node[right=of r1c10, xshift=0.1cm, draw=black, very thick, inner sep=0pt, minimum size=12pt] (r1c11) {\includegraphics[width=2cm]{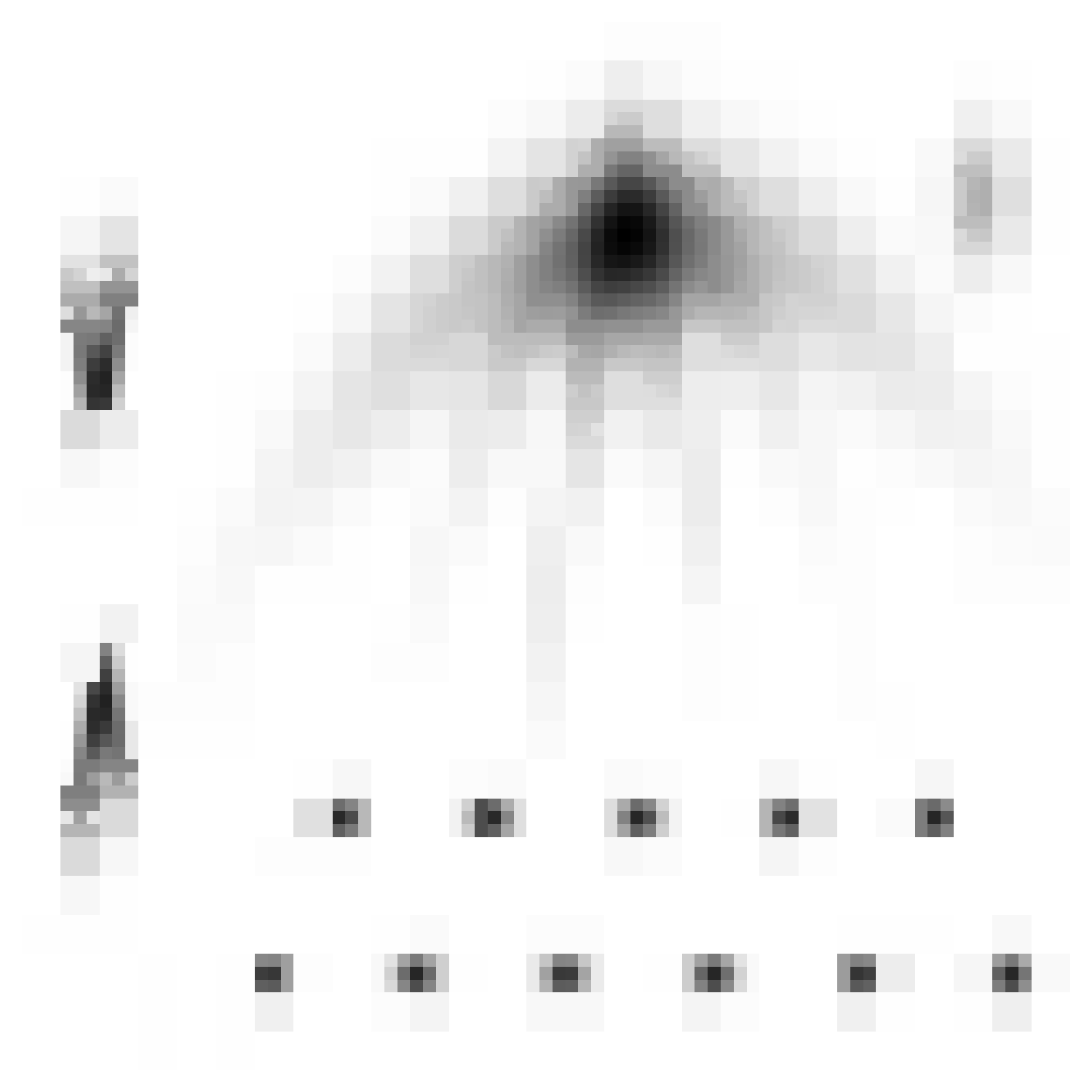}};
    \node[right=of r1c11, xshift=0.1cm, draw=black, very thick, inner sep=0pt, minimum size=12pt] (r1c12) {\includegraphics[width=2cm]{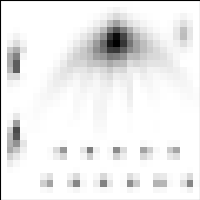}};

    \node[below=of r1c1, yshift=-0.1cm, draw=black, very thick, inner sep=0pt, minimum size=12pt] (r2c1) {\includegraphics[width=2cm]{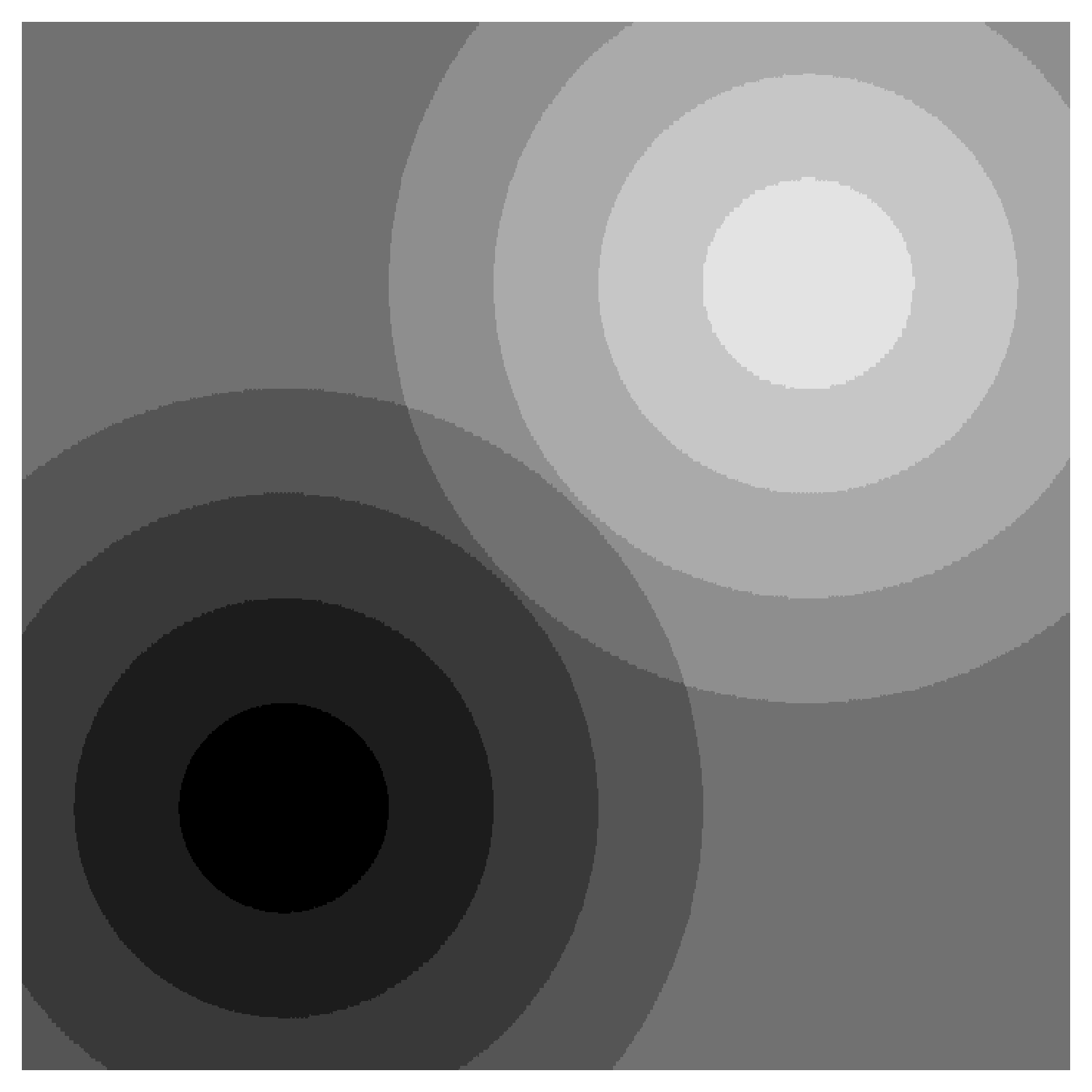}};
    \node[right=of r2c1, xshift=0.1cm, draw=black, very thick, inner sep=0pt, minimum size=12pt] (r2c2) {\includegraphics[width=2cm]{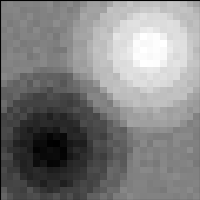}};

    \node[right=of r2c2, xshift=0.5cm, draw=black, very thick, inner sep=0pt, minimum size=12pt] (r2c3) {\includegraphics[width=2cm]{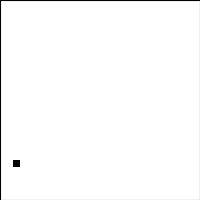}};
    \node[right=of r2c3, xshift=0.1cm, draw=black, very thick, inner sep=0pt, minimum size=12pt] (r2c4) {\includegraphics[width=2cm]{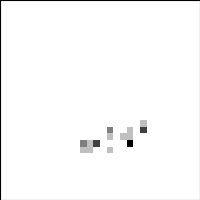}};
    \node[right=of r2c4, xshift=0.1cm, draw=black, very thick, inner sep=0pt, minimum size=12pt] (r2c5) {\includegraphics[width=2cm]{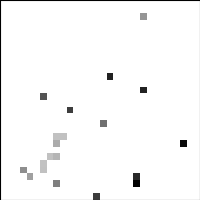}};
    \node[right=of r2c5, xshift=0.1cm, draw=black, very thick, inner sep=0pt, minimum size=12pt] (r2c6) {\includegraphics[width=2cm]{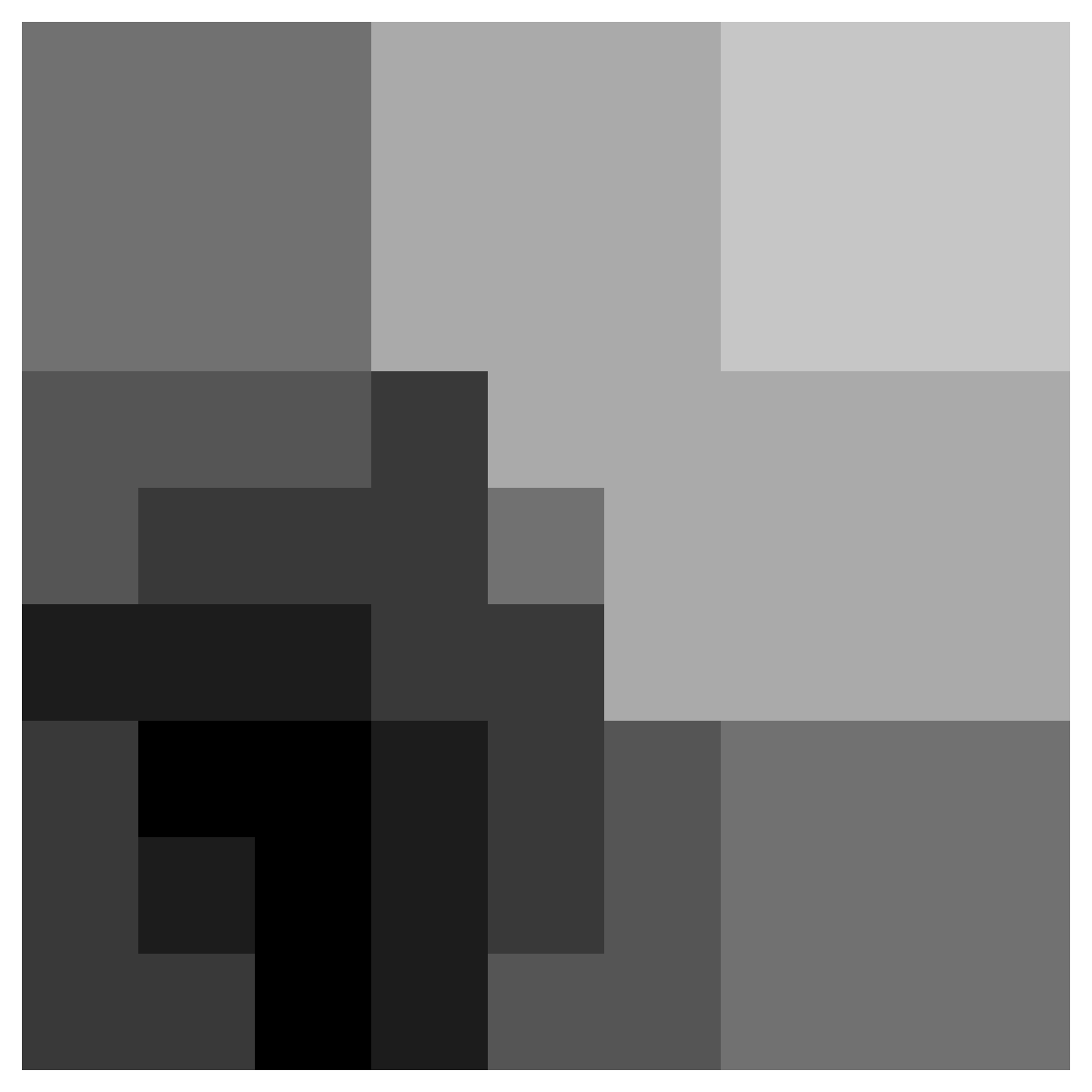}};
    \node[right=of r2c6, xshift=0.1cm, draw=black, very thick, inner sep=0pt, minimum size=12pt] (r2c7) {\includegraphics[width=2cm]{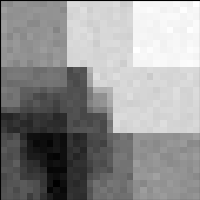}};

    \node[right=of r2c7, xshift=0.5cm, draw=black, very thick, inner sep=0pt, minimum size=12pt] (r2c8) {\includegraphics[width=2cm]{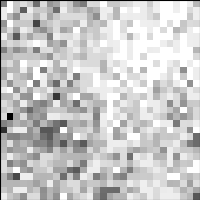}};
    \node[right=of r2c8, xshift=0.1cm, draw=black, very thick, inner sep=0pt, minimum size=12pt] (r2c9) {\includegraphics[width=2cm]{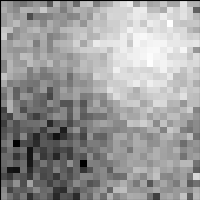}};
    \node[right=of r2c9, xshift=0.1cm, draw=black, very thick, inner sep=0pt, minimum size=12pt] (r2c10) {\includegraphics[width=2cm]{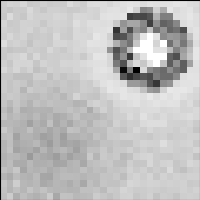}};
    \node[right=of r2c10, xshift=0.1cm, draw=black, very thick, inner sep=0pt, minimum size=12pt] (r2c11) {\includegraphics[width=2cm]{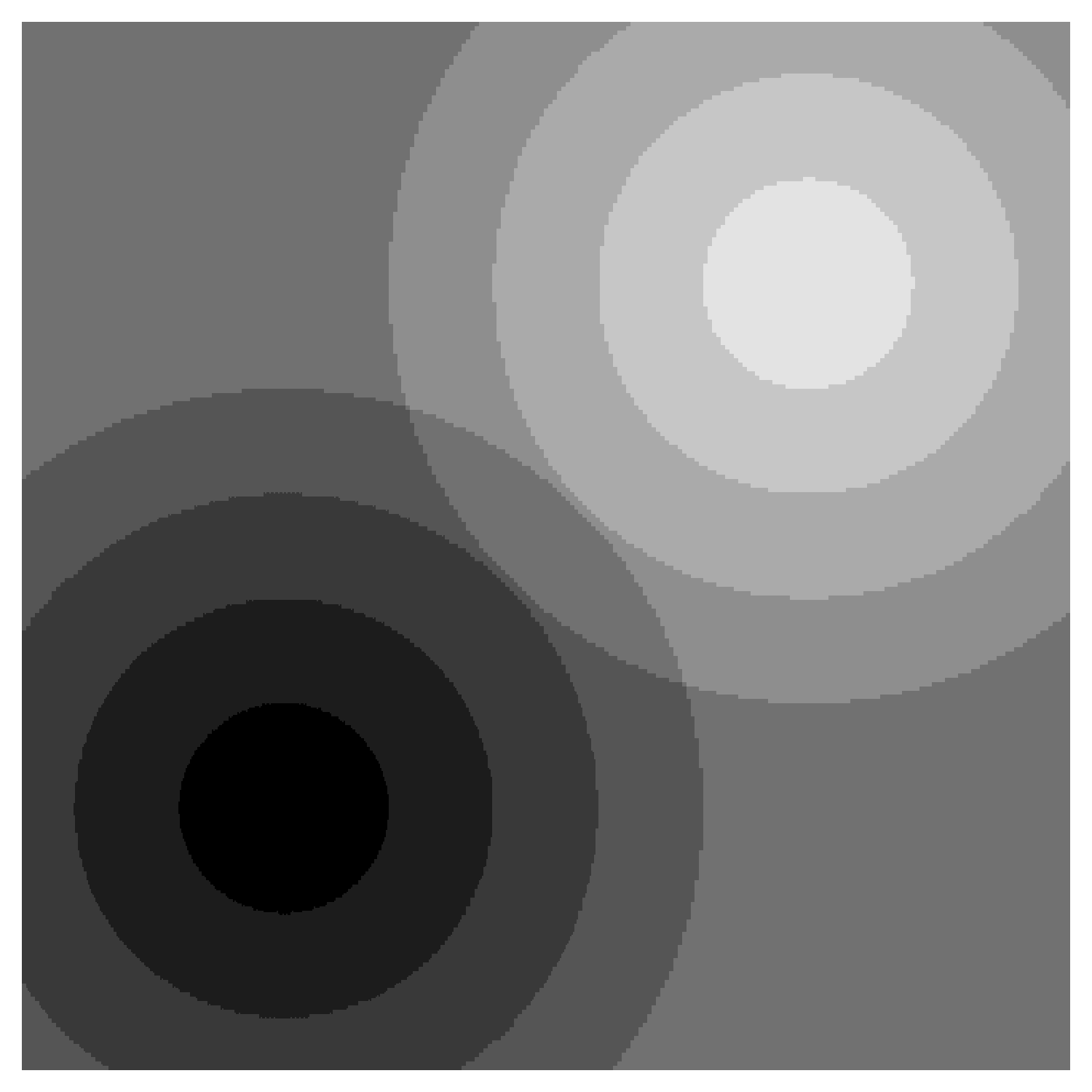}};
    \node[right=of r2c11, xshift=0.1cm, draw=black, very thick, inner sep=0pt, minimum size=12pt] (r2c12) {\includegraphics[width=2cm]{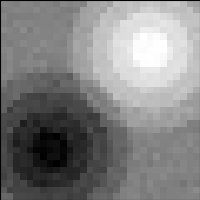}};

    \node[above=of r1c1, xshift=1.05cm, text width=2cm, text centered] {\footnotesize{Ground truth}};
    \node[above=of r1c1, xshift=4.8cm, text width=2cm, text centered] {\footnotesize{slice s.}};
    \node[above=of r1c1, xshift=6.9cm, text width=2cm, text centered] {\footnotesize{PTMCMC}};
    \node[above=of r1c1, xshift=9.0cm, text width=2cm, text centered] {\footnotesize{DNS}};
    \node[above=of r1c1, xshift=12.2cm, text width=2cm, text centered] {\footnotesize{DEFER}};

    \node[above=of r1c1, xshift=15.85cm, text width=2cm, text centered] {\footnotesize{slice s.}};
    \node[above=of r1c1, xshift=17.95cm, text width=2cm, text centered] {\footnotesize{PTMCMC}};
    \node[above=of r1c1, xshift=20.05cm, text width=2cm, text centered] {\footnotesize{DNS}};
    \node[above=of r1c1, xshift=23.25cm, text width=2cm, text centered] {\footnotesize{DEFER}};

    \node[below=of r2c1, yshift=-0.15cm, text width=2cm, text centered] {\large{$f$}};
    \node[below=of r2c1, yshift=-0.15cm, xshift=2.1cm, text width=2cm, text centered] {\large{$\bm{\theta} \sim f$}};

    \node[below=of r2c1, yshift=-0.15cm, xshift=4.7cm, text width=2cm, text centered] {\large{$\bm{\theta} \rsimdots f$}};
    \node[below=of r2c1, yshift=-0.15cm, xshift=6.8cm, text width=2cm, text centered] {\large{$\bm{\theta} \rsimdots f$}};
    \node[below=of r2c1, yshift=-0.15cm, xshift=9.0cm, text width=2cm, text centered] {\large{$\bm{\theta} \rsimdots f$}};
    \node[below=of r2c1, yshift=-0.05cm, xshift=11.1cm, text width=2cm, text centered] {\large{$\hat{f}$}};
    \node[below=of r2c1, yshift=-0.05cm, xshift=13.2cm, text width=2cm, text centered] {\large{$\bm{\theta} \sim \hat{f}$}};

    \node[below=of r2c1, yshift=-0.15cm, xshift=15.85cm, text width=2cm, text centered] {\large{$\bm{\theta} \rsimdots f$}};
    \node[below=of r2c1, yshift=-0.15cm, xshift=17.95cm, text width=2cm, text centered] {\large{$\bm{\theta} \rsimdots f$}};
    \node[below=of r2c1, yshift=-0.15cm, xshift=20.15cm, text width=2cm, text centered] {\large{$\bm{\theta} \rsimdots f$}};
    \node[below=of r2c1, yshift=-0.05cm, xshift=22.25cm, text width=2cm, text centered] {\large{$\hat{f}$}};
    \node[below=of r2c1, yshift=-0.05cm, xshift=24.35cm, text width=2cm, text centered] {\large{$\bm{\theta} \sim \hat{f}$}};

  \end{tikzpicture}
  }
  \vspace{-0.2cm}
  \caption{
  Robustness to characteristics in the density surface.
  In the upper row, a surface with heavy correlations and multi-modality is shown,
  and in the lower row a surface with discontinuities and a valley.
  The output of the methods is shown after approx. 150 (left) and 1k (right) density evaluations for the first function,
  to illustrate the relative inefficiency of MCMC in capturing modes.
  The same is shown after approx. 30 and 100k evaluations for the second function,
  to illustrate the early coarse approximation by DEFER as well as its asymptotic exactness in contrast to DNS.
  As different to the other methods, DEFER outputs a function $\hat{f}$.
  }
  \label{fig:qualitative}
  \vspace{-0.1cm}
\end{figure*}

\section{Methodology}
\label{sec:motivation}

Let $f : \Omega \to \mathbb{R}_{+}$ be an explicitly unknown,
unnormalised density function defined on a hyperrectangular sample space $\Omega \subset \mathbb{R}^D$,
which can be evaluated for any $\bm{\theta} \in \Omega$
\footnote{
In practice, a function with a complicated non-convex domain may be treated
by placing it within a hyperrectangle and assume that $f$ evaluates to $0$
everywhere it is not defined.
Distributions with infinite support may be approximately treated if it is possible to enclose the typical set within the bounds of
the hyperrectangle.
}
.
Our goal is to construct a representation of the function that enables the tractability of downstream tasks,
such as fast, constant-time proportional sampling,
evidence estimation,
tractable expectations of functions,
deriving conditional and marginal densities,
and compute quantities like divergences.
A density function implies a distribution
\begin{equation}
  \mathcal{P}_{f}(\bm{\theta}) =
  \frac{
  f(\bm{\theta})
  }{
  \int_{\bm{\theta} \in \Omega} f(\bm{\theta}) d\bm{\theta}
  }
  =
  \frac{
  f(\bm{\theta})
  }{
  Z
  },
\end{equation}
where $Z$ constitutes the unknown normalising constant.
We will approximate both the function $f$ and the normalising constant $Z$.

Typically $f$ is formed via a joint distribution of data $\mathcal{D}$ and parameters $\bm{\theta}$,
where we refer to $Z$ as the \emph{evidence} and the distribution as the \emph{posterior}.
In other words,
\begin{equation}
  f(\bm{\theta}) := p(\mathcal{D}|\bm{\theta})\mathcal{P}(\bm{\theta}) = Z \mathcal{P}(\bm{\theta}|\mathcal{D}),
\end{equation}
where $\mathcal{P}(\bm{\theta})$ and $p(\mathcal{D}|\bm{\theta})$ denotes the prior distribution and the likelihood function, respectively.

\subsection{Representation of $\hat{f}$}
The combination of two data structures will represent our approximation $\hat{f}$.
Firstly, a non-overlapping partitioning $\Pi := \{ \pp_i \}$ of the domain $\Omega$ in an array,
with an observed value of $f$ at each respective centroid $\bm{\theta}_i$ of the corresponding partition $\pp_i$,
that represents a Riemann sum over the domain
with hyper-rectangular partitions of non-constant side lengths as illustrated in Figure~\ref{fig:partitions} and Figure~\ref{fig:partitioning}.
An estimate of the normalisation constant over the domain or a subdomain is obtained by summation of the
masses $\{V_i f(\bm{\theta}_{i})\}$ of the partitions within, where $V_i$ is the volume of a partition with index $i$.
Secondly, a tree-structure in which these partitions are organised,
forming a search tree over volumetric objects.
Together, these data structures permit integrals to be approximated by summation as in a quadrature rule,
density queries of $\bm{\theta}$ or partitions by tree search,
and constant-time sampling.
Constant-time sampling will be achieved by the following.
First, sample the index of a partition in constant-time using the alias method for categorical distributions~\cite{kronmal1979alias},
and then sample uniformly within the partition
\footnote{Note that constant time sampling is only available after the construction of the approximation,
as the alias method requires linear time pre-processing.
Without such pre-processing, logarithmic time sampling is available by tree-search.
}
.
We will later demonstrate that,
given such a representation,
many quantities and constructs can be estimated within the same framework,
including conditional and marginal distributions.

\subsection{Iterative construction requirements of $\hat{f}$}
We now consider how to obtain a sufficiently close approximation to $f$ in an efficient manner.
We address this in three ways.
Firstly, we prioritise where in the domain $\Omega$ the approximation should be refined.
Secondly, we allow for a large number of partitions.
Lastly, we provide guarantees that the algorithm asymptotically approaches $f$.
In a non-overlapping space partitioning,
the first requirement translates into a decision-problem over which partitions to divide at a given step in sequence.
The second requirement translates into making the decisions in an efficient manner,
with an algorithmic complexity that allows for fast decisions that remain fast also for a large number
of partitions.
And lastly,
for the asymptotic behaviour,
guaranteeing that all partitions will eventually be divided.

\section{Algorithm}
\label{sec:algorithm}

\begin{algorithm}[!ht]
  \caption{DEFER}
  \label{algo:algorithm}
  \begin{algorithmic}
    \STATE {\bfseries Input:} General density function $f$ defined over $\Omega$ with unknown normalisation constant $Z$. \\
    {\bfseries Output:} Approximation $\hat{f}$, $\hat{Z}$, as specified by the produced partitioning $\Pi_T$. \\
    \STATE Initialize t = 1 and initial partitioning $\Pi_1 = \{ \Omega \}$
    \REPEAT[\textit{makes density acquisitions at each iteration}]
    \STATE $ \{ \pp_i \} = \text{to\_divide}\left[\, \Pi_t\right]$
    \STATE divide each partition $\pp_i$, each one resulting in $\{\pp_j\}$ new partitions
    \STATE add all sets of $\{\pp_j\}$ into $\Pi_t$
    \STATE remove the divided partitions $ \{ \pp_i \} $ from $\Pi_t$
    \STATE set $t \rightarrow t + 1$ and update data structures
    \UNTIL{$N_t \geq N_{\text{max}}$}
  \end{algorithmic}
\end{algorithm}

Given the representation and requirements,
we construct the approximation through an iterative refinement procedure of the current tree-structured domain partitioning.
The algorithm starts from the base case of the whole sample space being one partition, with its density evaluated at the centre.
A subset of the existing partitions will be selected at each iteration to be divided further, as specified by a few criteria, which we will address later.
Note that the base case will always be divided as \emph{at least} one partition will always be divided at each iteration.
When a partition is divided, it will result in new partitions according to a division procedure,
where each new partition will receive an associated density by evaluating $f$ at the partition centre,
except for the centre partition which will inherit it from the divided (parent) partition.
The new partitions will subsequently be incorporated into a few data structures (and the divided partition will be excluded), including the tree-structured partitioning, as well as into some data structures to enable the division criteria to be efficiently checked for all partitions in subsequent iterations.
The outline of the algorithm is shown in Algorithm~\ref{algo:algorithm}.
The set of current partitions at a given iteration $t$ in the sequence of decisions is denoted by $\Pi_t$.
We define $N_t := |\Pi_t|$ to be the number of partitions at iteration $t$.
At each iteration, we divide \emph{all} partitions in $\Pi_t$ that meet \emph{any} of three criteria,
to produce an updated set of partitions $\Pi_{t+1}$ (referred to as {\tt to\_divide} in Algorithm~\ref{algo:algorithm}).
Note that only one criterion needs to be met for division of a given partition,
and multiple partitions may be divided at each iteration.

\paragraph{Partition division routine}
We will now describe the division routine, which is carried out on a given partition after it has been decided to be divided.
A partition division entails dividing the partition into three sub-partitions with equal side-lengths for each divided dimension.
The set of dimensions to divide is the set of dimensions of maximum length for the partition,
following normalisation of the domain $\Omega$ to a unit hyper-cube to avoid favouring dimensions spanning a wider value range.
An illustration is shown in Figure~\ref{fig:partitioning} (a),
where the initial partition (the whole domain $\Omega$) was in the first iteration divided
horizontally followed by vertically, and in the second iteration the centre partition was selected to be
divided further, and divided horizontally.
The divide order of the dimensions is the same as the descending order of the highest observed function value,
in common with~\cite{jones1993lipschitzian}.
In other words, after determining all the child partitions centroids and evaluating the density at those locations,
associate each dimension with the highest density value observed along that dimension and rank the dimensions accordingly.
Note that the centroid position do not depend on the dimension divide order even though the partition bounds do,
allowing the function to be evaluated before the forming of the partitions.
The forming of each new partition, with centroid $\bm{\theta}_j$, entails evaluating the true density function
and storing the density value $f(\bm{\theta}_j)$ together with the partition boundaries.
Note that each partition division results in a variable number of new partitions,
depending on the number of dimensions being divided.

\paragraph{Search-tree}
To form the search-tree described in Section~\ref{sec:motivation},
we store each partition $\Omega_i$ together with its associated density function observation
$f(\bm{\theta}_i)$ in a node.
When a partition is divided (see  Algorithm~\ref{algo:algorithm}),
a set of child nodes is created and stored in an array within the parent node.
Each partition ($\Omega_i$) is represented using two arrays;
with the lower and upper bounds for each dimension, respectively.
A query for the unique \emph{leaf} node (and partition) that is associated with a given $\bm{\theta}$ can be performed by traversing the tree from the root.
This is a consequence of the partitions associated with child nodes of each non-leaf node being non-overlapping,
and their union being equal to the partition of their parent.

\subsection{Partition division decision criteria}
\label{sec:division_criteria}

We now detail the three individually sufficient criteria for division, that we denote CR1 to CR3.
For the first criterion, with the aim of robustness to degeneracies in $f$ and
to maintain an informed and efficient exploration,
we will re-interpret~\cite{jones1993lipschitzian}
which proposed an approach to avoiding explicit assumptions of the Lipschitz constant in the context of global optimisation.
In~\cite{jones1993lipschitzian} the domain of the function to be optimised was iteratively split into
finer partitions, where partitions to be divided had the maximum upper bound function value
under \emph{any} possible maximum rate of change (Lipschitz constant) between zero and infinity.
We will translate and adapt this idea to our context of mass-based prioritisation over the whole function.
To ensure the guarantee of the approximation's asymptotic exactness, at least one criterion will need to be eventually fulfilled for \emph{all} partitions, see Section~\ref{sec:background}.
This first sufficient criterion (CR1) will fulfil this.
We will also derive two complementary criteria that exploit typical density functions structures,
motivated for increased sample-efficiency in handling strong correlations.

\subsubsection{Sufficient partition division criterion 1}
\label{algo_section:direct}

One could imagine simply dividing the partition which currently has the largest estimated mass $V_k f(\bm{\theta}_k)$,
where $V_k$ is the volume of the partition $\pp_k$, and $f(\bm{\theta}_k)$ is the associated density value evaluated
at its centroid $\bm{\theta}_k$.
However, this would not consider that the density at the centroid $\bm{\theta}_k$ typically is less representative for the partition the larger the partition is.
The purpose of this criterion will be to prioritise the division of partitions by their \emph{potential for} (true) mass.
An approach that may easily come to mind is to fit a function model to the density values and partition centroids to assess
how the density function may behave outside the evaluated centroids.
However, fitting a function model, as well as taking the function model into account,
comes at a computational cost which often can dominate the cost of evaluating the density function more times.
Moreover, an interpolating function model implies a function with no discontinuities, which we cannot assume in our setting.
We will derive a rule that allows us to consider all partitions and density evaluations simultaneously, while carrying out decisions, in \emph{logarithmic} time with respect to the number of partitions so far.
Furthermore, the rule will handle discontinuities and rapidly changing surfaces and does not require an assumption or prior of the Lipschitz constant.

\paragraph{Criterion}
The criterion (CR1) is the following.
A partition $\pp_k$ will be divided if there exists \emph{any} rate-of-change constant
\footnote{Note that this is not an (explicit) Lipchitz constant assumption, as we simultaneously consider all possible $\bar{K} > 0$.}
$\bar{K} > 0$ such that
\begin{align}
  \label{eq:direct_criterion_1}
  V_k \cdot \left(f(\bm{\theta}_k) + \bar{K}\frac{d_k}{2} \right) \geq V_i \cdot \left( f(\bm{\theta}_i) + \bar{K}\frac{d_i}{2} \right), && \\
  \text{and} \quad V_k \cdot \left(f(\bm{\theta}_k) + \bar{K}\frac{d_k}{2} \right) \geq \beta \frac{\hat{Z}}{N_t + 1},
  \label{eq:direct_criterion_2}
\end{align}
for all $\forall\ i \in [1,N_{t}]$,
where $\hat{Z} = \sum_{k=1}^{N_t} \hat{Z}_k = \sum_{k=1}^{N_t} V_k \cdot f(\bm{\theta}_{k})$ is the normalising constant of the approximation,
$V_k$ is the volume of the partition $\pp_k$,
$\bm{\theta}_k$ is the centroid of the partition,
and $d_k$ is the diameter of the partition $d_k = \text{sup}_{\bm{\theta}_i, \bm{\theta}_j \in \pp_k}||\bm{\theta}_i - \bm{\theta}_j||$.
Here, $\beta$ is a positive parameter specifying what constitutes a non-trivial
amount of \emph{upper bound mass} of a partition in relation to the current estimate of
average mass; we fix $\beta = 1$ for all the experiments in this work.
Note that $\beta$ only controls precision relative to the average partition mass contribution so far,
and can in general be left at default.

The first statement (Equation~\ref{eq:direct_criterion_1}) is true if the partition of index $k$
has an upper bound on its true mass greater or equal to the upper bound of all the partitions
under \emph{any} choices of $\bar{K}>0$.
The second statement (Equation~\ref{eq:direct_criterion_2}) is true if the partition,
under the corresponding choice of $\bar{K}$,
has an upper bound on the true mass that constitutes a non-trivial amount of mass.

\begin{figure}[h!]
  \centering
  \vspace{0.2cm}
  \raisebox{-0.5\height}{\includegraphics[width=0.43\textwidth]{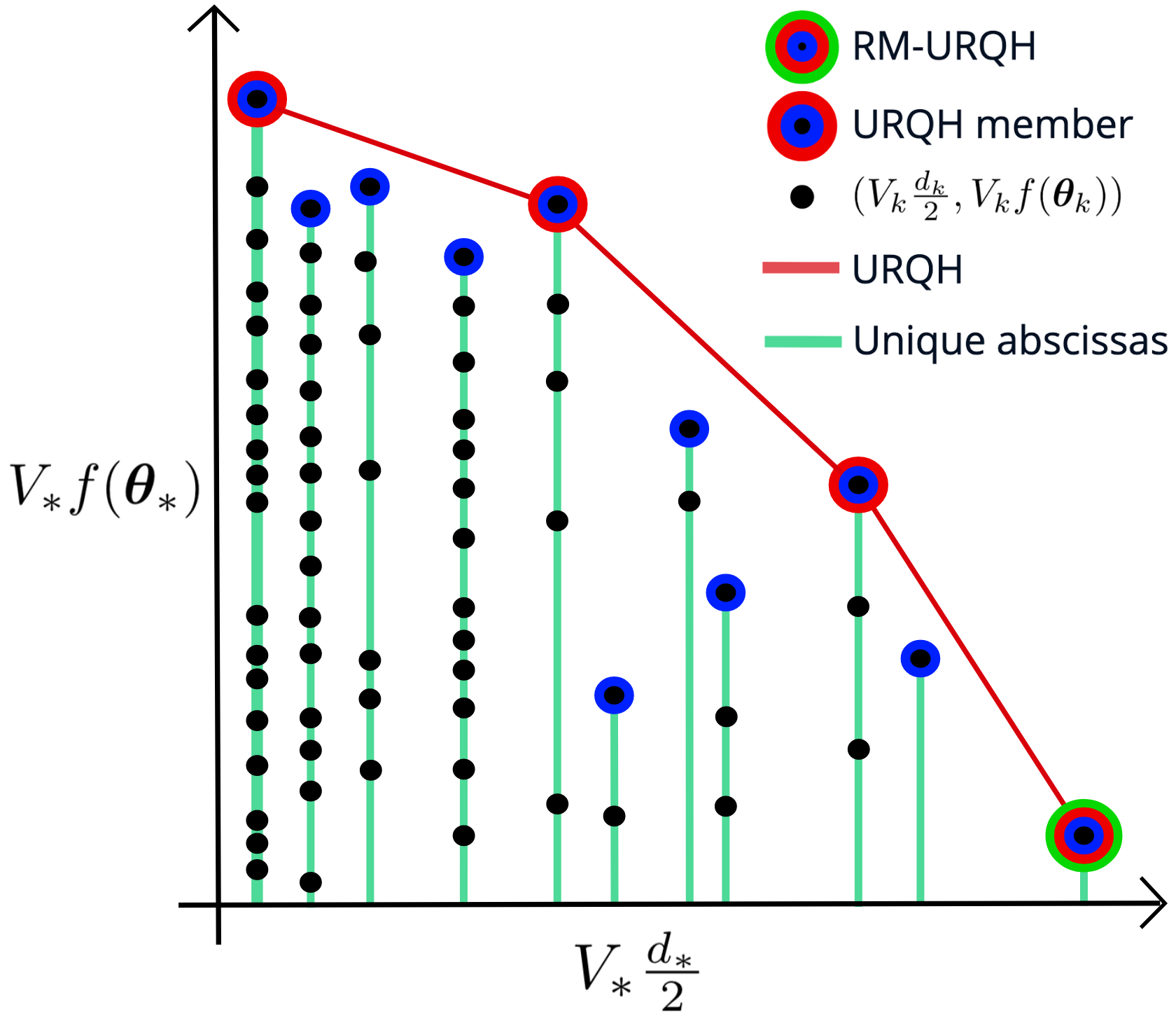}}
  \caption{
  Partitions and their associated density values are mapped to points in a 2D space for which the only partitions that can fulfil CR1 are also a member of the upper right quadrant of the convex hull,
  shown as the red line.
  Moreover, the only partitions whose points needs to be a part of the convex hull calculation are the ones that have the highest ordinate $V_j \cdot f(\bm{\theta}_j)$ (mass) per unique abscissa (green),
  shown with blue rings, which significantly reduces the needed computation.
  }
  \label{fig:convex_hull}
\end{figure}

\paragraph{Implementation}
We check the first statement (Equation~\ref{eq:direct_criterion_1}) using the following.
After construction of a partition $\pp_j$, with parameters $(f(\bm{\theta}_j), \bm{\theta}_j \in \pp_{i})$,
we map it to ordinate $V_j \cdot f(\bm{\theta}_j)$ and abscissa $V_j \cdot \nicefrac{d_j}{2}$ in a 2D space,
see Figure~\ref{fig:convex_hull}.
The criterion will be fulfilled for a partition $\pp_k$ \emph{if and only if}
its corresponding coordinate is a member of the upper-right quadrant of the convex hull (URQH) in this 2D space.

The ordinate and abscissa of each partition is stored in a hash map of heaps,
i.e.~where each entry of a hash table maps to a heap~\cite{williams1964algorithm} as its associated value.
This data structure map is used to efficiently provide access to the \emph{maximum} ordinate partition per unique abscissa at every
iteration $t$, which are kept sorted using the individual heaps, allowing fast updates.
These partitions are the only ones that have corresponding coordinates that \emph{may} be part of the URQH described,
and the only ones that need to be a part of a convex hull calculation.
The number of unique abscissas we will demonstrate is small and near constant with respect to the number of partitions.

The second statement (Equation~\ref{eq:direct_criterion_2}) is checked by considering the \emph{upper bound} of the $\bar{K}$ that a given partition $\pp_k$
used to fulfil the first statement.
This would be the $\bar{K}$ that,
were it any larger,
would result in the right hand neighbour on the URQH having a larger upper bound of the mass
$V_{i + 1} \cdot (f(\bm{\theta}_{i+1}) + \bar{K} \frac{d_{i+1}}{2}) > V_{i} \cdot (f(\bm{\theta}_{i+1}) + \bar{K}\frac{d_{i}}{2})$,
which would violate Equation~\ref{eq:direct_criterion_1}.
In other words, we have $\bar{K}_i^{\text{upper}} := 2\frac{V_{i+1}f(\bm{\theta}_{i+1})  - V_{i}f(\bm{\theta}_{i})}{V_{i}d_{i} - V_{i+1}d_{i+1}}$,
except for the right-most member which will have positively infinite upper-bound on $\bar{K}$.
As such, statement two is fulfilled if $V_i \cdot (f(\bm{\theta}_i) + \bar{K}_i^{\text{upper}}\frac{d_i}{2}) \geq \beta \frac{\hat{Z}}{N_t + 1}$ is true.

\paragraph{Fulfilment of asymptotic guarantee}
As of Equation~\ref{eq:direct_criterion_1} and~\ref{eq:direct_criterion_2}, it is clear that the right-most member of URQH (RM-URQH) will always be divided as its upper bound is positively infinite.
Note that this is true for any (finite) choice of $\beta$.
Furthermore,
as the RM-URQH will always be among the partitions with the largest diameter,
the largest diameter will asymptotically tend to zero.
As a consequence,
all partitions will eventually be divided,
which in turn guarantees asymptotic convergence to $f$ as of a partitioning of the whole domain constitutes a Riemann sum.

\subsubsection{Exploiting characteristics of density functions (CR2 and CR3)}
We now add search behaviour to exploit desirable heuristics targeting both locally linear correlations and spatial proximity in $\Omega$.

\paragraph{Set of high mass partitions} We define $H$ as a set of high (relative) mass partitions at the current iteration.
A partition $\pp_j$ will be a member of this set when
\begin{equation}
  \hat{Z}_j \geq \hat{Z}_{M}
  \quad \textrm{and} \quad
  \hat{Z}_j \geq \alpha \frac{\hat{Z}}{N_t + 1}
  \label{eq:high_mass_partition_definition}
\end{equation}
subject to the constraint that $1 < |H| \leq M$.
The value $\hat{Z}_M$ is defined to be the mass of the partition of the $M^{\text{th}}$ highest mass, $\hat{Z}_* = V_* \cdot f(\bm{\theta}_*) $, and
$\alpha$ is a positive parameter specifying what constitutes a \emph{large} (outlier) mass ratio relative to the average partition.
In practice, we set $M = \text{min}(5, D)$
and fix $\alpha = 20$ for all experiments in this work
\footnote{
This setting we found to work well empirically,
but we did not observe a large sensitivity to the choice of $\alpha$.
We also observed that $\hat{Z}_j$ is typically either similar to the average (as of CR1 aiming to keep true masses roughly equal),
or very much larger than the average.
Thus $\alpha = 20$ is mainly set to not trigger the rule when CR1 already manages to prioritise among the partitions effectively.
}.

\paragraph{CR2: Sufficient partition division criterion 2}
See Figure~\ref{fig:partitioning} for intuition.
Existing partitions with a high estimated mass, relative to other partitions, are grouped in the set
$H$ and denoted with red centroids. Were we to assume (at least some) linear correlation between the centroids in $H$,
we may also assume that high mass is more likely to concentrate along an affine subspace defined by
linear combinations of the centroids denoted $\Phi_{\text{linear}}$, illustrated by the maroon line.
Partitions intersecting this ($|H|-1$)-dimensional hyper-plane are candidates for division.

Given the set $H$ of high mass partitions, we construct a finite set of \emph{representer points} $\RR_{\text{linear}}$ based on the affine subspace constructed from linear combinations of the centroids of the partitions in $H$ that we denote $\Phi_{\text{linear}}$.
We address the representer points in the appendix.
A partition $\pp_k$ will be divided if any point $\rr \in \RR_{\text{linear}}$ falls within the partition.
That is, the set $\{ \rr \mid \rr \in \pp_k, \rr \in \RR_{\text{linear}} \}$ is not empty.

\paragraph{CR3: Sufficient partition division criterion 3}
\label{sec:defer:cr3}

We will now address the neighbourhoods of the $H$ partitions,
illustrated in Figure~\ref{fig:partitioning}.
The figure shows that we consider spatial proximity to elements of $H$ through sets of D-dimensional balls
$\Phi_{\text{balls}}$ centred on elements of $H$. We define $\Phi_{\text{balls}}$ as the union of the interiors of $D$-dimensional balls
centred at respective centroid of the partitions $H$.
The diameter of a given $D$-ball corresponding to partition $\pp_j \in H$ is $\phi \, d_h$,
where $\phi$ is constrained to be $\phi > 1$ to guarantee that positions
\emph{outside} of $\pp_j$ exists in all directions from $\bm{\theta}_j$.
A given $\phi$ leads to a volume ratio $\nu$ of $\pp_j$ relative to the ball as
$\nu = \nicefrac{V_j \cdot \Gamma(\nicefrac{D}{2} + 1)}{(\sqrt{\pi} \frac{\phi d_j}{2})^D}$.
In practice we fix $\phi = 1.2$ in all experiments in this work,
which we found works well empirically with little to no benefit of targeted tuning.

We take the set $H$ of high mass partitions and construct a discrete set $\RR_{\text{balls}}$ of additional representer points based on spatial proximity to the centroids in $H$,
denoted $\Phi_{\text{balls}}$,
and check the condition analogously to CR2.

\paragraph{Summary}
\label{sec:method_summary}
We have now addressed the criteria CR1 to CR3, describing the full algorithm Algorithm~\ref{algo:algorithm}.
The combined time complexity of a step $t$ is $\mathcal{O}(\log N_t + U \log U)$,
where $U$ is the number of \emph{unique} abscissas (see Section~\ref{algo_section:direct}).
For an average number of steps proportional to $N_T$ this results in a total average time complexity of the algorithm as
$\mathcal{O}(N_T(\log N_T + \bar{U} \log \bar{U}))$.
The space complexity (including storage of partitions) of the algorithm is linear,~i.e.~remains proportional to the number of observations.
See the appendix for the analysis.
We will show empirically that $\bar{U}$ is sufficiently small,
and close to constant with respect to $N_T$,
leading to a fast and scalable algorithm.

\begin{figure*}[t]
  \includegraphics[width=1.0\textwidth]{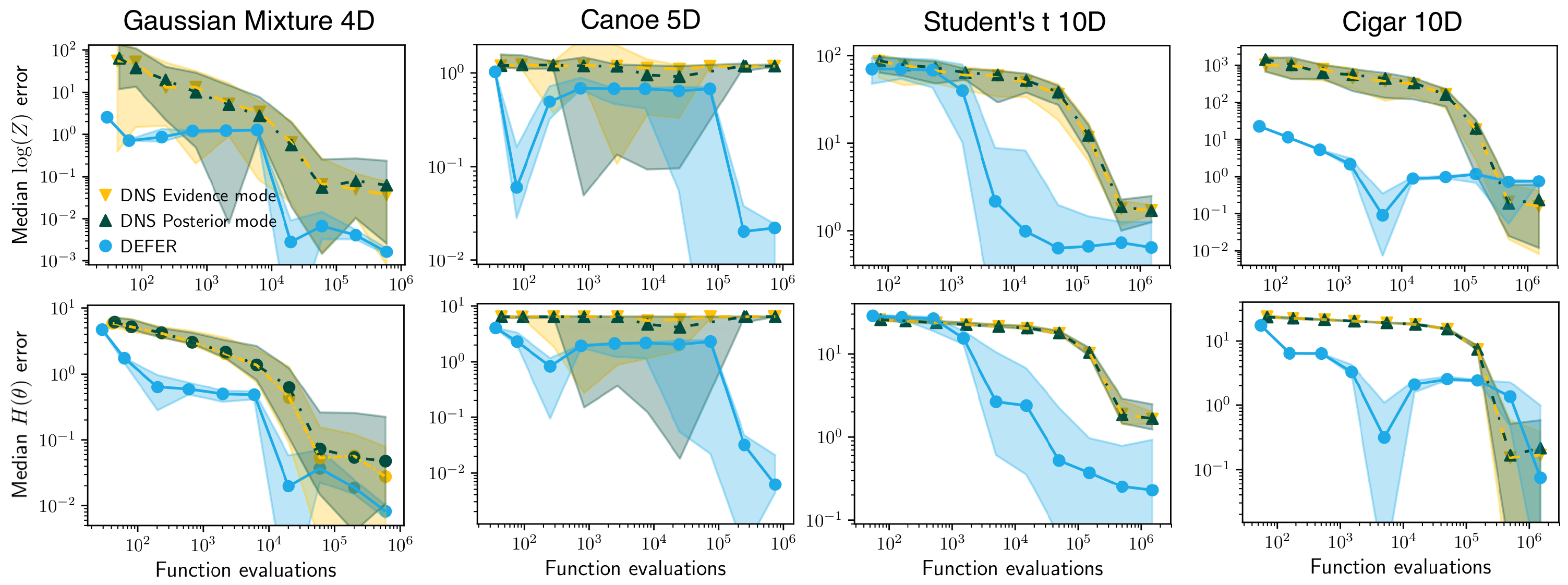}
  \vspace{-0.4cm}
  \caption{
  Upper row: Median absolute error of $\log Z$.
  Lower row: Median absolute error of the entropy $H(\bm{\theta})$.
  For both metrics, shaded areas are the 95\% CI of the median.
  }
  \label{fig:timelines}
\end{figure*}

\begin{figure*}
  \setlength{\tabcolsep}{2pt}
  \renewcommand{\arraystretch}{1.0}
  \begin{minipage}{0.66\textwidth}
    \captionsetup{width=0.8\linewidth}
    \begin{center}
      \scalebox{0.90}{
      \centering
      \begin{tabular}{ccccccccccccccc}

        \includegraphics[width=0.06\textwidth, frame]{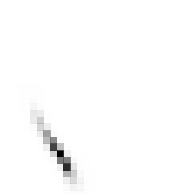} &
        \includegraphics[width=0.06\textwidth, frame]{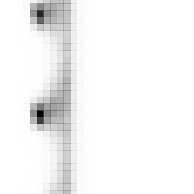} &
        \includegraphics[width=0.06\textwidth, frame]{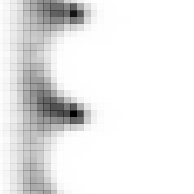} &
        \includegraphics[width=0.06\textwidth, frame]{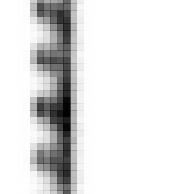} &
        \includegraphics[width=0.06\textwidth, frame]{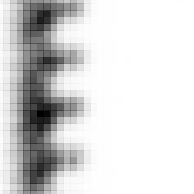} &
        \includegraphics[width=0.06\textwidth, frame]{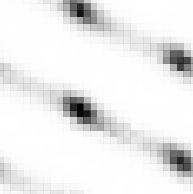} &
        \includegraphics[width=0.06\textwidth, frame]{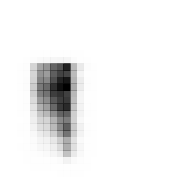} &
        \includegraphics[width=0.06\textwidth, frame]{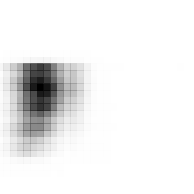} &
        \includegraphics[width=0.06\textwidth, frame]{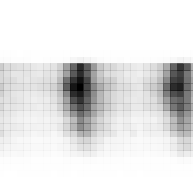} &
        \includegraphics[width=0.06\textwidth, frame]{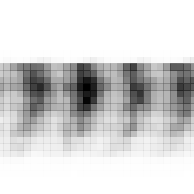} &
        \includegraphics[width=0.06\textwidth, frame]{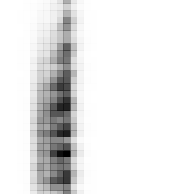} &
        \includegraphics[width=0.06\textwidth, frame]{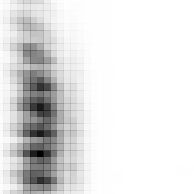} &
        \includegraphics[width=0.06\textwidth, frame]{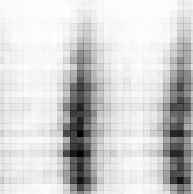} &
        \includegraphics[width=0.06\textwidth, frame]{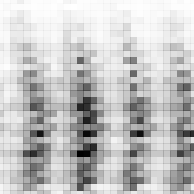} &
        \includegraphics[width=0.06\textwidth, frame]{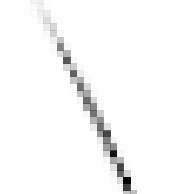}

        \\

        \includegraphics[width=0.06\textwidth, frame]{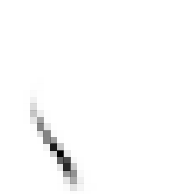} &
        \includegraphics[width=0.06\textwidth, frame]{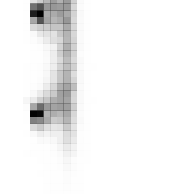} &
        \includegraphics[width=0.06\textwidth, frame]{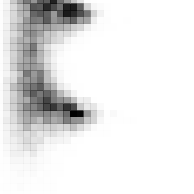} &
        \includegraphics[width=0.06\textwidth, frame]{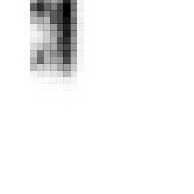} &
        \includegraphics[width=0.06\textwidth, frame]{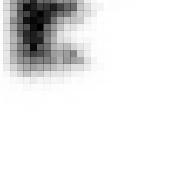} &
        \includegraphics[width=0.06\textwidth, frame]{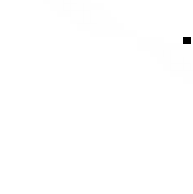} &
        \includegraphics[width=0.06\textwidth, frame]{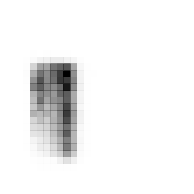} &
        \includegraphics[width=0.06\textwidth, frame]{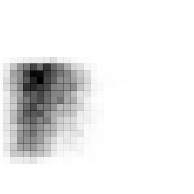} &
        \includegraphics[width=0.06\textwidth, frame]{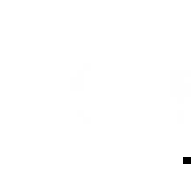} &
        \includegraphics[width=0.06\textwidth, frame]{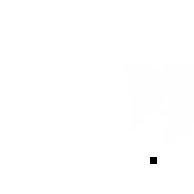} &
        \includegraphics[width=0.06\textwidth, frame]{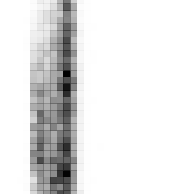} &
        \includegraphics[width=0.06\textwidth, frame]{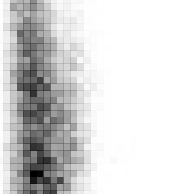} &
        \includegraphics[width=0.06\textwidth, frame]{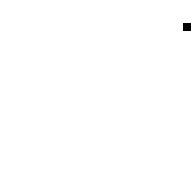} &
        \includegraphics[width=0.06\textwidth, frame]{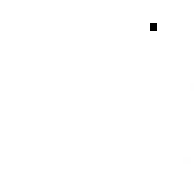} &
        \includegraphics[width=0.06\textwidth, frame]{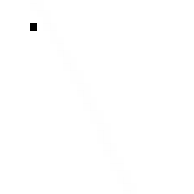}

        \\

        \includegraphics[width=0.06\textwidth, frame]{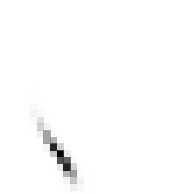} &
        \includegraphics[width=0.06\textwidth, frame]{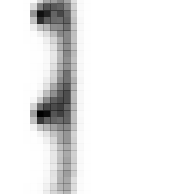} &
        \includegraphics[width=0.06\textwidth, frame]{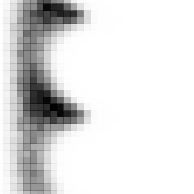} &
        \includegraphics[width=0.06\textwidth, frame]{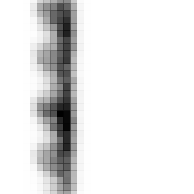} &
        \includegraphics[width=0.06\textwidth, frame]{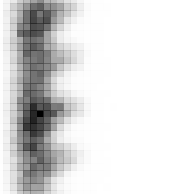} &
        \includegraphics[width=0.06\textwidth, frame]{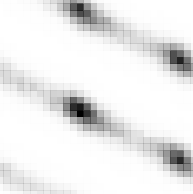} &
        \includegraphics[width=0.06\textwidth, frame]{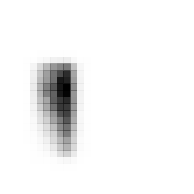} &
        \includegraphics[width=0.06\textwidth, frame]{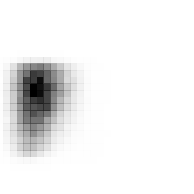} &
        \includegraphics[width=0.06\textwidth, frame]{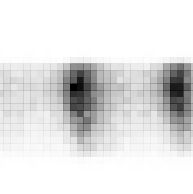} &
        \includegraphics[width=0.06\textwidth, frame]{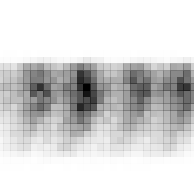} &
        \includegraphics[width=0.06\textwidth, frame]{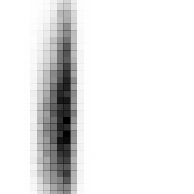} &
        \includegraphics[width=0.06\textwidth, frame]{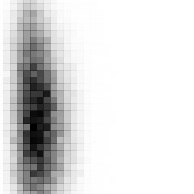} &
        \includegraphics[width=0.06\textwidth, frame]{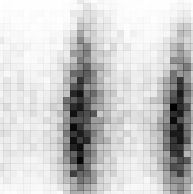} &
        \includegraphics[width=0.06\textwidth, frame]{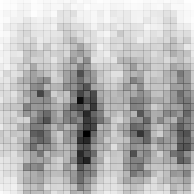} &
        \includegraphics[width=0.06\textwidth, frame]{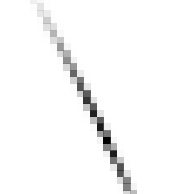}

      \end{tabular}
      }
    \end{center}
    \vspace{-0.35cm}
    \caption{
    Parameter estimation for gravitational-wave physics.
    Shown are histograms of the two-dimensional marginals of a 6D parameter inference problem~\cite{ashton2019bilby},
    with samples produced after 5M density function evaluations.
    Upper row: from the DEFER approximation.
    Second row: using PTMCMC.
    Bottom row: using DNS.
    }
    \label{fig:gw_physics}
  \end{minipage}
  \hspace{0.5cm}
  \begin{minipage}{.30\textwidth}
    \centering
    \captionsetup{width=0.8\linewidth}
    \includegraphics[width=0.90\linewidth]{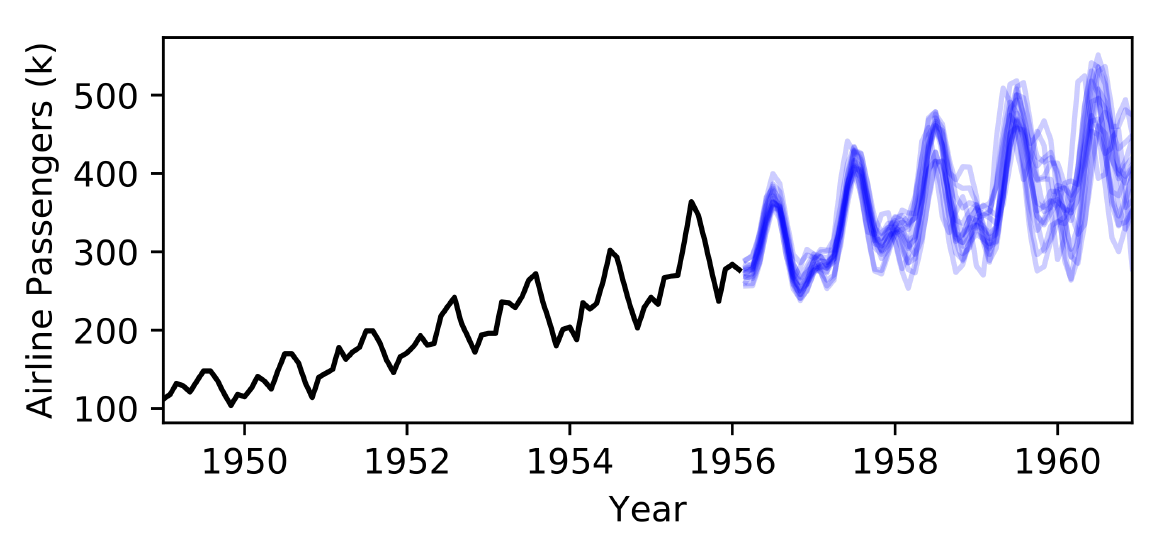}
    \vspace{-0.10cm}
    \captionof{figure}{
    Time-series forecasting using Gaussian Processes with the Spectral Mixture kernel.
    The data is shown in black, and posterior GP samples in blue.
    }
    \label{fig:spectral}
  \end{minipage}
  \vspace{-0.25cm}
\end{figure*}

\section{Experiments}

As discussed in Section~\ref{sec:introduction},
the focus of this work is to produce approximations to density functions with support over the whole sample space,
which in turn can be used for a number of down-stream tasks.
However, it is crucial that the presented algorithm produces approximations that have good quality whilst
being computationally efficient.
We will begin this section by assessing this using comparison to modern approximate inference methods,
and we will later provide examples of the larger set of down-stream tasks we may use the new method for.
For an ablation study of criterion (CR) 1 to 3,
and further setup or problem details, see the appendix.

We choose dynamic nested sampling (DNS)~\cite{higson2019dynamic}, slice sampling~\cite{neal2003slice},
and parallel-tempering MCMC (PTMCMC)~\cite{justin_ellis_2017_1037579} as the baselines
because they are easy-to-use or able to handle multi-modality and work in the absence of gradient information.
As of the capability of DNS to also estimate evidence,
and to be robust to multi-modal and degenerate posteriors with relatively little tuning,
it has the largest overlap in features with DEFER whilst being a recent and strong baseline.
We will therefore make more elaborate comparisons with this method.
Note, however, that DNS does not support proxy density queries, nor other still missing features to be addressed later.
PTMCMC and slice sampling neither support mass integration nor proxy density queries (see Section~\ref{sec:background}),
but is compared with when applicable for reference.

\subsection{Approximation quality and sample-efficiency}

We first test the robustness of the inference methods against challenging properties (see Figure~\ref{fig:qualitative})
including multi-modality, strong correlations, and discontinuities,
on low dimensional examples that are easy to visualize.
We illustrate the asymptotic guarantees of DEFER
in the presence of such properties in the density surface.
We also note its sample-efficiency in representing all modes when compared to MCMC,
and in capturing detail when compared to both DNS and the MCMC methods.
In one of the shown examples DNS fail to recover the ground truth,
which may be related to the algorithm's determination of likelihood contours~\cite{higson2019dynamic}.

We compare with DNS quantitatively both for evidence and parameter estimation.
For evidence estimation, we measure the median absolute error in log evidence (or $\log Z$)
over 20 runs with various budgets.
For parameter estimation, we measure the error in estimated differential entropy,
due to lack of summary statistics for multi-modal distributions.
Results for a few functions with various challenging characteristics
are shown in Figure~\ref{fig:timelines},
such as very small or elongated typical sets
(see appendix for the functions).
We remind the reader that although the surface of the approximation always tends towards the true surface,
integral estimates can oscillate slightly as local overestimations and underestimations can cancel out to some degree.
DEFER is able to match and typically significantly surpass
the performance of DNS.
Especially in getting close to the solution already at low $N_T$ budgets,
sometimes needing order of magnitudes
fewer function evaluations, but it also performs better using the higher $N_T$ budgets.
The gaps between the methods are noticeably larger for estimating differential entropy.
Estimation of entropy is more sensitive to the matching of the shape of a distribution, and not only its integral.
On Canoe,
DNS completely misses the concentrated mass,
but DEFER finds it after 100k evaluations,
which leads to a sharp drop in the error of both evidence and differential entropy.

\subsection{Real-world density sufaces}

Synthetic distributions like the ones addressed are convenient to use for assessment as they have known properties.
However, it is important to confirm that DEFER can be applied to real-world distributions and density functions.
To do this, we will apply DEFER on a parameter estimation task from gravitional-wave (GW) physics research~\cite{abbott2019gwtc, ligo2020gw190412},
and a hyperparameter inference task from~\cite{wilson2013gaussian}.

In GW research physically motivated likelihood functions and priors~\cite{PhysRevD.101.103004} are used,
often without gradients available, 
and the induced density surfaces are typically complicated, multi-modal, or discontinuous.
Inference on these problems is often prohibitively slow,
warranting actions such as re-sampling, density re-weighting and local density integration.
DEFER outputs a density function approximation with support for all these tasks 
making use of the domain-indexed search tree over partitions.
We apply DEFER to a simulated signal example from~\cite{ashton2019bilby}.
Figure~\ref{fig:gw_physics} shows all the 2D marginals of a 6D problem using the
`IMRPhenomPv2' waveform approximant.
Inferred parameters are, for example, the luminosity distance, and the spin magnitudes of binary black-holes.
We note the complicated interactions between parameters,
showing the importance of handling multi-modality and strong correlations.
Importantly, DEFER is able to handle the surface well without any tuning parameters.
As PTMCMC, DEFER asymptotically approaches the unknown ground truth surface.
See the appendix for additional GW experiments.

We apply DEFER to a Gaussian Process time-series regression model with a spectral mixture kernel (SMK)~\cite{wilson2013gaussian}
to infer the posterior of hyperparameters of the kernel,
which is known to be heavily multi-modal and complicated.
With a budget of 50k function evaluations, the negative log-likelihoods on the test data are $377.66$, $365.97$, $236.89$ and $\bm{205.50}$,
for slice sampling, PTMCMC, DNS and DEFER, respectively.
For predictions using DEFER, see Figure~\ref{fig:spectral}.
See appendix for further details and plots.

\begin{figure}
  \includegraphics[width=0.99\linewidth]{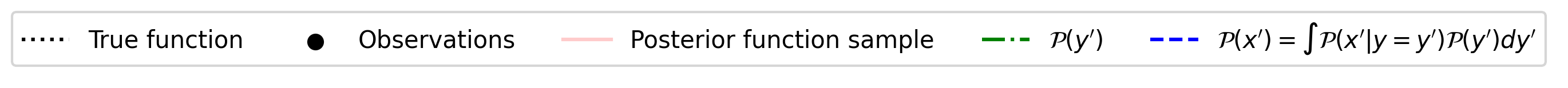}\vspace{-0.2cm}\\
  \includegraphics[width=0.99\linewidth]{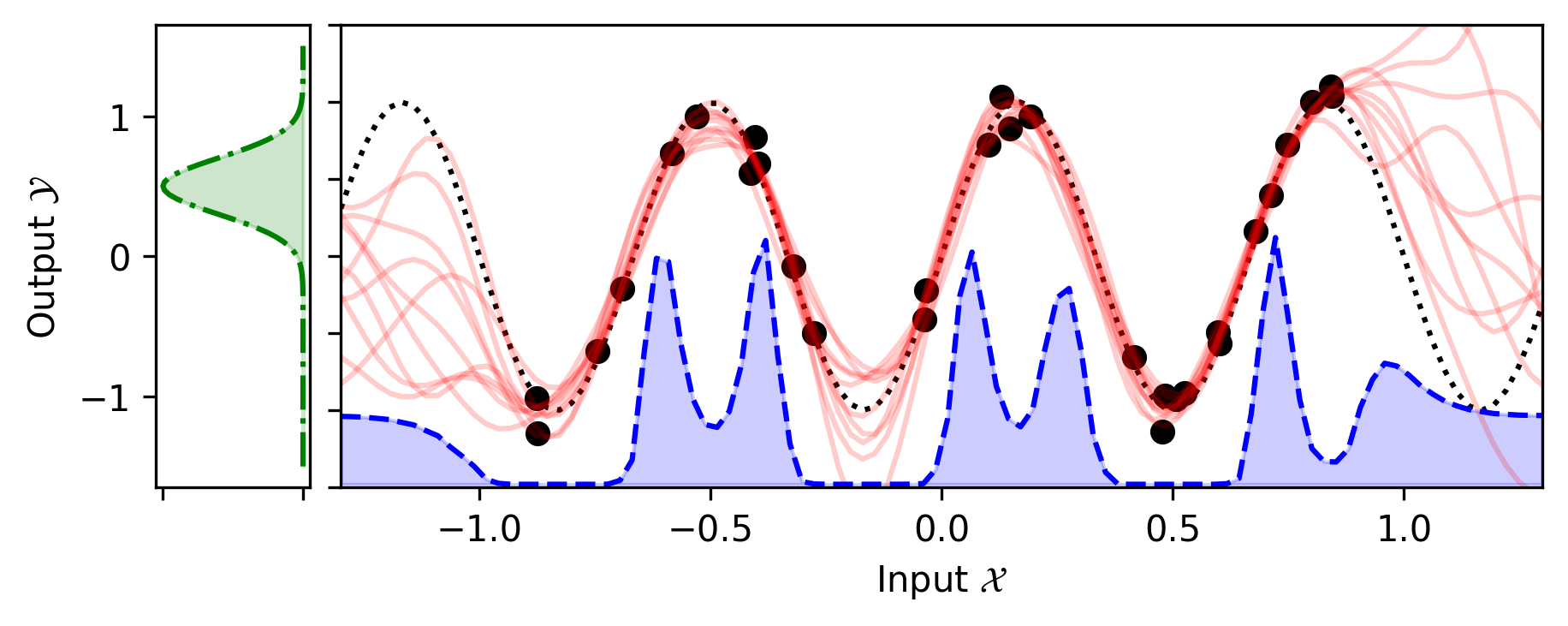}
  \vspace{-0.30cm}
  \caption{
  Input distributions.
  Shown in pink are function ($f$) samples from posterior Gaussian Processes,
  with hyperparameters $\bm{\theta}$ ancestrally sampled from the posterior $\mathcal{P}(\bm{\theta}|\mathcal{D})$.
  $\mathcal{P}(x')$ represent the distribution of the input which generated $y'$,
  where $y' := f(x')$ and $y' \sim \mathcal{P}(y')$.
  DEFER makes it possible to derive an approximation to this input distribution by a combination of capabilities.
  The free-form approximation produced is flexible enough to \emph{represent} the multi-modal input distribution.
  Furthermore, we are able to estimate the intractable integral $\mathcal{P}(x') \propto \int \mathcal{P}(y=y'|x) \mathcal{P}(x) \mathcal{P}(y') dx dy'$
  for each $x' \in \mathcal{X}$. $\mathcal{P}(x)$ is here set to be uniform over the range of the figure.
  }
  \label{fig:input_dist}
  \vspace{-0.30cm}
\end{figure}

\subsection{Runtime performance}
\label{sec:defer_runtime}

In the appendix we confirm both that DEFER has a similar algorithmic cost to the other
methods and that the DEFER has a near-constant cost per function evaluation with respect to $N_T$.
With algorithmic cost, we refer to the cost per 'decision' of where to evaluate the density function,
which is computed from the total (wall-clock) time of inference minus the total function evaluation time,
divided by the number of function evaluations made.
In practice, using our implementation of DEFER and setup, the decision time per evaluation were around 0.3 to 0.5 milliseconds
also after millions of density evaluations.

\subsection{Down-stream task and application examples}

We are now finished with the empirical comparisons,
confirming the quality of the resulting approximations and the efficiency of DEFER to construct them.
Importantly, we have treated density functions in a way that lets us be agnostic to the problem giving rise to the density function.
The approximation produced is a tree from which we can produce a piecewise constant function defined over
the domain or any axis-aligned hyper-rectangular subdomain.
This way we can transform a myriad of problems involving intractable integrals,
such as density marginalisations,
or deriving conditional distributions,
into queries of the approximation and summation.

For example, consider the problem of estimating mutual information
$I(\bm{\theta}_{a}; \bm{\theta}_{b}) = \mathbb{E}_{\mathcal{P}(\bm{\theta}_{a},\bm{\theta}_{b})}[\log \frac{\mathcal{P}(\bm{\theta}_{a},\bm{\theta}_{b})}{\mathcal{P}(\bm{\theta}_{a})\mathcal{P}(\bm{\theta}_{b})}]$.
Normally this would require a specialised algorithm, but we may now instead
estimate it by using DEFER to approximate the joint $\mathcal{P}(\bm{\theta}_{a},\bm{\theta}_{b})$,
and then use the capabilities of the approximation.
Another example is to propagate uncertainty,
such as deriving the distribution of the input to an uncertain function
producing an uncertain output, see Figure~\ref{fig:input_dist}.
We may also apply DEFER to multiple distributions,
allowing us to estimate divergences.

We provide code and examples for such applications and more at
\href{https://github.com/bodin-e/defer}{https://github.com/bodin-e/defer}.

\section{Conclusion}
We have presented a new approach to general Bayesian computation and approximate inference based on recursive partitioning.
The approach is shown to be competitive to state-of-the-art methods on black-box density functions,
with high sample-efficiency and scalability,
allowing complicated surfaces to be estimated with high precision.
The algorithm produces a function representation that can be used for a diverse set of tasks beyond the computation of evidence and sampling,
and is flexible and easy to use.

\vspace{0.41cm}
\noindent\textbf{Acknowledgments}~~
We thank Virginia D'Emilio, Rhys Green and Vivien Raymond (Cardiff University) for input and useful discussions relating to gravitational-wave physics,
Ivan \mbox{Ustyuzhaninov} (University of T\"ubingen) for useful input relating to the representer points,
and Ieva Kazlauskaite, Vidhi Lalchand and Pierre Thodoroff (\mbox{University} of Cambridge) for comments that greatly improved the manuscript.
Supported by the Engineering and Physical Sciences Research Council,
The Royal Swedish Academy of Engineering Sciences, UKRI CAMERA project (EP/M023281/1, EP/T022523/1) and the Royal Society.

\bibliography{main}
\bibliographystyle{icml2021}

\appendix

\section{Down-stream tasks and applications}

\begin{itemize}
  \item
  Analytical expectations of functions with respect to the approximation, $\mathbb{E}_{\hat{\mathcal{P}}(\bm{\theta})}[g(\bm{\theta})]$,
  where $\hat{\mathcal{P}}(\bm{\theta}) \propto \hat{f}(\bm{\theta})$, and $g$ is an arbitrary function of $\bm{\theta}$.
  \item
  Using $\hat{f}(\bm{\theta})$ as a proxy, allowing density queries without resorting to $f$,
  and constant-time re-sampling.
  \item
  Conditional approximations $\hat{\mathcal{P}}(\bm{\theta}_{a}|\bm{\theta}_{b})$, where $\bm{\theta} = \{ \bm{\theta}_{a}, \bm{\theta}_{b} \}$.
  without requiring more evaluations of $f$,
  \item
  Marginal approximations $\hat{\mathcal{P}}(\bm{\theta}_{a}) = \int \mathcal{P}(\bm{\theta}_{a}, \bm{\theta}_{b}) d\bm{\theta}_{b}$,
  where $\mathcal{P}(\bm{\theta}_{a})$ may optionally be substituted with $\hat{\mathcal{P}}(\bm{\theta}_{a}, \bm{\theta}_{b})$.
  Note that this constitute an integral density estimation problem for every $\bm{\theta}_{a} \in \Omega_a$,
  and where DEFER chooses a finite collection of points in a decision loop
  and constructs a tree akin to the other problems.
  For the inner loop of estimating each integral $\int \mathcal{P}(\bm{\theta}_{a}=\bm{\theta}_{a}, \bm{\theta}_{b}) d\bm{\theta}_{b}$
  we may use an arbitrary integration method, including DEFER.
  \item Marginalisation also through arbitrary conditionals,
  such as
  $\int \mathcal{P}(\bm{\theta}_{a}|c(\bm{\theta}_{b})) \mathcal{P}(\bm{\theta}_{b}) d\bm{\theta}_{b}$ where
  $c$ is a boolean-valued function. For an example, see Figure 8 in the paper.
  \item Composites of use-cases above, like estimating mutual information
  $I(\bm{\theta}_{a}; \bm{\theta}_{b}) = \mathbb{E}_{\mathcal{P}(\bm{\theta}_{a},\bm{\theta}_{b})}[\log \frac{\mathcal{P}(\bm{\theta}_{a},\bm{\theta}_{b})}{\mathcal{P}(\bm{\theta}_{a})\mathcal{P}(\bm{\theta}_{b})}]$,
  where the joint and both marginals may be approximated first,
  followed by the analytical expectation of density ratio term with respect to $\hat{\mathcal{P}}(\bm{\theta}_{a},\bm{\theta}_{b})$,
  querying the marginal approximation densities using tree search.
  \item Any use-case above in a (axis-aligned, hyper-rectangular) subregion of $\Omega$, including mass integration or sampling,
  by forming a tree from a query region of the tree.
  \item Divergence estimation between different distributions over the same domain (or an overlap via the above).
\end{itemize}

\section{Algorithm}

\paragraph{Keeping track of $\hat{Z}$ and the highest mass partitions}
At every step of the algorithm,
we keep track of the current total mass estimate $\hat{Z}$,
and the top $M$ mass partitions, as required for checking CR2 and CR3.

We implement both of these through aggregators which are updated at each \emph{include} and \emph{exclude} of a partition in the \emph{leaf set}.
The leaf set is the partitions that cover the domain in a non-overlapping fashion at the currently finest resolution,
constituting the current Riemann sum.
'Include' refers to the creation of a new child node (and partition), and 'exclude' to the removal of the previous (parent)
node from this set.
To easily keep track of all the updates a partition division should lead to,
we wrap the 'divide function' in another function that after division (the forming of child partitions),
makes all the related updates, including the exclusion of the divided parent node.

The aggregator for the total mass estimate keeps track of the accumulate sum,
where inclusion is addition and exclusion is subtraction.
The aggregator for the top $M$ partitions keeps track of such a current set of size up to $M$,
where set updates are handled within the inclusion and exclusion functions.

\paragraph{Note on implementation}
Densities of typical density functions can have exceedingly small values, risking underflow using default float precisions.
In the implementation, we use a 128-bit float to present the total mass aggregate and the density evaluation stored in each respective partition node object.
An alternative is to represent the logarithm of the total mass and the densities,
but these values would need to either be transformed to higher precision before being exponentiated and used to avoid underflow,
or all checks and operations would need to be performed on the logarithm, which is more complicated.

\subsection{Representer points}
\label{sec:representer_points}

To check criterion CR2 and CR3 there are representer points to be chosen.
The representer points are used to efficiently (but approximately) check for overlaps between respective partitions
and the two spaces $\bm{\Phi}_{\mathrm{linear}}$ and $\bm{\Phi}_{\mathrm{balls}}$, associated with CR2 and CR3 respectively.
Note that this is an approximate method for checking overlaps,
and although there may be no false positives,
there may be false negatives.

For each representer point, a tree search is performed to find the associated partition,
which will be divided.
Note that if a partition fulfils multiple individually sufficient criteria (CR1 to CR3),
or is 'hit' by multiple representer points,
it will still only be divided once.

\paragraph{CR2}
We begin with the determination of $\bm{\Phi}_{\mathrm{linear}}$, followed by $\bm{R}_{\mathrm{linear}}$.
To form the linear subspace $\bm{\Phi}_{\mathrm{linear}}$ we carry out the following steps.
We first obtain the corresponding centroids $\Theta_H$ of the $H$ partitions as stacked vectors,
which is then used to form a basis for the $(H - 1)$-dimensional hyperplane being $\bm{\Phi}_{\mathrm{linear}}$.
Specifically, we let $\bm{\theta}_{j}$ be the centroid (column vector) among them closest to the centre of the unit cube,
and $E = \Theta_{H, \neq j} - \bm{\theta}_{j}$ be the stacked basis vectors of the hyperplane.
Then we obtain an orthogonal basis $A$ using QR factorization of $E$.
Let $\tilde{E}$ be the re-scaled version of $E$ having unit-norm.
Now we can map a vector $u$ from a unit-cube onto a given hyper-plane as $\bm{\theta}_{j} + u A \tilde{E}$,
as specified by the given centroids.

From $\bm{\Phi}_{\mathrm{linear}}$ a discrete set of points $\bm{R}_{\mathrm{linear}}$ is now to be chosen.
We will concentrate the points on all lower-dimensional hyper-planes formed by all combinations
of the $H$ high mass partitions,
which in line with the assumed heuristic that mass tends to concentrate on linear subspaces formed by these partitions.
In practice this is implemented by ahead of time determining all $\sum_{s=1}^S \binom{M}{s+1}$ combinations
of indices up to $M$ where $s$ is the dimensionality of a lower-dimensional hyperplane.
These index combinations are then iterated through at runtime at step $t$, each one creating a subset $H_\ast$ of high mass partitions forming a
linear subspace.
If a given set of centroids is colinear, it is skipped, as they would then only form a hyperplane of dimension \emph{less than} $s$,
and thus already be taken care of by another hyperplane.
For each $H_\ast$ we add a representer point at their average location in $\Omega$ (the weighted centre of the simplex they form).
In addition, to spread points also throughout the linear subspace (and the constituent lower-dimensional linear subspaces),
we add $l$ representer points per $H_\ast$ by sampling uniformly in a unit cube and map the points onto each lower dimensional linear subspace,
in addition to the (highest dimensional) subspace $\bm{\Phi}_{\mathrm{linear}}$,
using the procedure described above (we use $l = 1$ in all experiments).

\paragraph{CR3}
We now address the choice of the discrete set of points $\bm{R}_{\mathrm{balls}}$ from $\bm{\Phi}_{\mathrm{balls}}$.
For each partition in $H$, we sample $b$ points uniformly within the corresponding D-ball using the Muller method~\cite{harman2010decompositional}.
In practice we set $b = D$ in all experiments.
The representer points are used as specified in the paper.

\subsection{Algorithmic complexity analysis}
\label{sec:complexity_analysis}

We will now address the complexity analysis of the resulting algorithm.
As discussed in the paper, the algorithm's time complexity must allow short decision times and scalability to a large number of partitions.
In practice, we aim to maintain sub-millisecond decision times even after millions of density function evaluations.
To simplify the analysis, we will assume that the number of iterations of the algorithm (Algorithm~1) is proportional to $N_T$.
It will always be true that the number of iterations \emph{is less than} $N_T$, as multiple partitions will be divided at each iteration,
and also, each partition division will yield multiple function evaluations.

\paragraph{CR1}
Updating the hash map of heaps for a new partition has average time complexity $\mathcal{O}(\log \frac{N_t}{U})$, where $U$ is the number of unique abscissas (see Section 4).
This is because the heap, found in constant time using the hash map, provides logarithmic time updates, where $\nicefrac{N_t}{U}$ is the average number of elements in a heap.
As $N_t > U$ this simplifies as the worst-case $\mathcal{O}(\log \frac{N_t}{U}) = \mathcal{O}(\log N_t - \log U) = \mathcal{O}(\log N_t)$.
Note that this requires the tree to be balanced in the sense that the number of levels of the tree grows logarithmically with respect to the number of nodes.
Maximum tree imbalance would happen if, at each iteration, the node that is the highest number of levels away from the root is chosen to be divided.
However, note that such a node (or partition) would have an associated volume that exponentially tends to zero, with a denominator in the base at least \emph{three}, $(\frac{1}{3})^t$, which as a result would be very quickly dominated by other partitions.
This is an important fact, as all partition division criteria are either based on upper bound \emph{mass}, proportional to volume, or volume together with spatial vicinity.
As a result, all criteria encourage the balancing of the tree.
In practice, we will later see empirically that a logarithmic tree lookup time is maintained,
evidenced by that runtime time cost of iterations of the whole algorithm is no worse than logarithmic.

Retrieving all top ordinate partitions from this structure at a given iteration $t$ has time complexity $\mathcal{O}(U)$
as the hash table with $U$ entries is traversed linearly, and the root of each heap is available in constant time.
Using the $U$ obtained coordinates, the convex hull is computed in $\mathcal{O}(U \log U)$ using Graham's scan~\cite{graham1972efficient}.
Following this, the hull's upper-right quadrant is obtained in $\mathcal{O}(U)$.

In summary we obtain the worst-case time complexity $\mathcal{O}(\log N_t + U \log U)$ for checking this criterion for all partitions at step $t$.
The space complexity of the hash map of heaps data structure is linear with respect to $N_t$.

\paragraph{CR2}
The checking of this criterion entails computation associated with $\sum_{s=1}^S \binom{M}{s+1}$ linear subspaces,
where $M$ is the \emph{maximum} number of high mass partitions ($H$).
The number of partitions that will be divided as a result grows proportionally to this constant, so we simplify the analysis to treating one such linear subspace and an associated constant set of representer points.
The two largest terms come from the QR factorization,
with time complexity $\mathcal{O}(M^2 D)$, and the tree-search for the partition of a representer point has time complexity $\mathcal{O}(\log N_t)$.
The space complexity is $\mathcal{O}(M)$ of keeping track of the high mass partitions.
As $D$ is constant with respect to $N_t$ and $M$ is upper bounded by $D$, we summarise this as time complexity $\mathcal{O}(\log N_t)$ and the space complexity as constant.

\paragraph{CR3}
For the check of this criterion, all up to $M$ high mass partitions are sampled in $\mathcal{O}(M D)$ time,
and the tree-search for a partition is $\mathcal{O}(\log N_t)$.
Analogous to CR2, we summarise this as a time complexity $\mathcal{O}(\log N_t)$ and the space complexity as constant.

\paragraph{Summary}
\label{sec:method_summary}
The combined time complexity of a step $t$ is $\mathcal{O}(\log N_t + U \log U)$,
where $U$ is the number of \emph{unique} abscissas (see paper).
For an average number of steps proportional to $N_T$ this results in a total average time complexity of the algorithm as
$\mathcal{O}(N_T(\log N_T + \bar{U} \log \bar{U}))$.
The space complexity (including storage of partitions) of the algorithm is linear,~i.e.~remains proportional to the number of observations.
In Section~\ref{sec:runtime} we show empirically that $\bar{U}$ is sufficiently small,
and close to constant with respect to $N_T$,
leading to a fast and scalable algorithm.

\begin{figure*}[t]
  \includegraphics[width=0.32\textwidth]{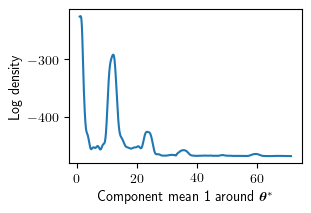}
  \includegraphics[width=0.32\textwidth]{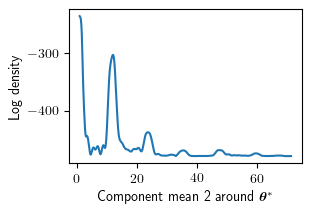}
  \includegraphics[width=0.32\textwidth]{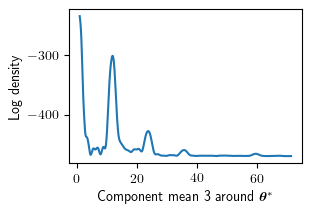}
  \\
  \includegraphics[width=0.32\textwidth]{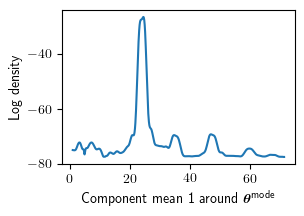}
  \includegraphics[width=0.32\textwidth]{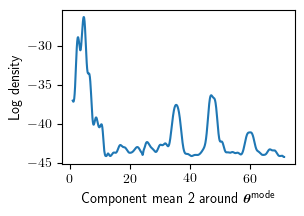}
  \includegraphics[width=0.32\textwidth]{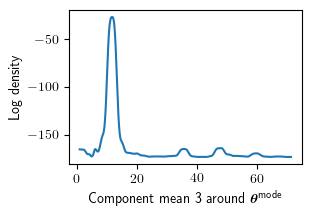}
  \caption{
  Multi-modality in the posterior of the parameters of the Spectral Mixture kernel.
  Shown is how the log density of the model changes with respect to each component mean within
  the Gaussian mixture representing the spectral density of the kernel.
  In the upper row the density changes with respect to each mean are shown at an uniformly sampled position in the (10D) domain.
  In the lower row these shown at the (estimated) mode of the parameter posterior.
  The estimation of the posterior mode was done using DEFER.
  }
  \label{fig:spectral_means}
\end{figure*}

\begin{figure*}[t]
  \centering
  \includegraphics[width=0.44\textwidth]{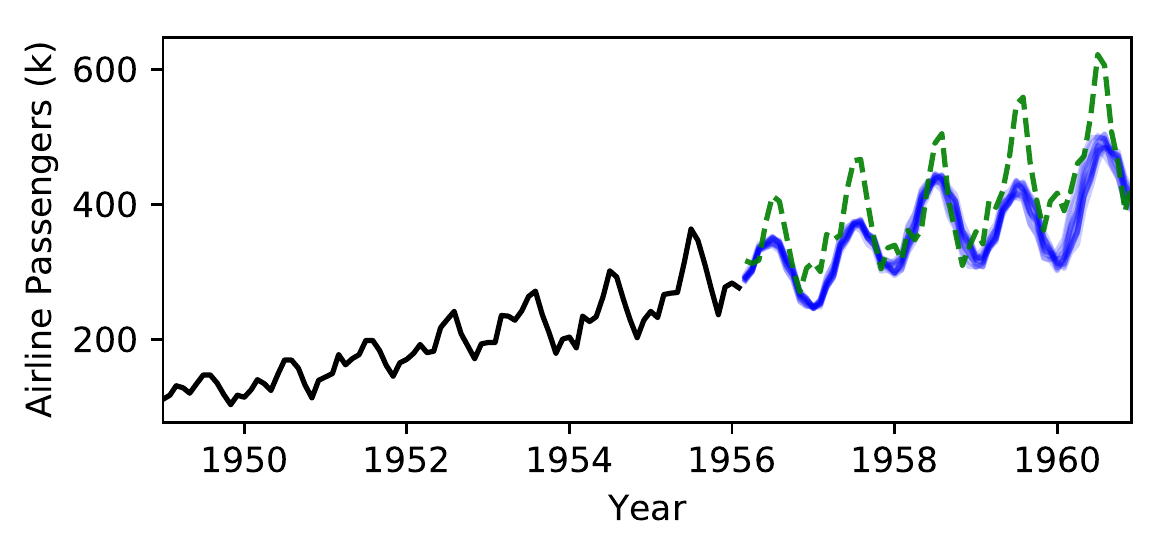}
  \includegraphics[width=0.44\textwidth]{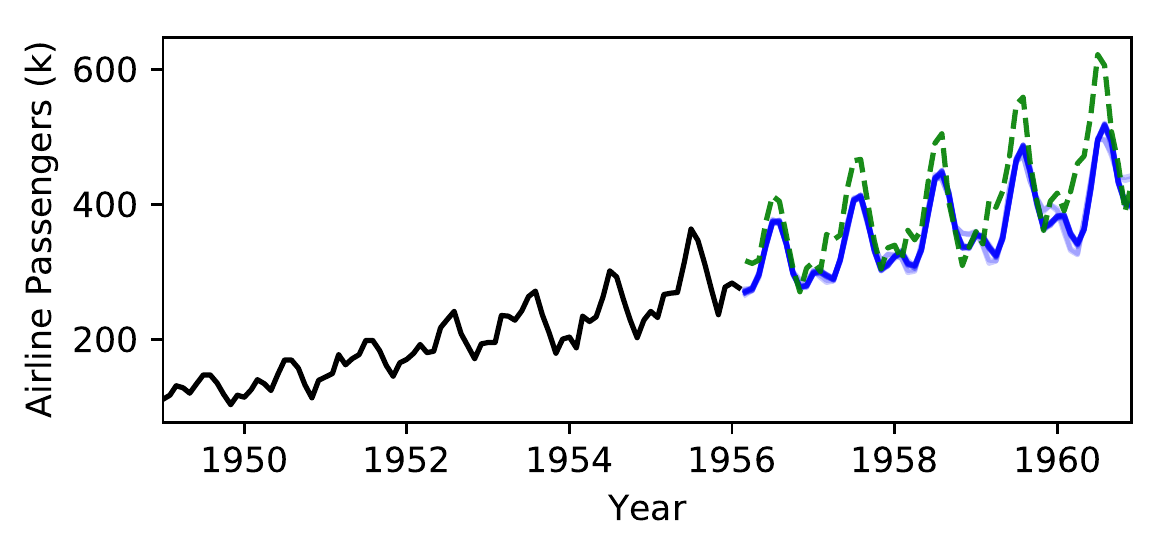}
  \includegraphics[width=0.44\textwidth]{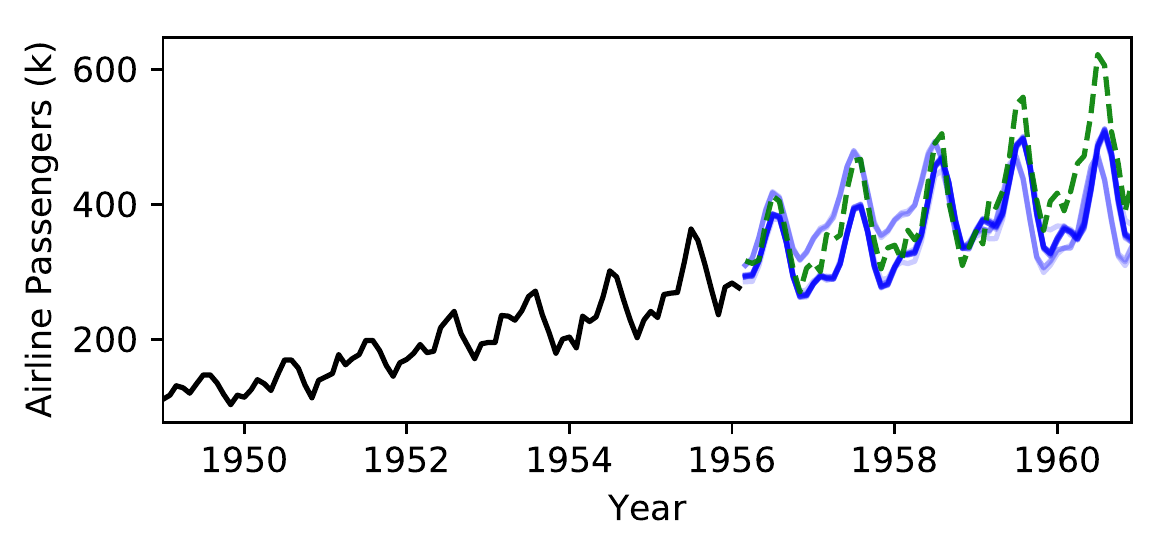}
  \includegraphics[width=0.44\textwidth]{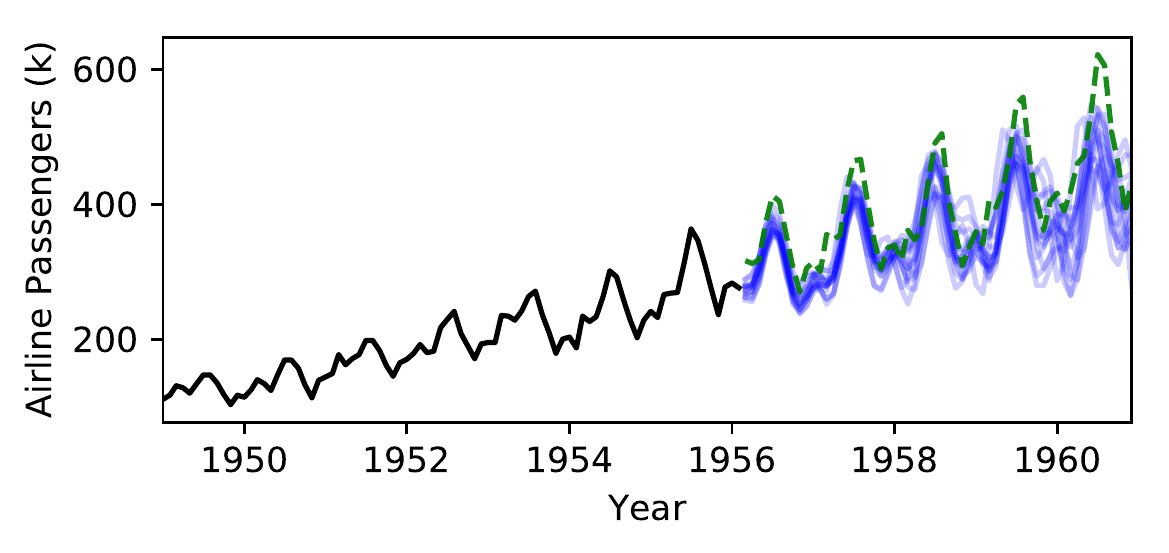}
  \caption{
  Time-series forecasting using Gaussian Processes with the Spectral Mixture~\cite{wilson2013gaussian} kernel.
  The data is shown in black, ground truth in green and posterior GP samples in blue.
  Shown from the top left is slice sampling, followed by PTMCMC on its right, then DNS and lastly DEFER in the bottom right.
  }
  \label{fig:spectral}
\end{figure*}

\section{Experiments}

\subsection{Runtime experiments}
\label{sec:runtime}

\begin{figure}[h]
  \centering

  \includegraphics[width=0.50\textwidth]{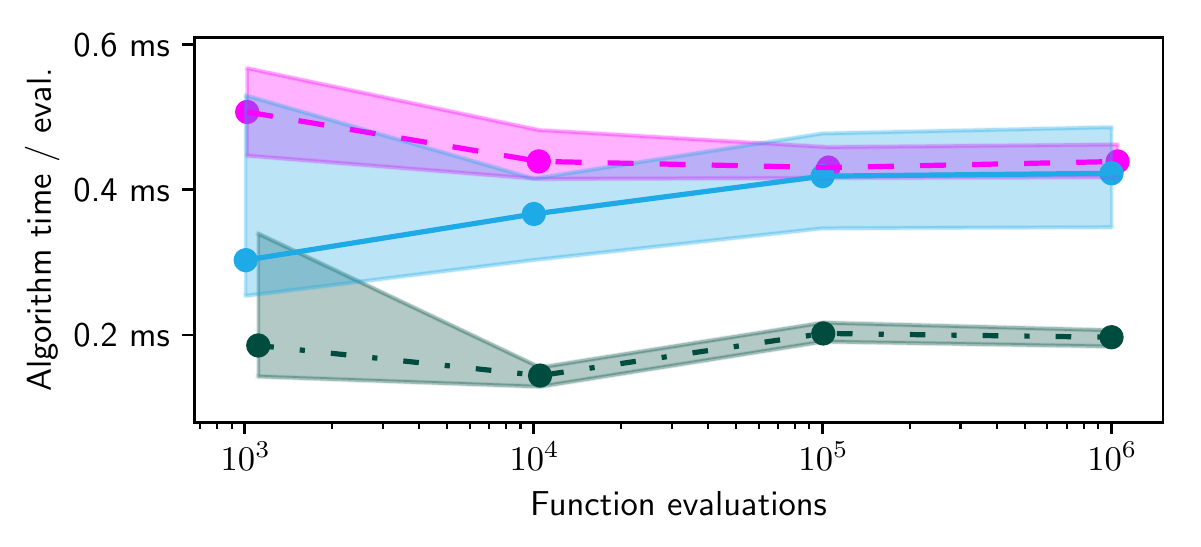}

  \includegraphics[width=0.50\textwidth]{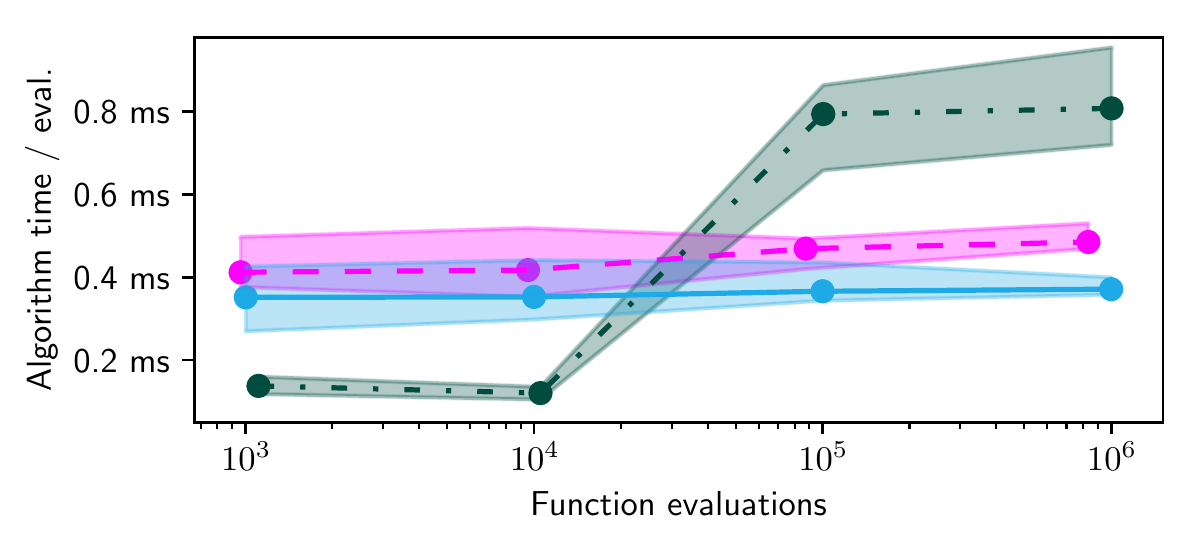}

  \includegraphics[width=0.50\textwidth]{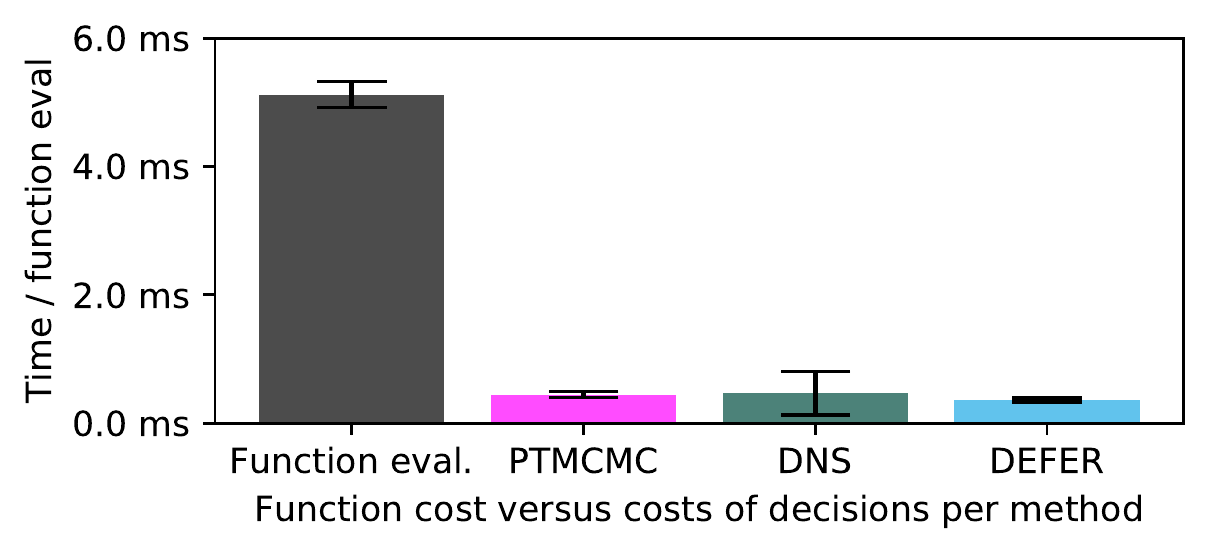}

  \caption{
  Algorithmic time.
  Shown in the two plots from the left is the associated runtime cost of each algorithm to decide where to evaluate the density function,
  for the Student's t 10D density function and gravitational-wave example, respectively (for legend see the bar plot on the right).
  Shown on the right is the time spent on evaluating the function versus the algorithm cost of respective method,
  with mean and standard deviation across 20 runs using the various settings of function evaluation budgets $N_T$.
  }
  \label{fig:table_timing}
\end{figure}

The DNS and DEFER were run (single-threaded) on a 2.6 GHz Intel Core i7, and PTMCMC used multiple cores due to its implementation.
In Figure~\ref{fig:table_timing} we note both that DEFER has a similar algorithmic cost to the other methods,
and that the DEFER has a near constant cost per function evaluation with respect to $N_T$.
With algorithmic cost we refer to the cost per 'decision' of where to evaluate the density function,
which is computed from the total (wall-clock) time of inference minus the total function evaluation time,
divided by the number of function evaluations made.
We also illustrate the common situation where the function evaluation time dominates the decision time,
making sample-efficiency a critical concern for total efficiency in these cases.

\begin{figure}[h]
  \centering
  \includegraphics[width=0.50\textwidth]{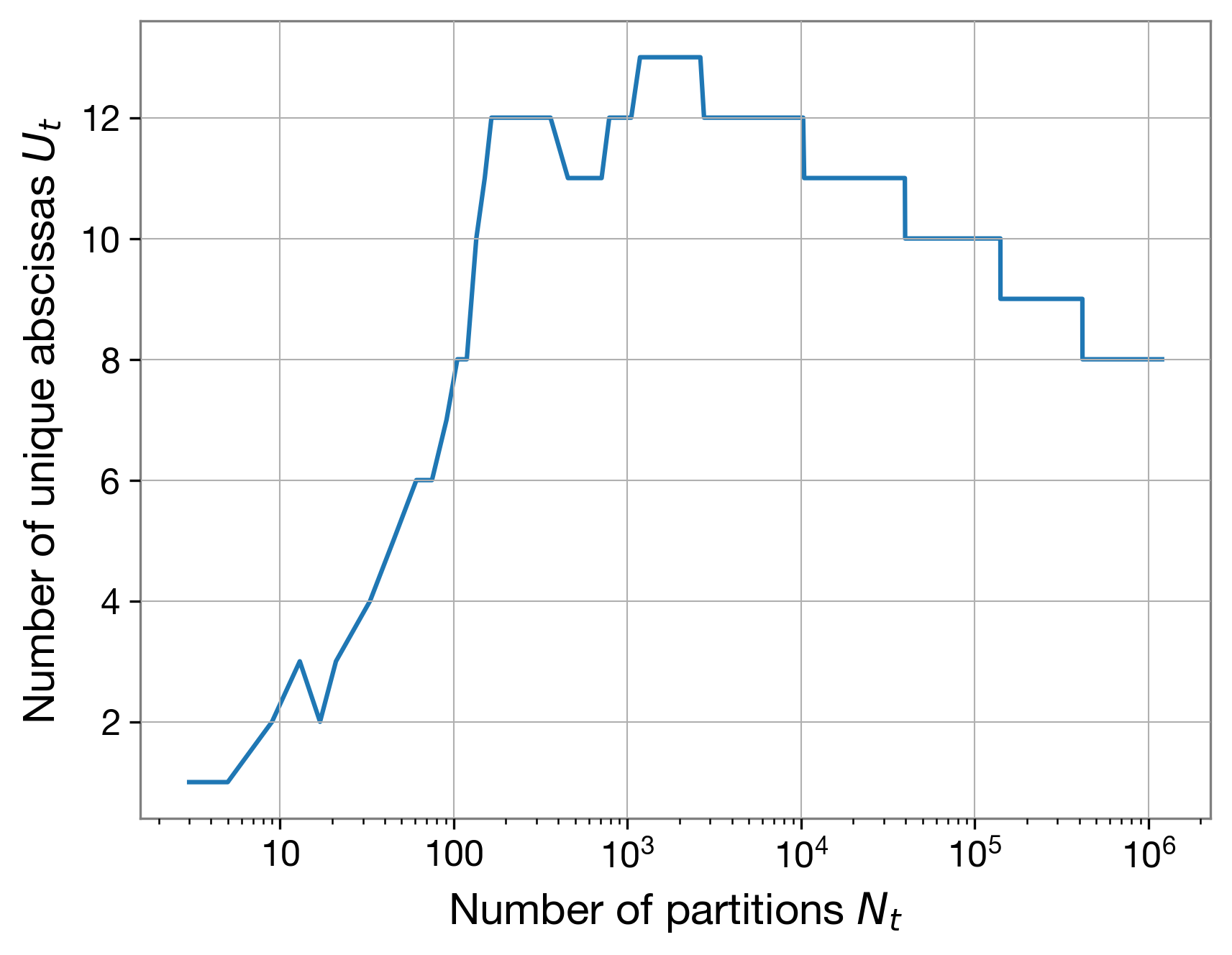}

  \caption{
  Shown in the plot is the number of unique abscissas $U_t$ per number of partitions $N_t$,
  as the algorithm executes.
  Note that $U_t$ does not continue growing with $N_t$ after a while, here after about 100 partitions.
  Used for the illustration is the Student's t 10D density function.
  }
  \label{fig:u_per_n}
\end{figure}

As discussed in Section~\ref{sec:complexity_analysis}, the time complexity of the algorithm is
$\mathcal{O}(N_T(\log N_T + \bar{U} \log \bar{U}))$.
To achieve the near linear scalability empirically illustrated,
it is clear that $\bar{U}$ must be close to constant with respect to $N_T$.
In Figure~\ref{fig:u_per_n} we confirm this explicitly.

\subsection{Ablation study of criteria}

\begin{figure*}[h]
  \centering

  \includegraphics[width=0.06\textwidth, frame]{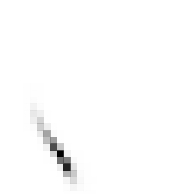}
  \includegraphics[width=0.06\textwidth, frame]{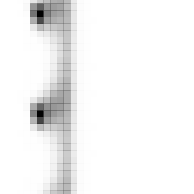}
  \includegraphics[width=0.06\textwidth, frame]{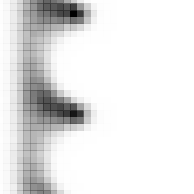}
  \includegraphics[width=0.06\textwidth, frame]{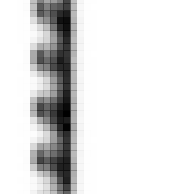}
  \includegraphics[width=0.06\textwidth, frame]{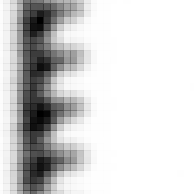}
  \includegraphics[width=0.06\textwidth, frame]{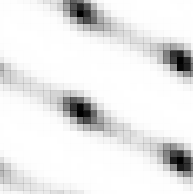}
  \includegraphics[width=0.06\textwidth, frame]{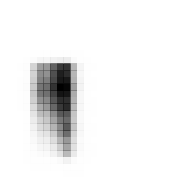}
  \includegraphics[width=0.06\textwidth, frame]{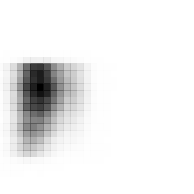}
  \includegraphics[width=0.06\textwidth, frame]{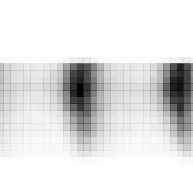}
  \includegraphics[width=0.06\textwidth, frame]{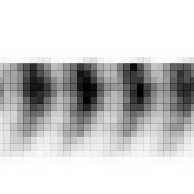}
  \includegraphics[width=0.06\textwidth, frame]{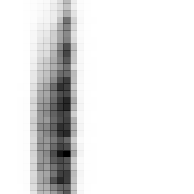}
  \includegraphics[width=0.06\textwidth, frame]{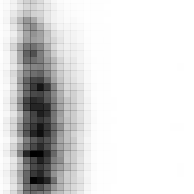}
  \includegraphics[width=0.06\textwidth, frame]{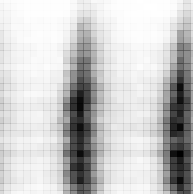}
  \includegraphics[width=0.06\textwidth, frame]{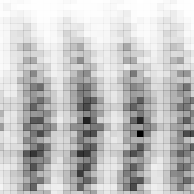}
  \includegraphics[width=0.06\textwidth, frame]{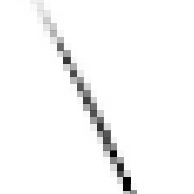}

  \vspace{0.5cm}

  \includegraphics[width=0.06\textwidth, frame]{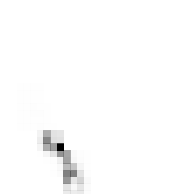}
  \includegraphics[width=0.06\textwidth, frame]{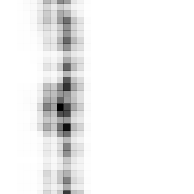}
  \includegraphics[width=0.06\textwidth, frame]{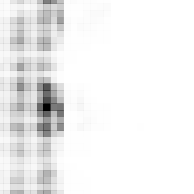}
  \includegraphics[width=0.06\textwidth, frame]{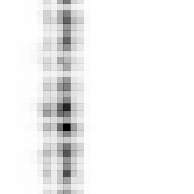}
  \includegraphics[width=0.06\textwidth, frame]{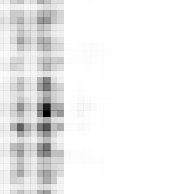}
  \includegraphics[width=0.06\textwidth, frame]{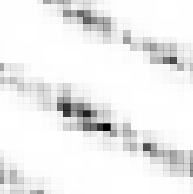}
  \includegraphics[width=0.06\textwidth, frame]{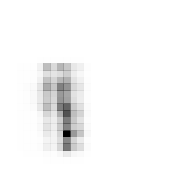}
  \includegraphics[width=0.06\textwidth, frame]{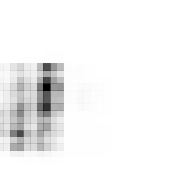}
  \includegraphics[width=0.06\textwidth, frame]{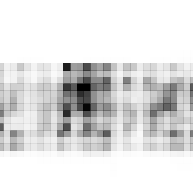}
  \includegraphics[width=0.06\textwidth, frame]{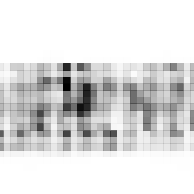}
  \includegraphics[width=0.06\textwidth, frame]{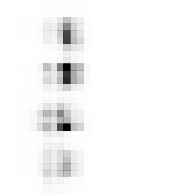}
  \includegraphics[width=0.06\textwidth, frame]{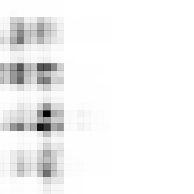}
  \includegraphics[width=0.06\textwidth, frame]{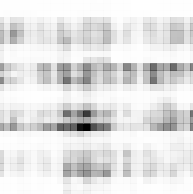}
  \includegraphics[width=0.06\textwidth, frame]{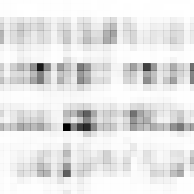}
  \includegraphics[width=0.06\textwidth, frame]{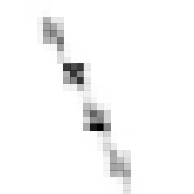}

  \includegraphics[width=0.06\textwidth, frame]{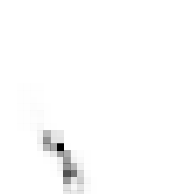}
  \includegraphics[width=0.06\textwidth, frame]{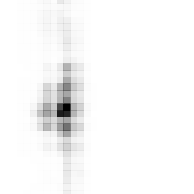}
  \includegraphics[width=0.06\textwidth, frame]{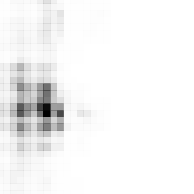}
  \includegraphics[width=0.06\textwidth, frame]{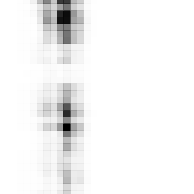}
  \includegraphics[width=0.06\textwidth, frame]{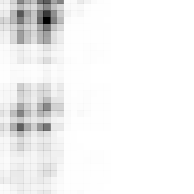}
  \includegraphics[width=0.06\textwidth, frame]{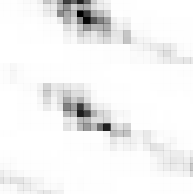}
  \includegraphics[width=0.06\textwidth, frame]{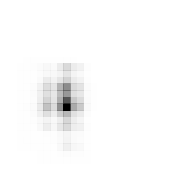}
  \includegraphics[width=0.06\textwidth, frame]{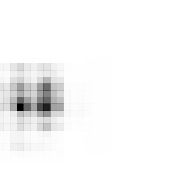}
  \includegraphics[width=0.06\textwidth, frame]{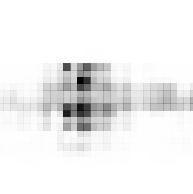}
  \includegraphics[width=0.06\textwidth, frame]{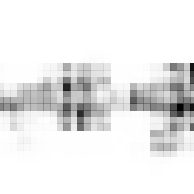}
  \includegraphics[width=0.06\textwidth, frame]{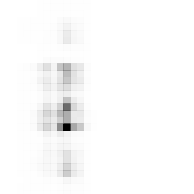}
  \includegraphics[width=0.06\textwidth, frame]{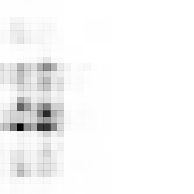}
  \includegraphics[width=0.06\textwidth, frame]{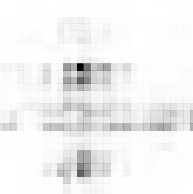}
  \includegraphics[width=0.06\textwidth, frame]{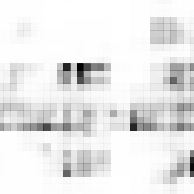}
  \includegraphics[width=0.06\textwidth, frame]{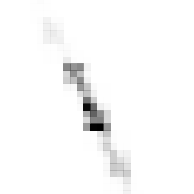}

  \includegraphics[width=0.06\textwidth, frame]{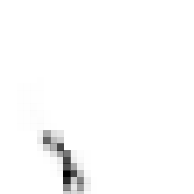}
  \includegraphics[width=0.06\textwidth, frame]{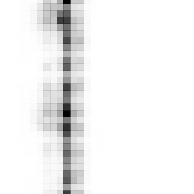}
  \includegraphics[width=0.06\textwidth, frame]{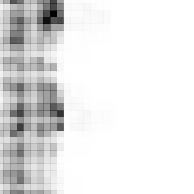}
  \includegraphics[width=0.06\textwidth, frame]{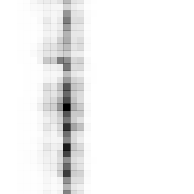}
  \includegraphics[width=0.06\textwidth, frame]{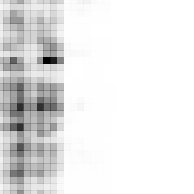}
  \includegraphics[width=0.06\textwidth, frame]{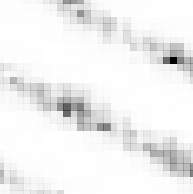}
  \includegraphics[width=0.06\textwidth, frame]{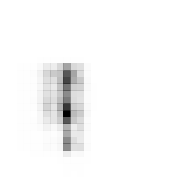}
  \includegraphics[width=0.06\textwidth, frame]{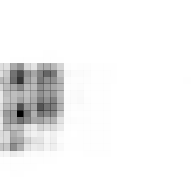}
  \includegraphics[width=0.06\textwidth, frame]{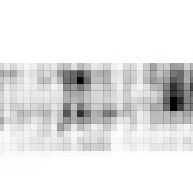}
  \includegraphics[width=0.06\textwidth, frame]{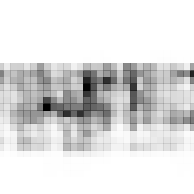}
  \includegraphics[width=0.06\textwidth, frame]{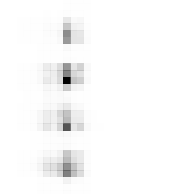}
  \includegraphics[width=0.06\textwidth, frame]{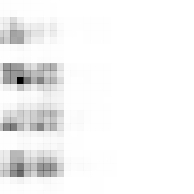}
  \includegraphics[width=0.06\textwidth, frame]{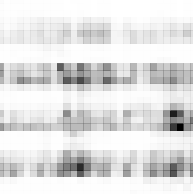}
  \includegraphics[width=0.06\textwidth, frame]{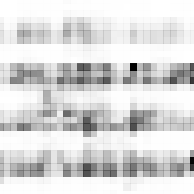}
  \includegraphics[width=0.06\textwidth, frame]{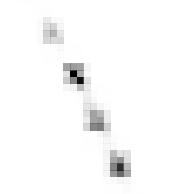}

  \vspace{0.5cm}

  \includegraphics[width=0.06\textwidth, frame]{figures/experiments/gw_subplots/ablation_study/gw_injected_6_defer_evals_5000001_bins_30_subplot_6}
  \includegraphics[width=0.06\textwidth, frame]{figures/experiments/gw_subplots/ablation_study/gw_injected_6_defer_evals_5000001_bins_30_subplot_12}
  \includegraphics[width=0.06\textwidth, frame]{figures/experiments/gw_subplots/ablation_study/gw_injected_6_defer_evals_5000001_bins_30_subplot_13}
  \includegraphics[width=0.06\textwidth, frame]{figures/experiments/gw_subplots/ablation_study/gw_injected_6_defer_evals_5000001_bins_30_subplot_18}
  \includegraphics[width=0.06\textwidth, frame]{figures/experiments/gw_subplots/ablation_study/gw_injected_6_defer_evals_5000001_bins_30_subplot_19}
  \includegraphics[width=0.06\textwidth, frame]{figures/experiments/gw_subplots/ablation_study/gw_injected_6_defer_evals_5000001_bins_30_subplot_20}
  \includegraphics[width=0.06\textwidth, frame]{figures/experiments/gw_subplots/ablation_study/gw_injected_6_defer_evals_5000001_bins_30_subplot_24}
  \includegraphics[width=0.06\textwidth, frame]{figures/experiments/gw_subplots/ablation_study/gw_injected_6_defer_evals_5000001_bins_30_subplot_25}
  \includegraphics[width=0.06\textwidth, frame]{figures/experiments/gw_subplots/ablation_study/gw_injected_6_defer_evals_5000001_bins_30_subplot_26}
  \includegraphics[width=0.06\textwidth, frame]{figures/experiments/gw_subplots/ablation_study/gw_injected_6_defer_evals_5000001_bins_30_subplot_27}
  \includegraphics[width=0.06\textwidth, frame]{figures/experiments/gw_subplots/ablation_study/gw_injected_6_defer_evals_5000001_bins_30_subplot_30}
  \includegraphics[width=0.06\textwidth, frame]{figures/experiments/gw_subplots/ablation_study/gw_injected_6_defer_evals_5000001_bins_30_subplot_31}
  \includegraphics[width=0.06\textwidth, frame]{figures/experiments/gw_subplots/ablation_study/gw_injected_6_defer_evals_5000001_bins_30_subplot_32}
  \includegraphics[width=0.06\textwidth, frame]{figures/experiments/gw_subplots/ablation_study/gw_injected_6_defer_evals_5000001_bins_30_subplot_33}
  \includegraphics[width=0.06\textwidth, frame]{figures/experiments/gw_subplots/ablation_study/gw_injected_6_defer_evals_5000001_bins_30_subplot_34}

  \includegraphics[width=0.06\textwidth, frame]{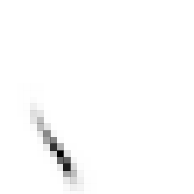}
  \includegraphics[width=0.06\textwidth, frame]{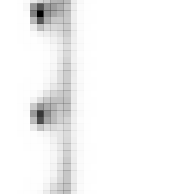}
  \includegraphics[width=0.06\textwidth, frame]{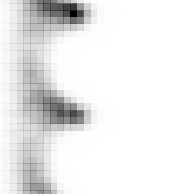}
  \includegraphics[width=0.06\textwidth, frame]{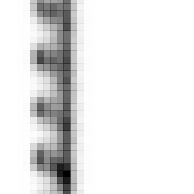}
  \includegraphics[width=0.06\textwidth, frame]{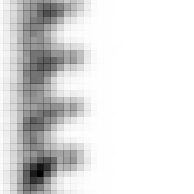}
  \includegraphics[width=0.06\textwidth, frame]{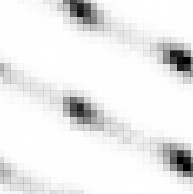}
  \includegraphics[width=0.06\textwidth, frame]{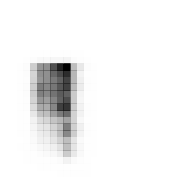}
  \includegraphics[width=0.06\textwidth, frame]{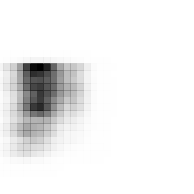}
  \includegraphics[width=0.06\textwidth, frame]{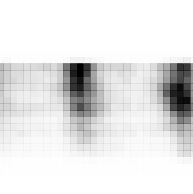}
  \includegraphics[width=0.06\textwidth, frame]{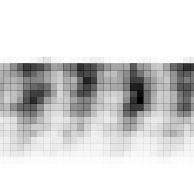}
  \includegraphics[width=0.06\textwidth, frame]{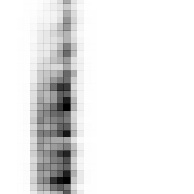}
  \includegraphics[width=0.06\textwidth, frame]{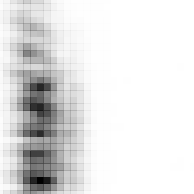}
  \includegraphics[width=0.06\textwidth, frame]{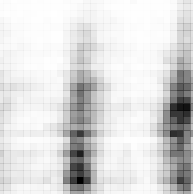}
  \includegraphics[width=0.06\textwidth, frame]{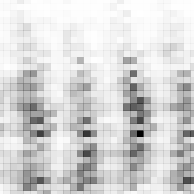}
  \includegraphics[width=0.06\textwidth, frame]{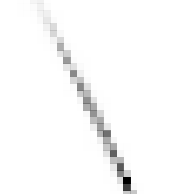}

  \includegraphics[width=0.06\textwidth, frame]{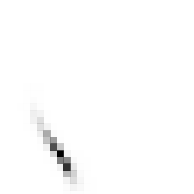}
  \includegraphics[width=0.06\textwidth, frame]{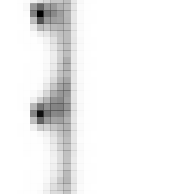}
  \includegraphics[width=0.06\textwidth, frame]{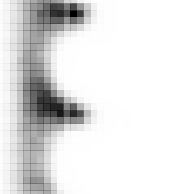}
  \includegraphics[width=0.06\textwidth, frame]{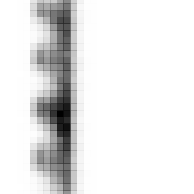}
  \includegraphics[width=0.06\textwidth, frame]{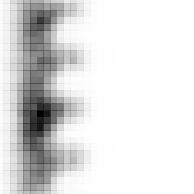}
  \includegraphics[width=0.06\textwidth, frame]{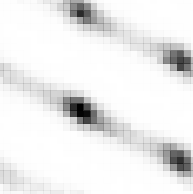}
  \includegraphics[width=0.06\textwidth, frame]{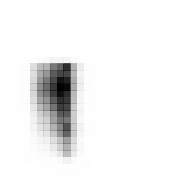}
  \includegraphics[width=0.06\textwidth, frame]{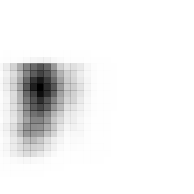}
  \includegraphics[width=0.06\textwidth, frame]{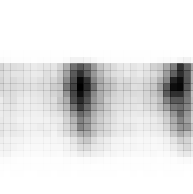}
  \includegraphics[width=0.06\textwidth, frame]{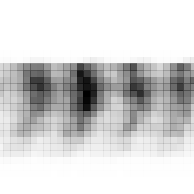}
  \includegraphics[width=0.06\textwidth, frame]{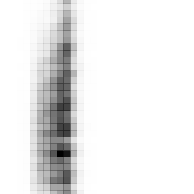}
  \includegraphics[width=0.06\textwidth, frame]{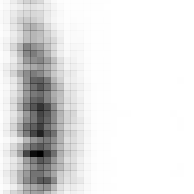}
  \includegraphics[width=0.06\textwidth, frame]{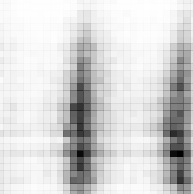}
  \includegraphics[width=0.06\textwidth, frame]{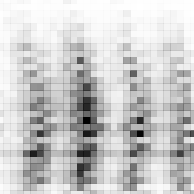}
  \includegraphics[width=0.06\textwidth, frame]{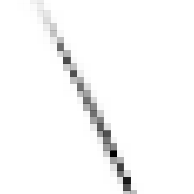}

  \caption{
  Visual ablation study of criteria on the gravitional-physics example.
  Shown are histograms of the two-dimensional marginals of the 6D problem (see Section~\ref{sec:gw_problem_details}),
  with samples produced after 200k density functions evaluations in the middle three rows,
  and after 5M evaluations in the bottom three rows.
  Shown in order, per group of three, is DEFER using all criteria (CR1-3),
  excluding CR2 (CR1,3), and excluding CR3 (CR1,2).
  The top row shows DEFER (using all criteria) after 10M evaluations for reference.
  Note that CR1 cannot be excluded as CR2 and CR3 would not explore the sample space at all on their own.
  We see that CR2, which complements the search along affine subspaces of high mass partitions,
  helps to 'fill in gaps' in various directions earlier than otherwise.
  CR3 has little effect in this example, but has seemingly no negative impact.
  For another example, see Figure~\ref{fig:gw2_ablation}.
  }
  \label{fig:gw_ablation}
\end{figure*}

\begin{figure*}[h]
  \centering

  \includegraphics[width=0.06\textwidth, frame]{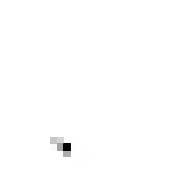}
  \includegraphics[width=0.06\textwidth, frame]{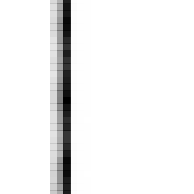}
  \includegraphics[width=0.06\textwidth, frame]{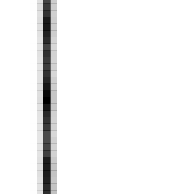}
  \includegraphics[width=0.06\textwidth, frame]{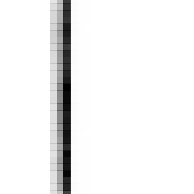}
  \includegraphics[width=0.06\textwidth, frame]{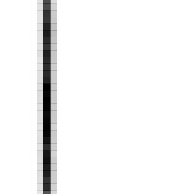}
  \includegraphics[width=0.06\textwidth, frame]{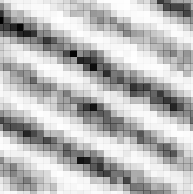}
  \includegraphics[width=0.06\textwidth, frame]{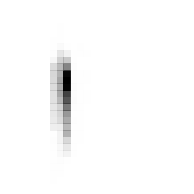}
  \includegraphics[width=0.06\textwidth, frame]{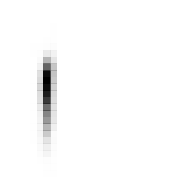}
  \includegraphics[width=0.06\textwidth, frame]{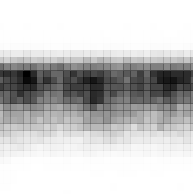}
  \includegraphics[width=0.06\textwidth, frame]{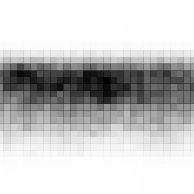}
  \includegraphics[width=0.06\textwidth, frame]{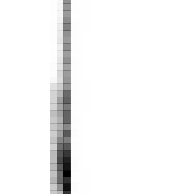}
  \includegraphics[width=0.06\textwidth, frame]{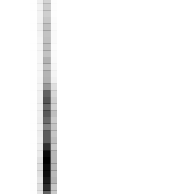}
  \includegraphics[width=0.06\textwidth, frame]{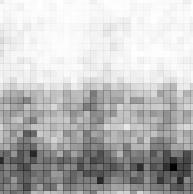}
  \includegraphics[width=0.06\textwidth, frame]{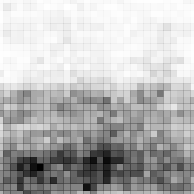}
  \includegraphics[width=0.06\textwidth, frame]{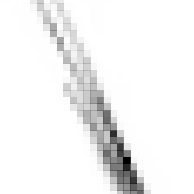}

  \vspace{0.5cm}

  \includegraphics[width=0.06\textwidth, frame]{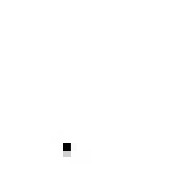}
  \includegraphics[width=0.06\textwidth, frame]{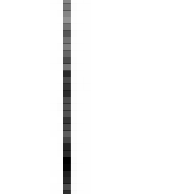}
  \includegraphics[width=0.06\textwidth, frame]{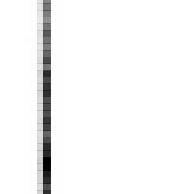}
  \includegraphics[width=0.06\textwidth, frame]{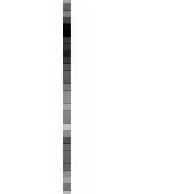}
  \includegraphics[width=0.06\textwidth, frame]{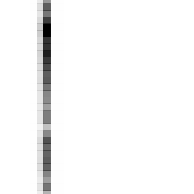}
  \includegraphics[width=0.06\textwidth, frame]{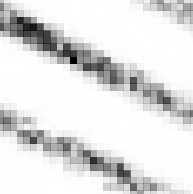}
  \includegraphics[width=0.06\textwidth, frame]{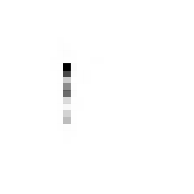}
  \includegraphics[width=0.06\textwidth, frame]{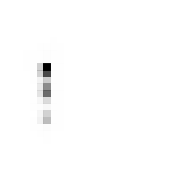}
  \includegraphics[width=0.06\textwidth, frame]{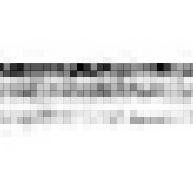}
  \includegraphics[width=0.06\textwidth, frame]{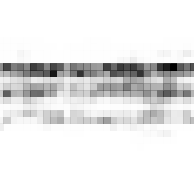}
  \includegraphics[width=0.06\textwidth, frame]{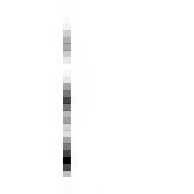}
  \includegraphics[width=0.06\textwidth, frame]{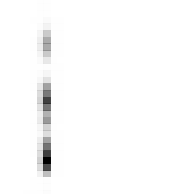}
  \includegraphics[width=0.06\textwidth, frame]{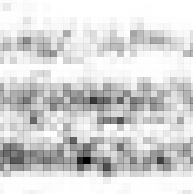}
  \includegraphics[width=0.06\textwidth, frame]{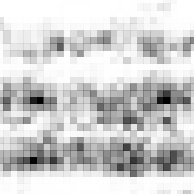}
  \includegraphics[width=0.06\textwidth, frame]{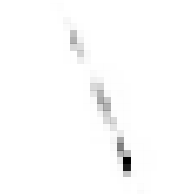}

  \includegraphics[width=0.06\textwidth, frame]{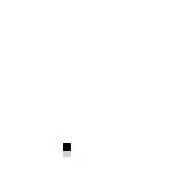}
  \includegraphics[width=0.06\textwidth, frame]{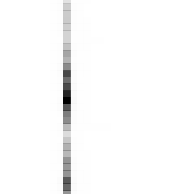}
  \includegraphics[width=0.06\textwidth, frame]{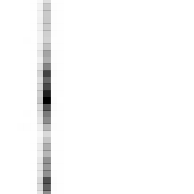}
  \includegraphics[width=0.06\textwidth, frame]{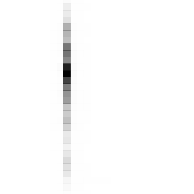}
  \includegraphics[width=0.06\textwidth, frame]{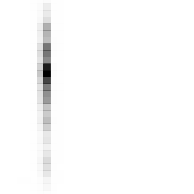}
  \includegraphics[width=0.06\textwidth, frame]{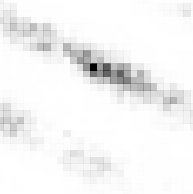}
  \includegraphics[width=0.06\textwidth, frame]{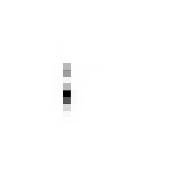}
  \includegraphics[width=0.06\textwidth, frame]{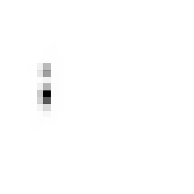}
  \includegraphics[width=0.06\textwidth, frame]{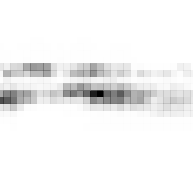}
  \includegraphics[width=0.06\textwidth, frame]{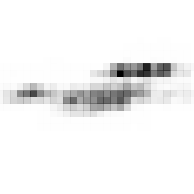}
  \includegraphics[width=0.06\textwidth, frame]{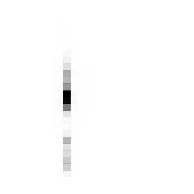}
  \includegraphics[width=0.06\textwidth, frame]{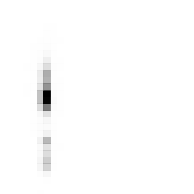}
  \includegraphics[width=0.06\textwidth, frame]{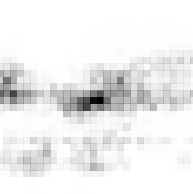}
  \includegraphics[width=0.06\textwidth, frame]{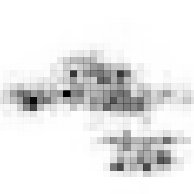}
  \includegraphics[width=0.06\textwidth, frame]{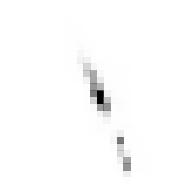}

  \includegraphics[width=0.06\textwidth, frame]{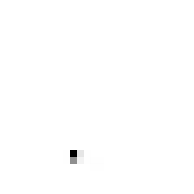}
  \includegraphics[width=0.06\textwidth, frame]{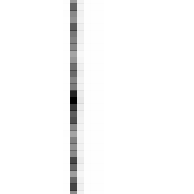}
  \includegraphics[width=0.06\textwidth, frame]{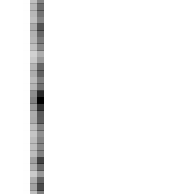}
  \includegraphics[width=0.06\textwidth, frame]{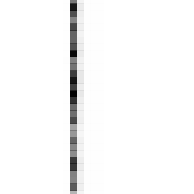}
  \includegraphics[width=0.06\textwidth, frame]{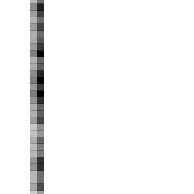}
  \includegraphics[width=0.06\textwidth, frame]{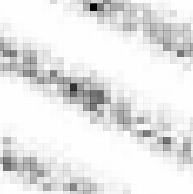}
  \includegraphics[width=0.06\textwidth, frame]{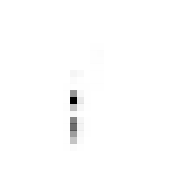}
  \includegraphics[width=0.06\textwidth, frame]{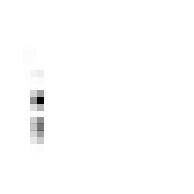}
  \includegraphics[width=0.06\textwidth, frame]{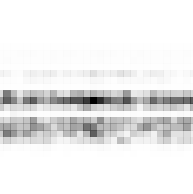}
  \includegraphics[width=0.06\textwidth, frame]{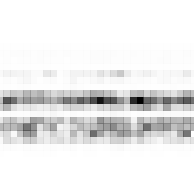}
  \includegraphics[width=0.06\textwidth, frame]{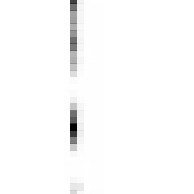}
  \includegraphics[width=0.06\textwidth, frame]{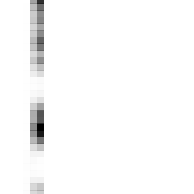}
  \includegraphics[width=0.06\textwidth, frame]{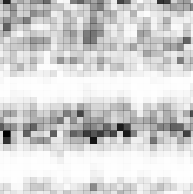}
  \includegraphics[width=0.06\textwidth, frame]{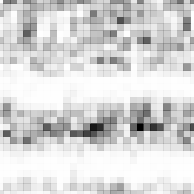}
  \includegraphics[width=0.06\textwidth, frame]{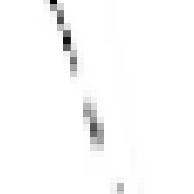}

  \vspace{0.5cm}

  \includegraphics[width=0.06\textwidth, frame]{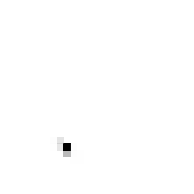}
  \includegraphics[width=0.06\textwidth, frame]{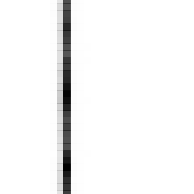}
  \includegraphics[width=0.06\textwidth, frame]{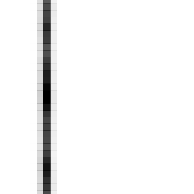}
  \includegraphics[width=0.06\textwidth, frame]{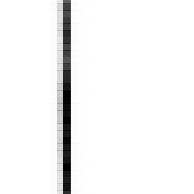}
  \includegraphics[width=0.06\textwidth, frame]{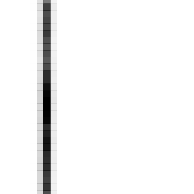}
  \includegraphics[width=0.06\textwidth, frame]{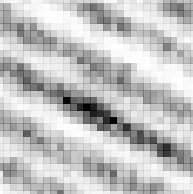}
  \includegraphics[width=0.06\textwidth, frame]{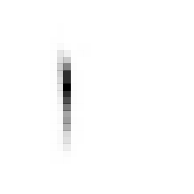}
  \includegraphics[width=0.06\textwidth, frame]{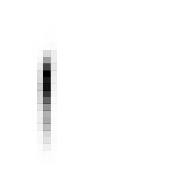}
  \includegraphics[width=0.06\textwidth, frame]{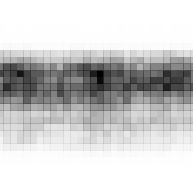}
  \includegraphics[width=0.06\textwidth, frame]{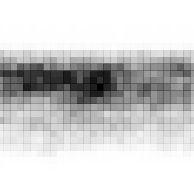}
  \includegraphics[width=0.06\textwidth, frame]{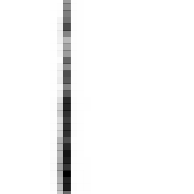}
  \includegraphics[width=0.06\textwidth, frame]{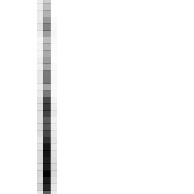}
  \includegraphics[width=0.06\textwidth, frame]{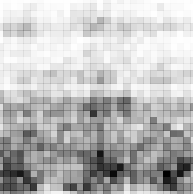}
  \includegraphics[width=0.06\textwidth, frame]{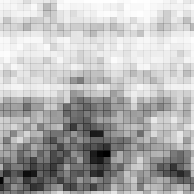}
  \includegraphics[width=0.06\textwidth, frame]{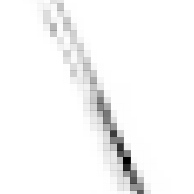}

  \includegraphics[width=0.06\textwidth, frame]{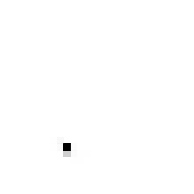}
  \includegraphics[width=0.06\textwidth, frame]{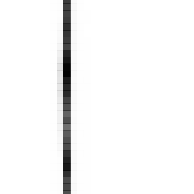}
  \includegraphics[width=0.06\textwidth, frame]{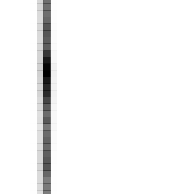}
  \includegraphics[width=0.06\textwidth, frame]{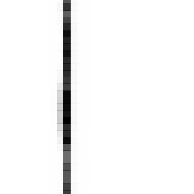}
  \includegraphics[width=0.06\textwidth, frame]{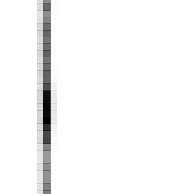}
  \includegraphics[width=0.06\textwidth, frame]{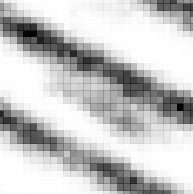}
  \includegraphics[width=0.06\textwidth, frame]{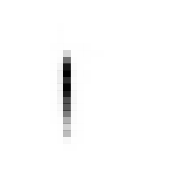}
  \includegraphics[width=0.06\textwidth, frame]{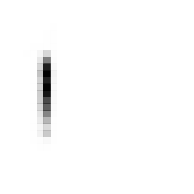}
  \includegraphics[width=0.06\textwidth, frame]{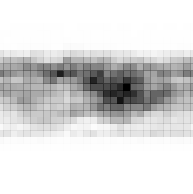}
  \includegraphics[width=0.06\textwidth, frame]{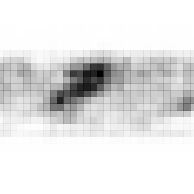}
  \includegraphics[width=0.06\textwidth, frame]{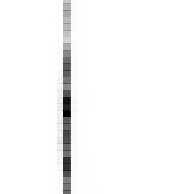}
  \includegraphics[width=0.06\textwidth, frame]{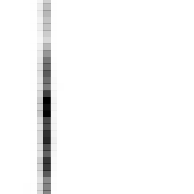}
  \includegraphics[width=0.06\textwidth, frame]{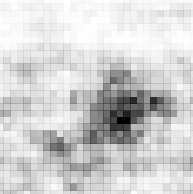}
  \includegraphics[width=0.06\textwidth, frame]{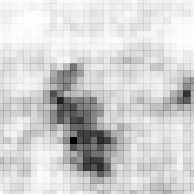}
  \includegraphics[width=0.06\textwidth, frame]{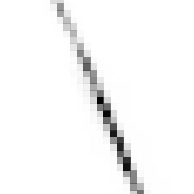}

  \includegraphics[width=0.06\textwidth, frame]{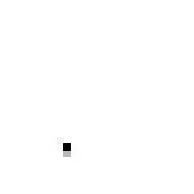}
  \includegraphics[width=0.06\textwidth, frame]{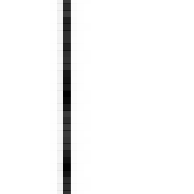}
  \includegraphics[width=0.06\textwidth, frame]{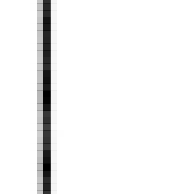}
  \includegraphics[width=0.06\textwidth, frame]{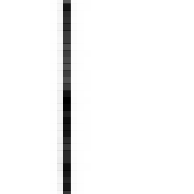}
  \includegraphics[width=0.06\textwidth, frame]{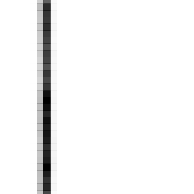}
  \includegraphics[width=0.06\textwidth, frame]{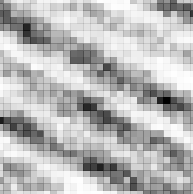}
  \includegraphics[width=0.06\textwidth, frame]{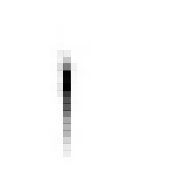}
  \includegraphics[width=0.06\textwidth, frame]{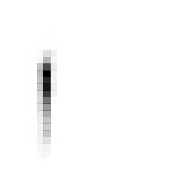}
  \includegraphics[width=0.06\textwidth, frame]{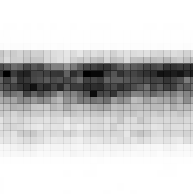}
  \includegraphics[width=0.06\textwidth, frame]{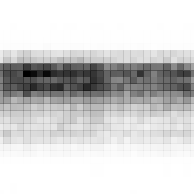}
  \includegraphics[width=0.06\textwidth, frame]{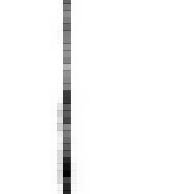}
  \includegraphics[width=0.06\textwidth, frame]{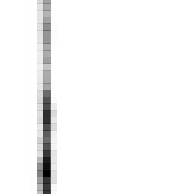}
  \includegraphics[width=0.06\textwidth, frame]{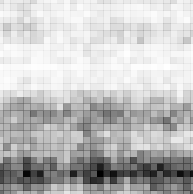}
  \includegraphics[width=0.06\textwidth, frame]{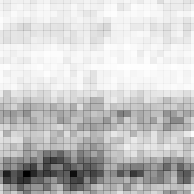}
  \includegraphics[width=0.06\textwidth, frame]{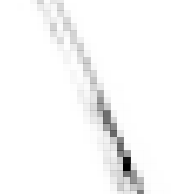}

  \caption{
  Visual ablation study of criteria on another gravitional-physics example,
  this time inferring the masses of the two black holes instead of the luminosity distance to the source and the orbital phase,
  together with the four other parameters (see Section~\ref{sec:gw_problem_details}).
  Shown are histograms of the two-dimensional marginals,
  with samples produced after 500k density functions evaluations in the middle three rows,
  and after 5M evaluations in the bottom three rows.
  Shown in order, per group of three, is DEFER using all criteria (CR1-3),
  excluding CR2 (CR1,3), and excluding CR3 (CR1,2).
  The top row shows DEFER (using all criteria) after 10M evaluations for reference.
  Note that CR1 cannot be excluded as CR2 and CR3 would not explore the sample space at all on their own.
  Similar to in Figure~\ref{fig:gw_ablation},
  we see that CR2 helps to fill in details earlier than otherwise,
  and the same is true for CR3 although to a lesser extent.
  }
  \label{fig:gw2_ablation}
\end{figure*}

\begin{figure*}[h]
  \centering

  \includegraphics[width=0.06\textwidth, frame]{figures/experiments/gw_subplots/ablation_study/gw_injected_6_defer_evals_200007_bins_30_subplot_6}
  \includegraphics[width=0.06\textwidth, frame]{figures/experiments/gw_subplots/ablation_study/gw_injected_6_defer_evals_200007_bins_30_subplot_12}
  \includegraphics[width=0.06\textwidth, frame]{figures/experiments/gw_subplots/ablation_study/gw_injected_6_defer_evals_200007_bins_30_subplot_13}
  \includegraphics[width=0.06\textwidth, frame]{figures/experiments/gw_subplots/ablation_study/gw_injected_6_defer_evals_200007_bins_30_subplot_18}
  \includegraphics[width=0.06\textwidth, frame]{figures/experiments/gw_subplots/ablation_study/gw_injected_6_defer_evals_200007_bins_30_subplot_19}
  \includegraphics[width=0.06\textwidth, frame]{figures/experiments/gw_subplots/ablation_study/gw_injected_6_defer_evals_200007_bins_30_subplot_20}
  \includegraphics[width=0.06\textwidth, frame]{figures/experiments/gw_subplots/ablation_study/gw_injected_6_defer_evals_200007_bins_30_subplot_24}
  \includegraphics[width=0.06\textwidth, frame]{figures/experiments/gw_subplots/ablation_study/gw_injected_6_defer_evals_200007_bins_30_subplot_25}
  \includegraphics[width=0.06\textwidth, frame]{figures/experiments/gw_subplots/ablation_study/gw_injected_6_defer_evals_200007_bins_30_subplot_26}
  \includegraphics[width=0.06\textwidth, frame]{figures/experiments/gw_subplots/ablation_study/gw_injected_6_defer_evals_200007_bins_30_subplot_27}
  \includegraphics[width=0.06\textwidth, frame]{figures/experiments/gw_subplots/ablation_study/gw_injected_6_defer_evals_200007_bins_30_subplot_30}
  \includegraphics[width=0.06\textwidth, frame]{figures/experiments/gw_subplots/ablation_study/gw_injected_6_defer_evals_200007_bins_30_subplot_31}
  \includegraphics[width=0.06\textwidth, frame]{figures/experiments/gw_subplots/ablation_study/gw_injected_6_defer_evals_200007_bins_30_subplot_32}
  \includegraphics[width=0.06\textwidth, frame]{figures/experiments/gw_subplots/ablation_study/gw_injected_6_defer_evals_200007_bins_30_subplot_33}
  \includegraphics[width=0.06\textwidth, frame]{figures/experiments/gw_subplots/ablation_study/gw_injected_6_defer_evals_200007_bins_30_subplot_34}

  \includegraphics[width=0.06\textwidth, frame]{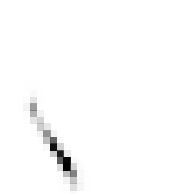}
  \includegraphics[width=0.06\textwidth, frame]{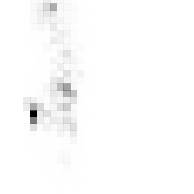}
  \includegraphics[width=0.06\textwidth, frame]{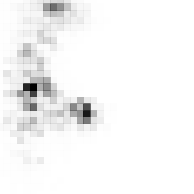}
  \includegraphics[width=0.06\textwidth, frame]{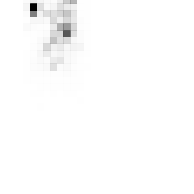}
  \includegraphics[width=0.06\textwidth, frame]{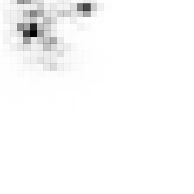}
  \includegraphics[width=0.06\textwidth, frame]{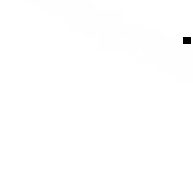}
  \includegraphics[width=0.06\textwidth, frame]{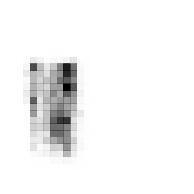}
  \includegraphics[width=0.06\textwidth, frame]{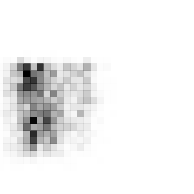}
  \includegraphics[width=0.06\textwidth, frame]{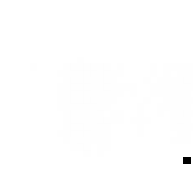}
  \includegraphics[width=0.06\textwidth, frame]{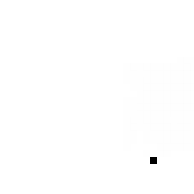}
  \includegraphics[width=0.06\textwidth, frame]{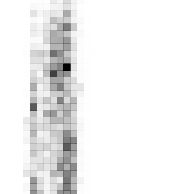}
  \includegraphics[width=0.06\textwidth, frame]{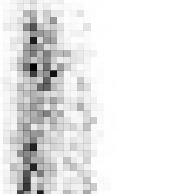}
  \includegraphics[width=0.06\textwidth, frame]{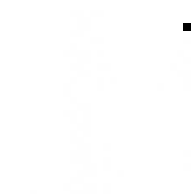}
  \includegraphics[width=0.06\textwidth, frame]{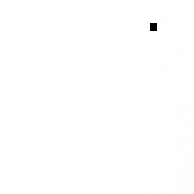}
  \includegraphics[width=0.06\textwidth, frame]{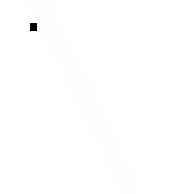}

  \includegraphics[width=0.06\textwidth, frame]{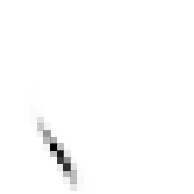}
  \includegraphics[width=0.06\textwidth, frame]{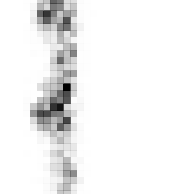}
  \includegraphics[width=0.06\textwidth, frame]{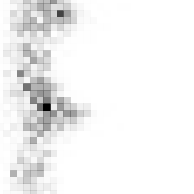}
  \includegraphics[width=0.06\textwidth, frame]{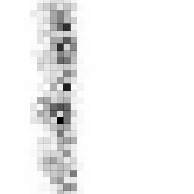}
  \includegraphics[width=0.06\textwidth, frame]{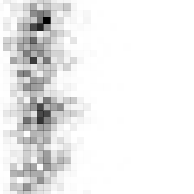}
  \includegraphics[width=0.06\textwidth, frame]{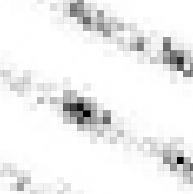}
  \includegraphics[width=0.06\textwidth, frame]{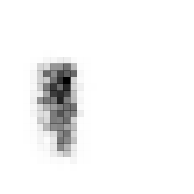}
  \includegraphics[width=0.06\textwidth, frame]{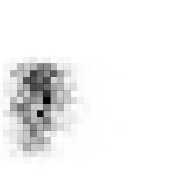}
  \includegraphics[width=0.06\textwidth, frame]{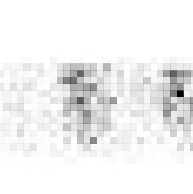}
  \includegraphics[width=0.06\textwidth, frame]{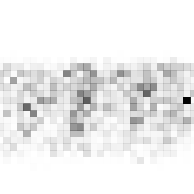}
  \includegraphics[width=0.06\textwidth, frame]{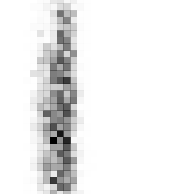}
  \includegraphics[width=0.06\textwidth, frame]{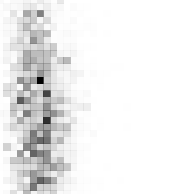}
  \includegraphics[width=0.06\textwidth, frame]{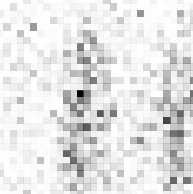}
  \includegraphics[width=0.06\textwidth, frame]{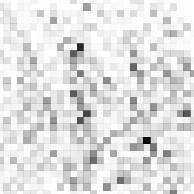}
  \includegraphics[width=0.06\textwidth, frame]{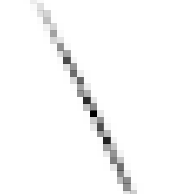}

  \vspace{0.5cm}

  \includegraphics[width=0.06\textwidth, frame]{figures/experiments/gw_subplots/ablation_study/gw_injected_6_defer_evals_5000001_bins_30_subplot_6}
  \includegraphics[width=0.06\textwidth, frame]{figures/experiments/gw_subplots/ablation_study/gw_injected_6_defer_evals_5000001_bins_30_subplot_12}
  \includegraphics[width=0.06\textwidth, frame]{figures/experiments/gw_subplots/ablation_study/gw_injected_6_defer_evals_5000001_bins_30_subplot_13}
  \includegraphics[width=0.06\textwidth, frame]{figures/experiments/gw_subplots/ablation_study/gw_injected_6_defer_evals_5000001_bins_30_subplot_18}
  \includegraphics[width=0.06\textwidth, frame]{figures/experiments/gw_subplots/ablation_study/gw_injected_6_defer_evals_5000001_bins_30_subplot_19}
  \includegraphics[width=0.06\textwidth, frame]{figures/experiments/gw_subplots/ablation_study/gw_injected_6_defer_evals_5000001_bins_30_subplot_20}
  \includegraphics[width=0.06\textwidth, frame]{figures/experiments/gw_subplots/ablation_study/gw_injected_6_defer_evals_5000001_bins_30_subplot_24}
  \includegraphics[width=0.06\textwidth, frame]{figures/experiments/gw_subplots/ablation_study/gw_injected_6_defer_evals_5000001_bins_30_subplot_25}
  \includegraphics[width=0.06\textwidth, frame]{figures/experiments/gw_subplots/ablation_study/gw_injected_6_defer_evals_5000001_bins_30_subplot_26}
  \includegraphics[width=0.06\textwidth, frame]{figures/experiments/gw_subplots/ablation_study/gw_injected_6_defer_evals_5000001_bins_30_subplot_27}
  \includegraphics[width=0.06\textwidth, frame]{figures/experiments/gw_subplots/ablation_study/gw_injected_6_defer_evals_5000001_bins_30_subplot_30}
  \includegraphics[width=0.06\textwidth, frame]{figures/experiments/gw_subplots/ablation_study/gw_injected_6_defer_evals_5000001_bins_30_subplot_31}
  \includegraphics[width=0.06\textwidth, frame]{figures/experiments/gw_subplots/ablation_study/gw_injected_6_defer_evals_5000001_bins_30_subplot_32}
  \includegraphics[width=0.06\textwidth, frame]{figures/experiments/gw_subplots/ablation_study/gw_injected_6_defer_evals_5000001_bins_30_subplot_33}
  \includegraphics[width=0.06\textwidth, frame]{figures/experiments/gw_subplots/ablation_study/gw_injected_6_defer_evals_5000001_bins_30_subplot_34}

  \includegraphics[width=0.06\textwidth, frame]{figures/experiments/gw_subplots/baselines/gw_injected_6_ptmcmc_evals_4486439_bins_30_subplot_6}
  \includegraphics[width=0.06\textwidth, frame]{figures/experiments/gw_subplots/baselines/gw_injected_6_ptmcmc_evals_4486439_bins_30_subplot_12}
  \includegraphics[width=0.06\textwidth, frame]{figures/experiments/gw_subplots/baselines/gw_injected_6_ptmcmc_evals_4486439_bins_30_subplot_13}
  \includegraphics[width=0.06\textwidth, frame]{figures/experiments/gw_subplots/baselines/gw_injected_6_ptmcmc_evals_4486439_bins_30_subplot_18}
  \includegraphics[width=0.06\textwidth, frame]{figures/experiments/gw_subplots/baselines/gw_injected_6_ptmcmc_evals_4486439_bins_30_subplot_19}
  \includegraphics[width=0.06\textwidth, frame]{figures/experiments/gw_subplots/baselines/gw_injected_6_ptmcmc_evals_4486439_bins_30_subplot_20}
  \includegraphics[width=0.06\textwidth, frame]{figures/experiments/gw_subplots/baselines/gw_injected_6_ptmcmc_evals_4486439_bins_30_subplot_24}
  \includegraphics[width=0.06\textwidth, frame]{figures/experiments/gw_subplots/baselines/gw_injected_6_ptmcmc_evals_4486439_bins_30_subplot_25}
  \includegraphics[width=0.06\textwidth, frame]{figures/experiments/gw_subplots/baselines/gw_injected_6_ptmcmc_evals_4486439_bins_30_subplot_26}
  \includegraphics[width=0.06\textwidth, frame]{figures/experiments/gw_subplots/baselines/gw_injected_6_ptmcmc_evals_4486439_bins_30_subplot_27}
  \includegraphics[width=0.06\textwidth, frame]{figures/experiments/gw_subplots/baselines/gw_injected_6_ptmcmc_evals_4486439_bins_30_subplot_30}
  \includegraphics[width=0.06\textwidth, frame]{figures/experiments/gw_subplots/baselines/gw_injected_6_ptmcmc_evals_4486439_bins_30_subplot_31}
  \includegraphics[width=0.06\textwidth, frame]{figures/experiments/gw_subplots/baselines/gw_injected_6_ptmcmc_evals_4486439_bins_30_subplot_32}
  \includegraphics[width=0.06\textwidth, frame]{figures/experiments/gw_subplots/baselines/gw_injected_6_ptmcmc_evals_4486439_bins_30_subplot_33}
  \includegraphics[width=0.06\textwidth, frame]{figures/experiments/gw_subplots/baselines/gw_injected_6_ptmcmc_evals_4486439_bins_30_subplot_34}

  \includegraphics[width=0.06\textwidth, frame]{figures/experiments/gw_subplots/baselines/gw_injected_6_dns_evals_5002325_bins_30_subplot_6}
  \includegraphics[width=0.06\textwidth, frame]{figures/experiments/gw_subplots/baselines/gw_injected_6_dns_evals_5002325_bins_30_subplot_12}
  \includegraphics[width=0.06\textwidth, frame]{figures/experiments/gw_subplots/baselines/gw_injected_6_dns_evals_5002325_bins_30_subplot_13}
  \includegraphics[width=0.06\textwidth, frame]{figures/experiments/gw_subplots/baselines/gw_injected_6_dns_evals_5002325_bins_30_subplot_18}
  \includegraphics[width=0.06\textwidth, frame]{figures/experiments/gw_subplots/baselines/gw_injected_6_dns_evals_5002325_bins_30_subplot_19}
  \includegraphics[width=0.06\textwidth, frame]{figures/experiments/gw_subplots/baselines/gw_injected_6_dns_evals_5002325_bins_30_subplot_20}
  \includegraphics[width=0.06\textwidth, frame]{figures/experiments/gw_subplots/baselines/gw_injected_6_dns_evals_5002325_bins_30_subplot_24}
  \includegraphics[width=0.06\textwidth, frame]{figures/experiments/gw_subplots/baselines/gw_injected_6_dns_evals_5002325_bins_30_subplot_25}
  \includegraphics[width=0.06\textwidth, frame]{figures/experiments/gw_subplots/baselines/gw_injected_6_dns_evals_5002325_bins_30_subplot_26}
  \includegraphics[width=0.06\textwidth, frame]{figures/experiments/gw_subplots/baselines/gw_injected_6_dns_evals_5002325_bins_30_subplot_27}
  \includegraphics[width=0.06\textwidth, frame]{figures/experiments/gw_subplots/baselines/gw_injected_6_dns_evals_5002325_bins_30_subplot_30}
  \includegraphics[width=0.06\textwidth, frame]{figures/experiments/gw_subplots/baselines/gw_injected_6_dns_evals_5002325_bins_30_subplot_31}
  \includegraphics[width=0.06\textwidth, frame]{figures/experiments/gw_subplots/baselines/gw_injected_6_dns_evals_5002325_bins_30_subplot_32}
  \includegraphics[width=0.06\textwidth, frame]{figures/experiments/gw_subplots/baselines/gw_injected_6_dns_evals_5002325_bins_30_subplot_33}
  \includegraphics[width=0.06\textwidth, frame]{figures/experiments/gw_subplots/baselines/gw_injected_6_dns_evals_5002325_bins_30_subplot_34}

  \vspace{0.5cm}

  \includegraphics[width=0.06\textwidth, frame]{figures/experiments/gw_subplots/ablation_study/gw_injected_6_defer_evals_10000003_bins_30_subplot_6}
  \includegraphics[width=0.06\textwidth, frame]{figures/experiments/gw_subplots/ablation_study/gw_injected_6_defer_evals_10000003_bins_30_subplot_12}
  \includegraphics[width=0.06\textwidth, frame]{figures/experiments/gw_subplots/ablation_study/gw_injected_6_defer_evals_10000003_bins_30_subplot_13}
  \includegraphics[width=0.06\textwidth, frame]{figures/experiments/gw_subplots/ablation_study/gw_injected_6_defer_evals_10000003_bins_30_subplot_18}
  \includegraphics[width=0.06\textwidth, frame]{figures/experiments/gw_subplots/ablation_study/gw_injected_6_defer_evals_10000003_bins_30_subplot_19}
  \includegraphics[width=0.06\textwidth, frame]{figures/experiments/gw_subplots/ablation_study/gw_injected_6_defer_evals_10000003_bins_30_subplot_20}
  \includegraphics[width=0.06\textwidth, frame]{figures/experiments/gw_subplots/ablation_study/gw_injected_6_defer_evals_10000003_bins_30_subplot_24}
  \includegraphics[width=0.06\textwidth, frame]{figures/experiments/gw_subplots/ablation_study/gw_injected_6_defer_evals_10000003_bins_30_subplot_25}
  \includegraphics[width=0.06\textwidth, frame]{figures/experiments/gw_subplots/ablation_study/gw_injected_6_defer_evals_10000003_bins_30_subplot_26}
  \includegraphics[width=0.06\textwidth, frame]{figures/experiments/gw_subplots/ablation_study/gw_injected_6_defer_evals_10000003_bins_30_subplot_27}
  \includegraphics[width=0.06\textwidth, frame]{figures/experiments/gw_subplots/ablation_study/gw_injected_6_defer_evals_10000003_bins_30_subplot_30}
  \includegraphics[width=0.06\textwidth, frame]{figures/experiments/gw_subplots/ablation_study/gw_injected_6_defer_evals_10000003_bins_30_subplot_31}
  \includegraphics[width=0.06\textwidth, frame]{figures/experiments/gw_subplots/ablation_study/gw_injected_6_defer_evals_10000003_bins_30_subplot_32}
  \includegraphics[width=0.06\textwidth, frame]{figures/experiments/gw_subplots/ablation_study/gw_injected_6_defer_evals_10000003_bins_30_subplot_33}
  \includegraphics[width=0.06\textwidth, frame]{figures/experiments/gw_subplots/ablation_study/gw_injected_6_defer_evals_10000003_bins_30_subplot_34}

  \includegraphics[width=0.06\textwidth, frame]{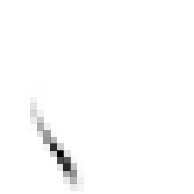}
  \includegraphics[width=0.06\textwidth, frame]{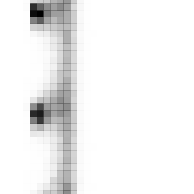}
  \includegraphics[width=0.06\textwidth, frame]{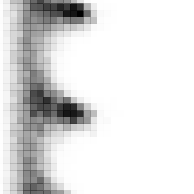}
  \includegraphics[width=0.06\textwidth, frame]{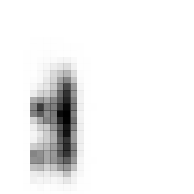}
  \includegraphics[width=0.06\textwidth, frame]{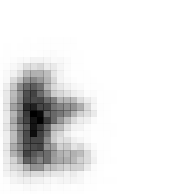}
  \includegraphics[width=0.06\textwidth, frame]{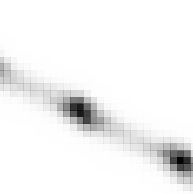}
  \includegraphics[width=0.06\textwidth, frame]{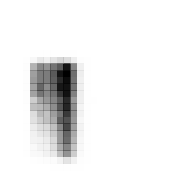}
  \includegraphics[width=0.06\textwidth, frame]{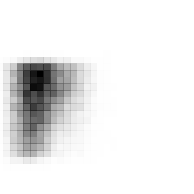}
  \includegraphics[width=0.06\textwidth, frame]{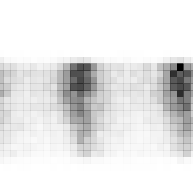}
  \includegraphics[width=0.06\textwidth, frame]{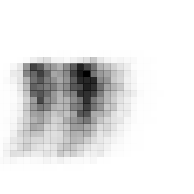}
  \includegraphics[width=0.06\textwidth, frame]{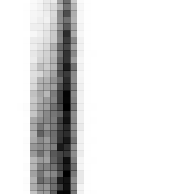}
  \includegraphics[width=0.06\textwidth, frame]{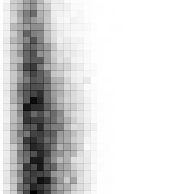}
  \includegraphics[width=0.06\textwidth, frame]{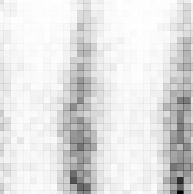}
  \includegraphics[width=0.06\textwidth, frame]{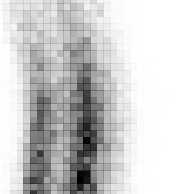}
  \includegraphics[width=0.06\textwidth, frame]{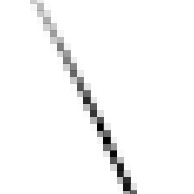}

  \includegraphics[width=0.06\textwidth, frame]{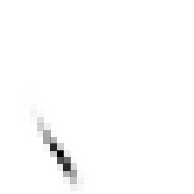}
  \includegraphics[width=0.06\textwidth, frame]{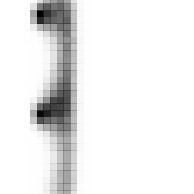}
  \includegraphics[width=0.06\textwidth, frame]{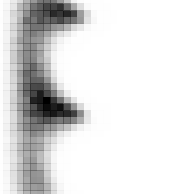}
  \includegraphics[width=0.06\textwidth, frame]{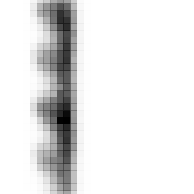}
  \includegraphics[width=0.06\textwidth, frame]{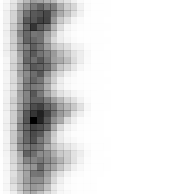}
  \includegraphics[width=0.06\textwidth, frame]{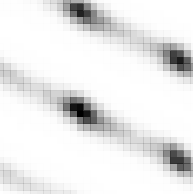}
  \includegraphics[width=0.06\textwidth, frame]{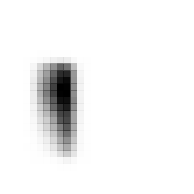}
  \includegraphics[width=0.06\textwidth, frame]{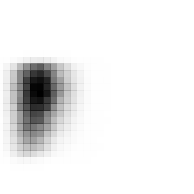}
  \includegraphics[width=0.06\textwidth, frame]{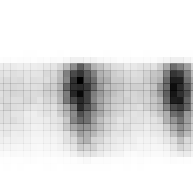}
  \includegraphics[width=0.06\textwidth, frame]{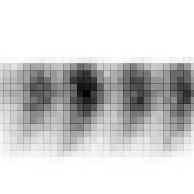}
  \includegraphics[width=0.06\textwidth, frame]{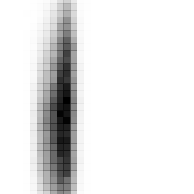}
  \includegraphics[width=0.06\textwidth, frame]{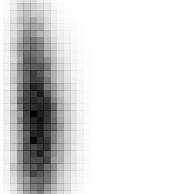}
  \includegraphics[width=0.06\textwidth, frame]{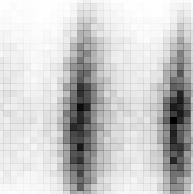}
  \includegraphics[width=0.06\textwidth, frame]{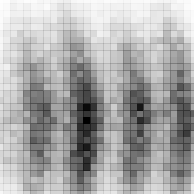}
  \includegraphics[width=0.06\textwidth, frame]{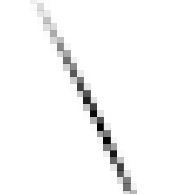}

  \caption{
  Gravitional-physics parameter inference using the different methods.
  Shown are histograms of the two-dimensional marginals of the 6D problem (see Section~\ref{sec:gw_problem_details}),
  with samples produced after 200k density functions evaluations in the upper three rows,
  after 5M evaluations in the middle three rows,
  and efter 10M evaluations in the bottom three rows.
  Shown in order, per group of three, is DEFER, PTMCMC and DNS.
  DEFER is able to capture the surface well.
  PTMCMC captures areas around some modes well, but fails to capture all modes.
  Note that we perform independent runs for each budget.
  For another gravitional-physics example, see Figure~\ref{fig:gw2_baselines}.
  }
  \label{fig:gw_baselines}
\end{figure*}

\begin{figure*}[h]
  \centering

  \includegraphics[width=0.06\textwidth, frame]{figures/experiments/gw2_subplots/ablation_study/gw_injected_2_6_defer_evals_500003_bins_30_subplot_6}
  \includegraphics[width=0.06\textwidth, frame]{figures/experiments/gw2_subplots/ablation_study/gw_injected_2_6_defer_evals_500003_bins_30_subplot_12}
  \includegraphics[width=0.06\textwidth, frame]{figures/experiments/gw2_subplots/ablation_study/gw_injected_2_6_defer_evals_500003_bins_30_subplot_13}
  \includegraphics[width=0.06\textwidth, frame]{figures/experiments/gw2_subplots/ablation_study/gw_injected_2_6_defer_evals_500003_bins_30_subplot_18}
  \includegraphics[width=0.06\textwidth, frame]{figures/experiments/gw2_subplots/ablation_study/gw_injected_2_6_defer_evals_500003_bins_30_subplot_19}
  \includegraphics[width=0.06\textwidth, frame]{figures/experiments/gw2_subplots/ablation_study/gw_injected_2_6_defer_evals_500003_bins_30_subplot_20}
  \includegraphics[width=0.06\textwidth, frame]{figures/experiments/gw2_subplots/ablation_study/gw_injected_2_6_defer_evals_500003_bins_30_subplot_24}
  \includegraphics[width=0.06\textwidth, frame]{figures/experiments/gw2_subplots/ablation_study/gw_injected_2_6_defer_evals_500003_bins_30_subplot_25}
  \includegraphics[width=0.06\textwidth, frame]{figures/experiments/gw2_subplots/ablation_study/gw_injected_2_6_defer_evals_500003_bins_30_subplot_26}
  \includegraphics[width=0.06\textwidth, frame]{figures/experiments/gw2_subplots/ablation_study/gw_injected_2_6_defer_evals_500003_bins_30_subplot_27}
  \includegraphics[width=0.06\textwidth, frame]{figures/experiments/gw2_subplots/ablation_study/gw_injected_2_6_defer_evals_500003_bins_30_subplot_30}
  \includegraphics[width=0.06\textwidth, frame]{figures/experiments/gw2_subplots/ablation_study/gw_injected_2_6_defer_evals_500003_bins_30_subplot_31}
  \includegraphics[width=0.06\textwidth, frame]{figures/experiments/gw2_subplots/ablation_study/gw_injected_2_6_defer_evals_500003_bins_30_subplot_32}
  \includegraphics[width=0.06\textwidth, frame]{figures/experiments/gw2_subplots/ablation_study/gw_injected_2_6_defer_evals_500003_bins_30_subplot_33}
  \includegraphics[width=0.06\textwidth, frame]{figures/experiments/gw2_subplots/ablation_study/gw_injected_2_6_defer_evals_500003_bins_30_subplot_34}

  \includegraphics[width=0.06\textwidth, frame]{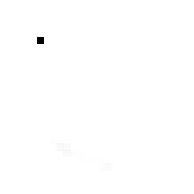}
  \includegraphics[width=0.06\textwidth, frame]{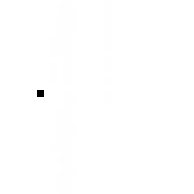}
  \includegraphics[width=0.06\textwidth, frame]{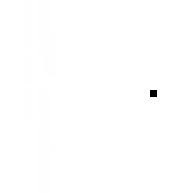}
  \includegraphics[width=0.06\textwidth, frame]{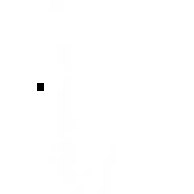}
  \includegraphics[width=0.06\textwidth, frame]{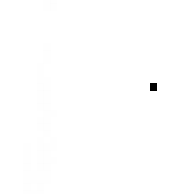}
  \includegraphics[width=0.06\textwidth, frame]{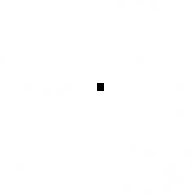}
  \includegraphics[width=0.06\textwidth, frame]{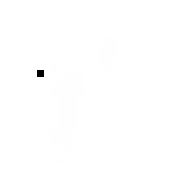}
  \includegraphics[width=0.06\textwidth, frame]{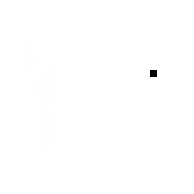}
  \includegraphics[width=0.06\textwidth, frame]{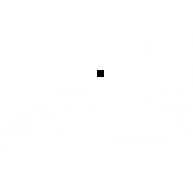}
  \includegraphics[width=0.06\textwidth, frame]{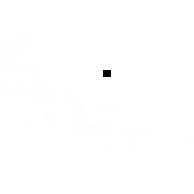}
  \includegraphics[width=0.06\textwidth, frame]{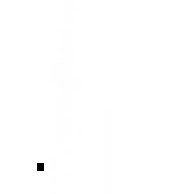}
  \includegraphics[width=0.06\textwidth, frame]{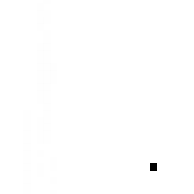}
  \includegraphics[width=0.06\textwidth, frame]{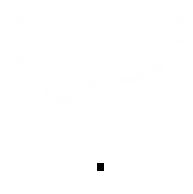}
  \includegraphics[width=0.06\textwidth, frame]{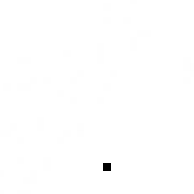}
  \includegraphics[width=0.06\textwidth, frame]{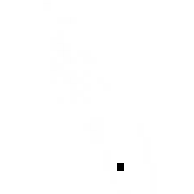}

  \includegraphics[width=0.06\textwidth, frame]{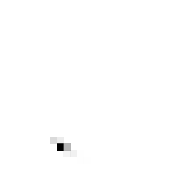}
  \includegraphics[width=0.06\textwidth, frame]{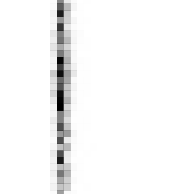}
  \includegraphics[width=0.06\textwidth, frame]{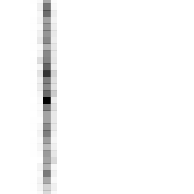}
  \includegraphics[width=0.06\textwidth, frame]{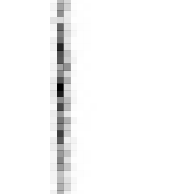}
  \includegraphics[width=0.06\textwidth, frame]{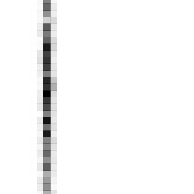}
  \includegraphics[width=0.06\textwidth, frame]{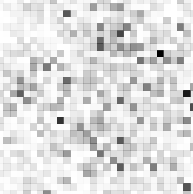}
  \includegraphics[width=0.06\textwidth, frame]{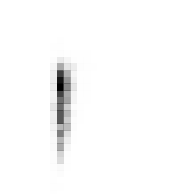}
  \includegraphics[width=0.06\textwidth, frame]{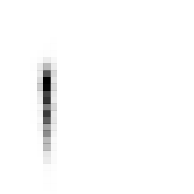}
  \includegraphics[width=0.06\textwidth, frame]{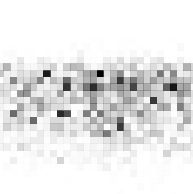}
  \includegraphics[width=0.06\textwidth, frame]{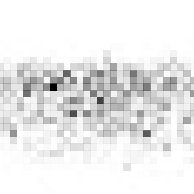}
  \includegraphics[width=0.06\textwidth, frame]{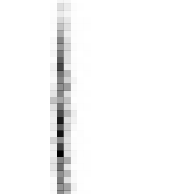}
  \includegraphics[width=0.06\textwidth, frame]{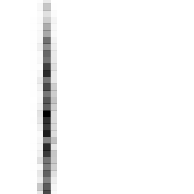}
  \includegraphics[width=0.06\textwidth, frame]{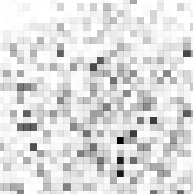}
  \includegraphics[width=0.06\textwidth, frame]{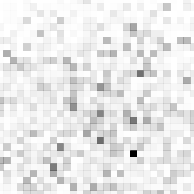}
  \includegraphics[width=0.06\textwidth, frame]{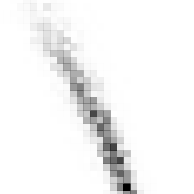}

  \vspace{0.5cm}

  \includegraphics[width=0.06\textwidth, frame]{figures/experiments/gw2_subplots/ablation_study/gw_injected_2_6_defer_evals_5000003_bins_30_subplot_6}
  \includegraphics[width=0.06\textwidth, frame]{figures/experiments/gw2_subplots/ablation_study/gw_injected_2_6_defer_evals_5000003_bins_30_subplot_12}
  \includegraphics[width=0.06\textwidth, frame]{figures/experiments/gw2_subplots/ablation_study/gw_injected_2_6_defer_evals_5000003_bins_30_subplot_13}
  \includegraphics[width=0.06\textwidth, frame]{figures/experiments/gw2_subplots/ablation_study/gw_injected_2_6_defer_evals_5000003_bins_30_subplot_18}
  \includegraphics[width=0.06\textwidth, frame]{figures/experiments/gw2_subplots/ablation_study/gw_injected_2_6_defer_evals_5000003_bins_30_subplot_19}
  \includegraphics[width=0.06\textwidth, frame]{figures/experiments/gw2_subplots/ablation_study/gw_injected_2_6_defer_evals_5000003_bins_30_subplot_20}
  \includegraphics[width=0.06\textwidth, frame]{figures/experiments/gw2_subplots/ablation_study/gw_injected_2_6_defer_evals_5000003_bins_30_subplot_24}
  \includegraphics[width=0.06\textwidth, frame]{figures/experiments/gw2_subplots/ablation_study/gw_injected_2_6_defer_evals_5000003_bins_30_subplot_25}
  \includegraphics[width=0.06\textwidth, frame]{figures/experiments/gw2_subplots/ablation_study/gw_injected_2_6_defer_evals_5000003_bins_30_subplot_26}
  \includegraphics[width=0.06\textwidth, frame]{figures/experiments/gw2_subplots/ablation_study/gw_injected_2_6_defer_evals_5000003_bins_30_subplot_27}
  \includegraphics[width=0.06\textwidth, frame]{figures/experiments/gw2_subplots/ablation_study/gw_injected_2_6_defer_evals_5000003_bins_30_subplot_30}
  \includegraphics[width=0.06\textwidth, frame]{figures/experiments/gw2_subplots/ablation_study/gw_injected_2_6_defer_evals_5000003_bins_30_subplot_31}
  \includegraphics[width=0.06\textwidth, frame]{figures/experiments/gw2_subplots/ablation_study/gw_injected_2_6_defer_evals_5000003_bins_30_subplot_32}
  \includegraphics[width=0.06\textwidth, frame]{figures/experiments/gw2_subplots/ablation_study/gw_injected_2_6_defer_evals_5000003_bins_30_subplot_33}
  \includegraphics[width=0.06\textwidth, frame]{figures/experiments/gw2_subplots/ablation_study/gw_injected_2_6_defer_evals_5000003_bins_30_subplot_34}

  \includegraphics[width=0.06\textwidth, frame]{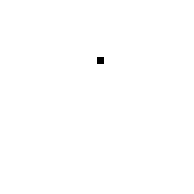}
  \includegraphics[width=0.06\textwidth, frame]{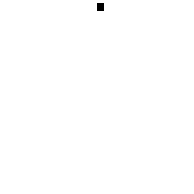}
  \includegraphics[width=0.06\textwidth, frame]{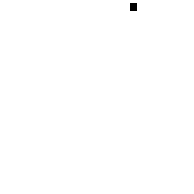}
  \includegraphics[width=0.06\textwidth, frame]{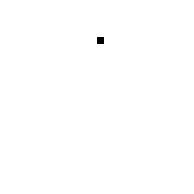}
  \includegraphics[width=0.06\textwidth, frame]{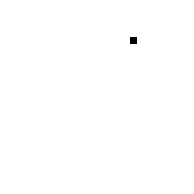}
  \includegraphics[width=0.06\textwidth, frame]{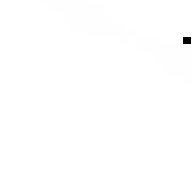}
  \includegraphics[width=0.06\textwidth, frame]{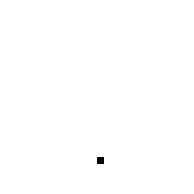}
  \includegraphics[width=0.06\textwidth, frame]{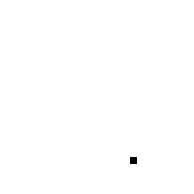}
  \includegraphics[width=0.06\textwidth, frame]{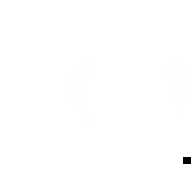}
  \includegraphics[width=0.06\textwidth, frame]{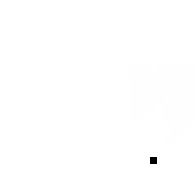}
  \includegraphics[width=0.06\textwidth, frame]{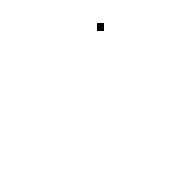}
  \includegraphics[width=0.06\textwidth, frame]{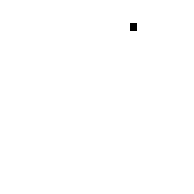}
  \includegraphics[width=0.06\textwidth, frame]{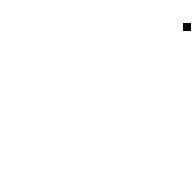}
  \includegraphics[width=0.06\textwidth, frame]{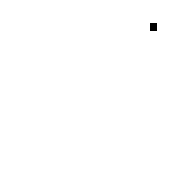}
  \includegraphics[width=0.06\textwidth, frame]{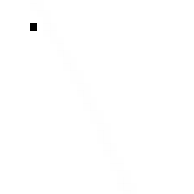}

  \includegraphics[width=0.06\textwidth, frame]{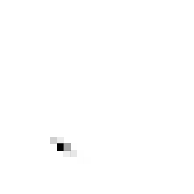}
  \includegraphics[width=0.06\textwidth, frame]{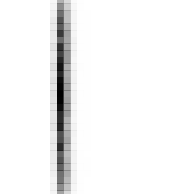}
  \includegraphics[width=0.06\textwidth, frame]{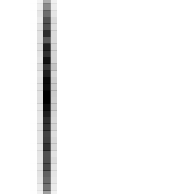}
  \includegraphics[width=0.06\textwidth, frame]{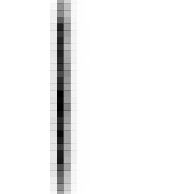}
  \includegraphics[width=0.06\textwidth, frame]{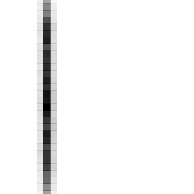}
  \includegraphics[width=0.06\textwidth, frame]{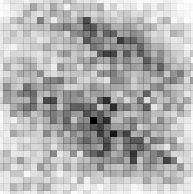}
  \includegraphics[width=0.06\textwidth, frame]{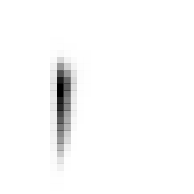}
  \includegraphics[width=0.06\textwidth, frame]{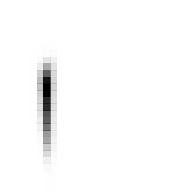}
  \includegraphics[width=0.06\textwidth, frame]{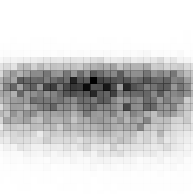}
  \includegraphics[width=0.06\textwidth, frame]{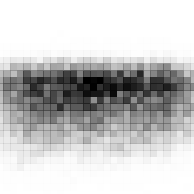}
  \includegraphics[width=0.06\textwidth, frame]{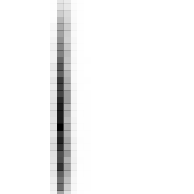}
  \includegraphics[width=0.06\textwidth, frame]{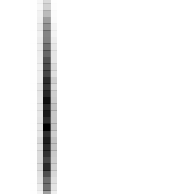}
  \includegraphics[width=0.06\textwidth, frame]{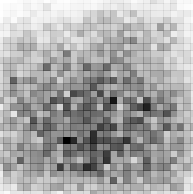}
  \includegraphics[width=0.06\textwidth, frame]{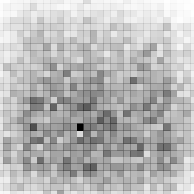}
  \includegraphics[width=0.06\textwidth, frame]{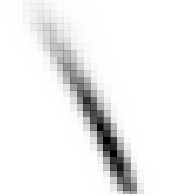}

  \vspace{0.5cm}

  \includegraphics[width=0.06\textwidth, frame]{figures/experiments/gw2_subplots/ablation_study/gw_injected_2_6_defer_evals_10000001_bins_30_subplot_6}
  \includegraphics[width=0.06\textwidth, frame]{figures/experiments/gw2_subplots/ablation_study/gw_injected_2_6_defer_evals_10000001_bins_30_subplot_12}
  \includegraphics[width=0.06\textwidth, frame]{figures/experiments/gw2_subplots/ablation_study/gw_injected_2_6_defer_evals_10000001_bins_30_subplot_13}
  \includegraphics[width=0.06\textwidth, frame]{figures/experiments/gw2_subplots/ablation_study/gw_injected_2_6_defer_evals_10000001_bins_30_subplot_18}
  \includegraphics[width=0.06\textwidth, frame]{figures/experiments/gw2_subplots/ablation_study/gw_injected_2_6_defer_evals_10000001_bins_30_subplot_19}
  \includegraphics[width=0.06\textwidth, frame]{figures/experiments/gw2_subplots/ablation_study/gw_injected_2_6_defer_evals_10000001_bins_30_subplot_20}
  \includegraphics[width=0.06\textwidth, frame]{figures/experiments/gw2_subplots/ablation_study/gw_injected_2_6_defer_evals_10000001_bins_30_subplot_24}
  \includegraphics[width=0.06\textwidth, frame]{figures/experiments/gw2_subplots/ablation_study/gw_injected_2_6_defer_evals_10000001_bins_30_subplot_25}
  \includegraphics[width=0.06\textwidth, frame]{figures/experiments/gw2_subplots/ablation_study/gw_injected_2_6_defer_evals_10000001_bins_30_subplot_26}
  \includegraphics[width=0.06\textwidth, frame]{figures/experiments/gw2_subplots/ablation_study/gw_injected_2_6_defer_evals_10000001_bins_30_subplot_27}
  \includegraphics[width=0.06\textwidth, frame]{figures/experiments/gw2_subplots/ablation_study/gw_injected_2_6_defer_evals_10000001_bins_30_subplot_30}
  \includegraphics[width=0.06\textwidth, frame]{figures/experiments/gw2_subplots/ablation_study/gw_injected_2_6_defer_evals_10000001_bins_30_subplot_31}
  \includegraphics[width=0.06\textwidth, frame]{figures/experiments/gw2_subplots/ablation_study/gw_injected_2_6_defer_evals_10000001_bins_30_subplot_32}
  \includegraphics[width=0.06\textwidth, frame]{figures/experiments/gw2_subplots/ablation_study/gw_injected_2_6_defer_evals_10000001_bins_30_subplot_33}
  \includegraphics[width=0.06\textwidth, frame]{figures/experiments/gw2_subplots/ablation_study/gw_injected_2_6_defer_evals_10000001_bins_30_subplot_34}

  \includegraphics[width=0.06\textwidth, frame]{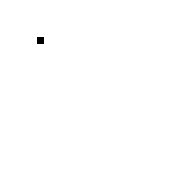}
  \includegraphics[width=0.06\textwidth, frame]{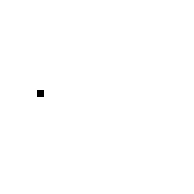}
  \includegraphics[width=0.06\textwidth, frame]{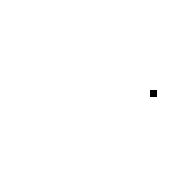}
  \includegraphics[width=0.06\textwidth, frame]{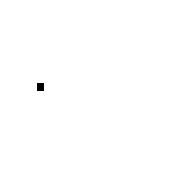}
  \includegraphics[width=0.06\textwidth, frame]{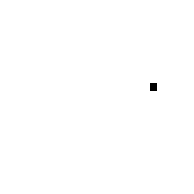}
  \includegraphics[width=0.06\textwidth, frame]{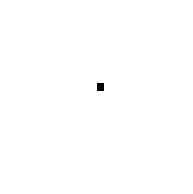}
  \includegraphics[width=0.06\textwidth, frame]{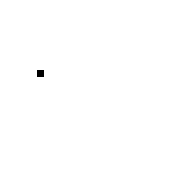}
  \includegraphics[width=0.06\textwidth, frame]{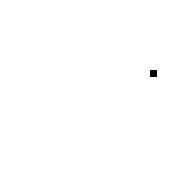}
  \includegraphics[width=0.06\textwidth, frame]{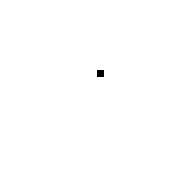}
  \includegraphics[width=0.06\textwidth, frame]{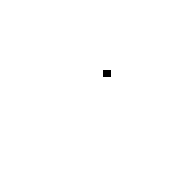}
  \includegraphics[width=0.06\textwidth, frame]{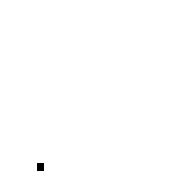}
  \includegraphics[width=0.06\textwidth, frame]{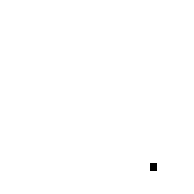}
  \includegraphics[width=0.06\textwidth, frame]{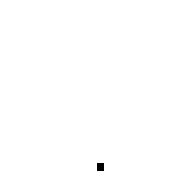}
  \includegraphics[width=0.06\textwidth, frame]{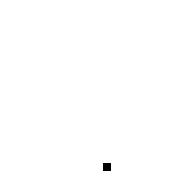}
  \includegraphics[width=0.06\textwidth, frame]{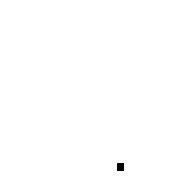}

  \includegraphics[width=0.06\textwidth, frame]{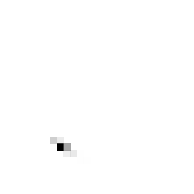}
  \includegraphics[width=0.06\textwidth, frame]{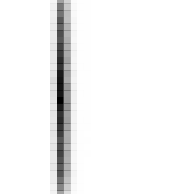}
  \includegraphics[width=0.06\textwidth, frame]{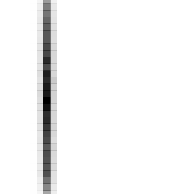}
  \includegraphics[width=0.06\textwidth, frame]{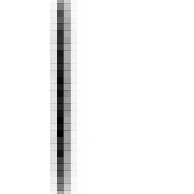}
  \includegraphics[width=0.06\textwidth, frame]{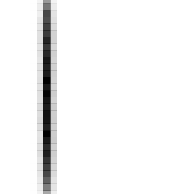}
  \includegraphics[width=0.06\textwidth, frame]{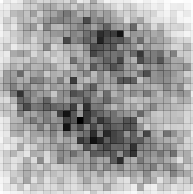}
  \includegraphics[width=0.06\textwidth, frame]{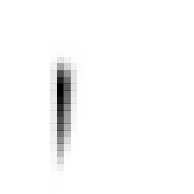}
  \includegraphics[width=0.06\textwidth, frame]{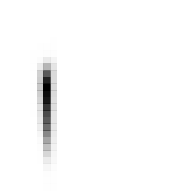}
  \includegraphics[width=0.06\textwidth, frame]{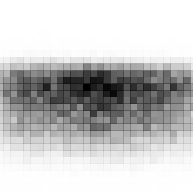}
  \includegraphics[width=0.06\textwidth, frame]{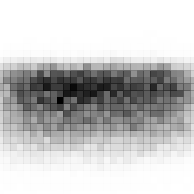}
  \includegraphics[width=0.06\textwidth, frame]{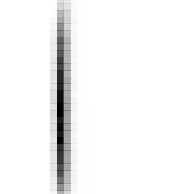}
  \includegraphics[width=0.06\textwidth, frame]{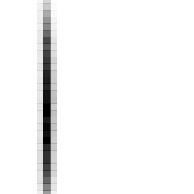}
  \includegraphics[width=0.06\textwidth, frame]{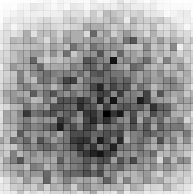}
  \includegraphics[width=0.06\textwidth, frame]{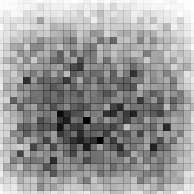}
  \includegraphics[width=0.06\textwidth, frame]{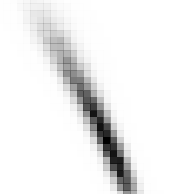}

  \caption{
  Gravitional-physics parameter inference using the different methods.
  Shown are histograms of the two-dimensional marginals of another 6D problem,
  this time inferring the masses of the two black holes instead of the luminosity distance to the source and the orbital phase,
  together with the four other parameters (see Section~\ref{sec:gw_problem_details}).
  The samples are produced after 500k density functions evaluations in the upper three rows,
  after 5M evaluations in the middle three rows,
  and efter 10M evaluations in the bottom three rows.
  Shown in order, per group of three, is DEFER, PTMCMC and DNS.
  DEFER is able to capture the surface in detail.
  PTMCMC got stuck on an insignificant mode within the available budget, and DNS seemingly captures a lower fidelity surface.
  Note that we perform independent runs for each budget.
  Dynamic Nested Sampling (DNS) is designed to automatically adapt the number of live points, which controls
  the fidelty of the nested sampling algorithm.
  A possible explanation for the lack of fidelity of DNS on this example could be a failure of the algorithm to detect a need for additional live points.
  Note that, on the other hand, a too high number of live points for a given problem lowers the overall sample-efficiency of the nested sampling algorithm.
  }
  \label{fig:gw2_baselines}
\end{figure*}

For a small visual ablation study of the partition division criterion (see Section 4 of the paper),
see Figure~\ref{fig:gw_ablation} and Figure~\ref{fig:gw2_ablation}.

\subsection{Baseline setups}
We used the following implementations:
DNS~\cite{speagle2020dynesty},
slice sampling~\cite{abadi2016tensorflow},
and PTMCMC~\cite{justin_ellis_2017_1037579}.

\paragraph{Slice sampling}
We use an initial step size corresponding to $0.05$ of the unit domain sides,
with max\_doublings set to 5.
The used burn-in ratio is 25\%.

\paragraph{PTMCMC}
We follow the default settings and
sample $p_0$ uniformly within the domain and set the initial covariance matrix
to be diagonal with variance 0.01.
We use covUpdate=500, 25\% burn-in and all parameters set to the default.

\paragraph{DNS}
We use the default (as all other settings) of nlive\_init=500, in practice nlive\_init=min(500, 2 + int($N_T$ / 10)) as the max number of function evaluations
$N_T$ in a few experiments are small.
For 'Posterior mode' pfrac = 1.0, and for 'Evidence mode' pfrac = 0.0.

\subsection{Real-world density surfaces}

\subsubsection{Gravitional-wave physics}

\paragraph{Problem background}
Motivation for use cases of this algorithm can be found in natural science research areas such as gravitational-wave physics~\cite{ligo2020gw190412, abbott2019gwtc, mandel2020alternative}.
In GW research, scientific knowledge is often expressed in the form of physically motivated likelihood functions and priors~\cite{PhysRevD.101.103004}.
An increasing sophistication has brought challenges from an inference standpoint.
Tractable gradients are often missing,
the functions may exhibit undefined (or zero density) regions,
and the induced density surfaces tend to be
multi-modal,
discontinuous,
have mass concentrated in tiny regions,
and exhibit complicated correlations.
As a consequence, inference can be prohibitively slow even for problems of around ten dimensions.
Simple techniques such as standard rejection sampling are typically infeasible due to small typical sets,
requiring a prohibitively large number of function evaluations to obtain representative samples.
Markov Chain Monte Carlo methods,
generally popular for their asymptotic guarantees~\cite{geyer1992practical} and strengths in high dimension~\cite{neal2011mcmc},
struggle in handling the multi-modality present in these problems.
Furthermore, MCMC does not provide evidence estimation, which often is of crucial importance in scientific applications and elsewhere.

A popular family of methods to deploy instead is Nested Sampling~\cite{higson2019dynamic,ligo2020gw190412},
known for performing well on multi-modal and degenerate posteriors,
as well as additionally providing evidence estimation.
Still,
the computation required to run an experiment can often be impractical,
such as weeks on a cluster.
Re-sampling, density re-weighting and local density integration have been identified as important tasks to save computation,
when considering different priors~\cite{mandel2020alternative} or estimating equations of state~\cite{vivanco2019measuring}.
Currently these tasks, as well as parameter and evidence estimation,
require a portfolio of different algorithms with various tuning parameters,
requiring significant expertise and effort to obtain reliable results.
DEFER outputs a density function approximation with support for these tasks;
the latter via making use of the domain-indexed search tree over partitions.

We apply DEFER to a simulated signal example from~\cite{ashton2019bilby} similar to~\cite{ligo2020gw190412}.
Shown in the corresponding figure in the paper are all the 2D marginals of a six-dimensional problem using the
`IMRPhenomPv2' waveform approximant.
Inferred parameters are, for example, the luminosity distance, and the spin magnitudes of binary black-holes.
We note the complicated interactions between parameters,
showing the importance of handling multi-modality and strong correlations.
Importantly, DEFER is able to handle the surface well without any tuning parameters.

To see results of the different baselines after varying budgets, see Figure~\ref{fig:gw_baselines} and Figure~\ref{fig:gw2_baselines},
where the latter figure is a modified example where the masses of the two black holes instead of the luminosity distance to the source and the orbital phase
are being inferred, keeping the other four parameters.

\paragraph{Details}
\label{sec:gw_problem_details}

The problem addressed concerns inferring model parameters to explain an event involving two black holes in a binary system.
This
\href{https://git.ligo.org/lscsoft/bilby/blob/master/examples/gw_examples/injection_examples/fast_tutorial.py}{
4D example (link)}
was turned into a 6D example with the injection of all parameters except the following six parameters, which are inferred:
\begin{itemize}
  \item The luminosity distance to the source dL.
  \item Phase: The orbital phase of the binary at the reference time.
  \item Theta JN: The inclination of the system's total angular momentum with respect to the line of sight.
  \item Psi: The polarisation angle describes the orientation of the projection of the binary's orbital momentum vector onto the plane on the sky.
  \item Dimensionless spin magnitude $a_1$.
  \item Dimensionless spin magnitude $a_2$.
\end{itemize}
For more details, see~\cite{veitch2015parameter}.
To produce a deterministic surface the seed of the pseudo-random generator must be fixed.
To set the global seed used by the waveform approximant,
while not fixing the seed for DEFER or the baseline stochastic methods,
we maintain the state of the pseudo-random generator used within and outside the function, respectively.

To see results of the different baselines after varying budgets, see Figure~\ref{fig:gw_baselines} and Figure~\ref{fig:gw2_baselines},
where the latter figure is a modified example where the masses of the two black holes instead of the luminosity distance to the source and the orbital phase
are being inferred, keeping the other four parameters.

\subsubsection{GP regression with spectral mixture kernel}

\paragraph{Problem background}
Apart from inferring a complex posterior surface, we also compare the quality of the posterior samples in terms of prediction accuracy.
We consider a time-series forecasting problem because the parameter posterior of such problems are often complex and multi-modal,
generally making approximate inference more difficult.
We use a Gaussian process (GP) time-series regression model with a spectral mixture kernel (SMK)~\cite{wilson2013gaussian}.
SMK can approximate any stationary kernel including periodic ones by learning the parameters of a Gaussian mixture to represent the kernel spectral density~\cite{bochner1959lectures}.
However, inference of the parameters of the mixture can be difficult as the induced density surface is multi-modal.
In Figure~\ref{fig:spectral_means} we confirm the presence of multi-modality,
which may cause local optimizers and step-wise sampling techniques like MCMC to get stuck in poor solutions.

We use the airline passengers data~\cite{wilson2013gaussian} and use the first 60\% of months for training and the rest for test.
The GP model has three mixture components plus a linear slope, which translates into a 10D parameter inference task.
With a budget of 50k function evaluations, the negative log-likelihoods on the test data are $377.66$, $365.97$, $236.89$ and $\bm{205.50}$,
for slice sampling, PTMCMC, DNS and DEFER, respectively.
For predictions, see Figure~\ref{fig:spectral}.
DEFER performs better than other methods,
which evidenced by the figure,
may be explained by better capturing multi-modality and thus a larger breadth of possible solutions.
In contrast, the MCMC techniques capture only a single mode each.

\paragraph{Details}
\href{https://gpytorch.readthedocs.io/en/latest/examples/01_Exact_GPs/Spectral_Mixture_GP_Regression.html}{
PyTorch Spectral Mixture kernel GP example}.
This example was adapted for the Airline Passengers dataset used in~\cite{wilson2013gaussian}.

\subsection{Synthetic density functions}

The synthetic density functions used in the quantitative experiments are the following.
All domains are scaled to be a unit hybercube.

\paragraph{Student's t}
The scale parameter was $0.01$.
The mean parameter was uniformly sampled $\mathcal{U}(0.2, 0.8)$ per dimension
for each of the 20 runs.
The degrees of freedom was set to $2.5 + (D / 2)$.

\paragraph{Canoe}
\begin{equation}
  f(\bm{\theta}) = \text{max}(v(\bm{\theta}), 0),
\end{equation}
\begin{equation}
  v(\bm{\theta}) = 2 + 5 \mathcal{N}\bm{\theta}|(\bm{\mu}, \bm{\Sigma}_{\text{inner}}) - 10 \mathcal{N}(\bm{\theta}|\bm{\mu}, \bm{\Sigma}_{\text{outer}}),
\end{equation}
where $\bm{\mu} = \bm{0.5}$, $\bm{\Sigma}_{\text{outer}} = a$,
$\bm{\Sigma}_{\text{inner}} = 0.01 (0.95 J_D + 0.05 I_D)$.
and $\bm{\Sigma}_{\text{outer}} = 0.02 (0.60 J_D + 0.40 I_D)$.
$J_D$ is a $D \times D$ matrix of all ones, and $I_D$ is the identity matrix.

\paragraph{Mixture of Gaussians}
\begin{equation}
  f(\bm{\theta}) = 2.5 \mathcal{N}(\bm{\theta}|\bm{\mu}_{\text{a}}, \bm{\Sigma}_{\text{a}}) + \mathcal{N}(\bm{\theta}|\bm{\mu}_{\text{b}}, \bm{\Sigma}_{\text{b}}),
\end{equation}
where $\bm{\mu}_a = [0.6326, 0.7401, 0.7232, 0.2471]$,
$\bm{\mu}_b = [0.5139, 0.4667, 0.3777, 0.7995]$
(sampled from $\mathcal{U}(0.2, 0.8)$),
and
\begin{equation}
  \bm{\Sigma}_{\text{a}} =
  0.01^2
  \begin{bmatrix}
    2.25 & -1.0 & 0 & 0 \\
    -1.0 & 2.25 & 0 & 0 \\
    0 & 0 & 2.25 & 0 \\
    0 & 0 & 0 & 2.25
  \end{bmatrix},
\end{equation}
\begin{equation}
  \bm{\Sigma}_{\text{b}} =
  0.01^2
  \begin{bmatrix}
    2.25^2 & -2.25 & 1.0 & -1.0 \\
    -2.25 & 2.25^2 & 0 & 0 \\
    1.0 & 0 & 2.25^2 & 0 \\
    -1.0 & 0 & 0 & 2.25^2
  \end{bmatrix}.
\end{equation}

\paragraph{Cigar}
\begin{equation}
  f(\bm{\theta}) = \mathcal{N}(\bm{\theta}|\bm{\mu}, \bm{\Sigma}),
\end{equation}
where $\bm{\mu} = \bm{0.5}$ and $\bm{\Sigma} = 0.01(0.99 J_D + 0.01 I_{D})$.
$J_D$ is a $D \times D$ matrix of all ones, and $I_D$ is the identity matrix.

\balance

\end{document}